\documentclass[10pt,twocolumn,letterpaper]{article}

\usepackage{wacv}              

\definecolor{wacvblue}{rgb}{0.21,0.49,0.74}
\usepackage[pagebackref,breaklinks,colorlinks,allcolors=wacvblue]{hyperref}

\def\wacvPaperID{591}
\def\confName{WACV}
\def\confYear{2027}

\usepackage[T1]{fontenc}
\usepackage{amsmath,amssymb,amsthm}
\usepackage{mathtools}
\usepackage{bm}
\usepackage{algorithm,algorithmic}
\usepackage{graphicx}
\usepackage{xcolor}
\usepackage[table]{xcolor}
\usepackage{hyperref}
\usepackage[margin=2.5cm]{geometry}
\usepackage{tikz}
\usetikzlibrary{arrows.meta,positioning,calc}
\usepackage{booktabs}
\usepackage{multirow}
\usepackage{caption,subcaption}
\usepackage{enumitem}
\usepackage[capitalize]{cleveref}
\usepackage[section]{placeins}

\crefname{figure}{Fig.}{Figs.}
\crefname{table}{Tab.}{Tabs.}
\crefname{equation}{Eq.}{Eqs.}
\crefname{section}{Sec.}{Secs.}
\crefname{theorem}{Thm.}{Thms.}

\Crefname{figure}{Figure}{Figures}
\Crefname{table}{Table}{Tables}
\Crefname{equation}{Equation}{Equations}
\Crefname{section}{Section}{Sections}
\Crefname{Theorem}{Theorem}{Theorems}

\graphicspath{{images/}}
\newcommand{\R}{\mathbb{R}}
\newcommand{\img}{\bm{x}}

\newcommand{\repr}{f}
\newcommand{\augm}{\mathcal{A}}

\newcommand{\our}{SEAMS}

\DeclareMathOperator{\sigmoid}{sigmoid}
\usepackage{comment}
\title{What Pixels Are Enough?\\
  SEAMS: Sufficiency Saliency via MSE-Preservation Soft-Masks}

\author{
\textbf{Magdalena Tr\k{e}dowicz}$^{2,6,7}$ \\
{\tt\small magdalena.tredowicz@doctoral.uj.edu.pl}
\and
\textbf{\L{}ukasz Struski}$^{1,6,7}$ \\
{\tt\small lukasz.struski@uj.edu.pl}
\and
\textbf{Arkadiusz Lewicki}$^{4,6,7}$ \\
{\tt\small alewicki@wsiz.edu.pl}
\and
\textbf{Karolina Pachota}$^{5}$ \\
{\tt\small karolina.pachota@student.uj.edu.pl}
\and
\textbf{Andrzej Grudzie\'n}$^{5}$ \\
{\tt\small andrzej.grudzien@uj.edu.pl}
\and
\textbf{Mateusz Jag\l{}a}$^{5}$ \\
{\tt\small mateusz.jagla@uj.edu.pl}
\and
\textbf{Jacek Tabor}$^{1,3,6,7}$ \\
{\tt\small jacek.tabor@uj.edu.pl}
\and
\parbox{\linewidth}{\small\raggedright
$^{1}$\,Faculty of Mathematics and Computer Science, Jagiellonian University, Poland; \quad
$^{2}$\,Doctoral School of Exact and Natural Sciences, Jagiellonian University, Poland; \quad
$^{3}$\,Centre for Credible Artificial Intelligence, Warsaw University of Technology, Poland; \quad
$^{4}$\,University of Information Technology and Management in Rzeszów, Poland; \quad
$^{5}$\,Jagiellonian University Medical College, Poland; \quad
$^{6}$\,Prometheus MedTech.AI, Inc., USA; \quad
$^{7}$\,UES Sp. z o.o., Poland
}
}

\begin{document}
\maketitle

\begin{abstract}
Saliency maps are most useful when they identify the image regions that are sufficient to preserve a model's behaviour. We introduce \our{}, a sufficiency-based saliency method that directly optimises a soft mask using a preservation objective. Given a frozen differentiable model output, such as a class probability, CLS embedding, or token representation, \our{} searches for a compact mask that preserves the selected output. The approach relies on a simple optimisation framework based on soft masks, a learnable budget, and a three-way image composite generated entirely from the query image. As a result, it requires no auxiliary distractor dataset, architecture-specific attribution mechanism, or differentiable top-k relaxation. Experiments with frozen ViT-S/16 and ConvNeXt models show that the same optimisation pipeline can generate object-level, class-conditioned, and token-level explanations by changing only the preserved target. The resulting masks are compact, interpretable, stable across random initialisations, and competitive on insertion and deletion benchmarks. Our results also indicate that different architectures often rely on different sufficient evidence while achieving similar preservation fidelity, highlighting the architecture-dependent nature of visual explanations.
\noindent Project page: \url{https://github.com/gmum/SEAMS}
\end{abstract}
\vspace{-0.4cm}

\section{Introduction}
\label{sec:intro}

\begin{figure}[t]
\centering

\newcommand{\myfig}[1]{\includegraphics[width=\linewidth]{images/#1}}

\newcommand{\salrow}[2]{%
    \par\noindent
    \begin{minipage}[c]{7px}
        \rotatebox{90}{\makebox[7px][c]{\scriptsize #1}}
    \end{minipage}%
    \begin{minipage}[c]{\dimexpr\columnwidth-7px\relax}
        \myfig{#2}
    \end{minipage}%
}

\begin{minipage}[c]{0.33\columnwidth}
    \centering\scriptsize Input
\end{minipage}%
\begin{minipage}[c]{0.33\columnwidth}
    \centering\scriptsize Mask
\end{minipage}%
\begin{minipage}[c]{0.33\columnwidth}
    \centering\scriptsize Mask on Grey
\end{minipage}

\salrow{ViT-S/16 @ 224}{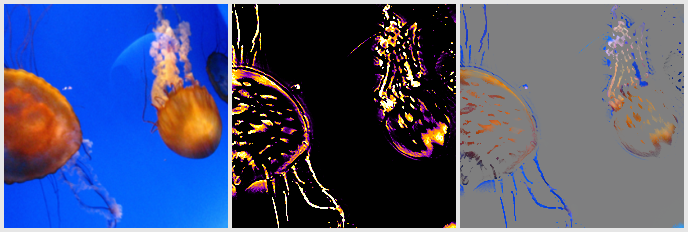}\\[-5px]
\salrow{DINOv2 ViT-S/14 @ 518}{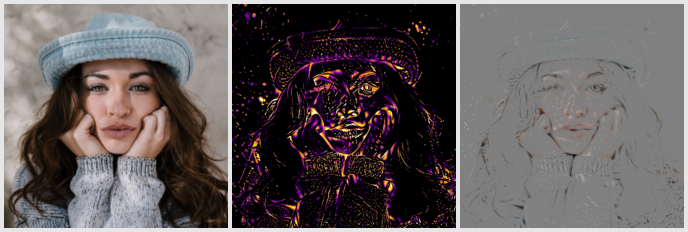}

\caption{\textbf{Sufficiency-based saliency across scales and architectures.} \our{} identifies compact image regions that are sufficient to preserve a selected model representation. Top: ImageNet image analysed with a supervised ViT-S/16 encoder. Bottom: DIV2K image analysed with a self-supervised DINOv2 ViT-S/14 encoder. From left to right: original image, optimised soft mask, and the corresponding saliency overlay. Despite differences in image resolution, training paradigm, and architecture, the same optimisation framework consistently identifies a small subset of pixels sufficient to preserve the target representation.}
\vspace{-0.6cm}
\label{fig:teaser-pair}
\end{figure}

Understanding which image regions are responsible for a neural network prediction remains a central problem in explainable artificial intelligence. In computer vision, explanations should be spatially precise enough to distinguish relevant object parts from background content, allowing researchers to verify whether a model relies on meaningful visual evidence. This challenge has become particularly important with the growing adoption of deep convolutional networks and Vision Transformers, whose internal representations are difficult to interpret directly. Most existing visual explanation methods are based on gradient attribution. Representative examples include vanilla saliency maps \cite{simonyan2014deep}, SmoothGrad \cite{smilkov2017smoothgrad}, DeepLIFT \cite{shrikumar2017learning}, Integrated Gradients \cite{sundararajan2017axiomatic}, and Layer-wise Relevance Propagation \cite{bach2015pixel}. More recently, transformer-oriented approaches such as DAVE \cite{wrobel2026dave} have demonstrated that high-quality explanations can also be obtained for ViT architectures. Although these methods differ substantially in their implementation, they share a common principle: they estimate how sensitive a model output is to small changes in the input.

However, sensitivity and sufficiency are fundamentally different concepts. A pixel may receive a large attribution score because a small perturbation strongly affects the output, while the same information may still be redundantly encoded elsewhere in the image. Conversely, pixels with relatively small gradients may nevertheless be necessary for preserving a representation or prediction. As a result, sensitivity-based explanations do not directly answer the question of which visual evidence is sufficient for maintaining model behaviour.
\begin{figure}[t]
\centering

\newcommand{\myfig}[1]{\includegraphics[width=\linewidth]
{images/fig2/vit_small_patch14_dinov2.lvd142m/#1}}

\newcommand{\salrow}[2]{%
    \par\noindent
    \begin{minipage}[c]{7px}
        \rotatebox{90}{\makebox[7px][c]{\scriptsize #1}}
    \end{minipage}%
    \begin{minipage}[c]{\dimexpr\columnwidth-7px\relax}
        \myfig{#2}
    \end{minipage}%
}

\begin{minipage}[c]{0.25\columnwidth}
    \centering\scriptsize \quad Input
\end{minipage}%
\begin{minipage}[c]{0.25\columnwidth}
    \centering\scriptsize \quad Class Probability
\end{minipage}%
\begin{minipage}[c]{0.25\columnwidth}
    \centering\scriptsize \quad CLS Embedding
\end{minipage}%
\begin{minipage}[c]{0.25\columnwidth}
    \centering\scriptsize Patch-Token
\end{minipage}

\salrow{Ant}{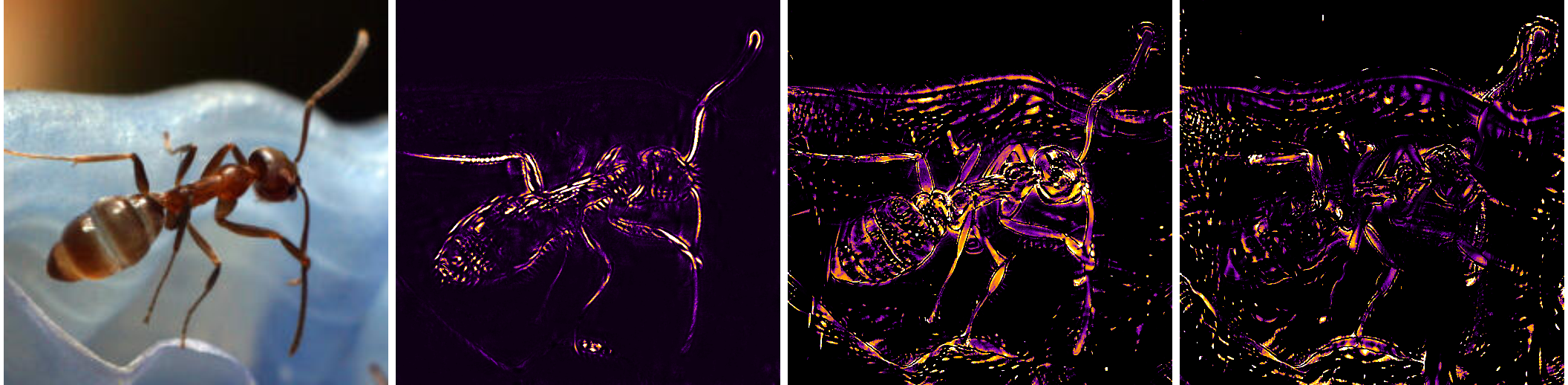}
\salrow{Lycaenid}{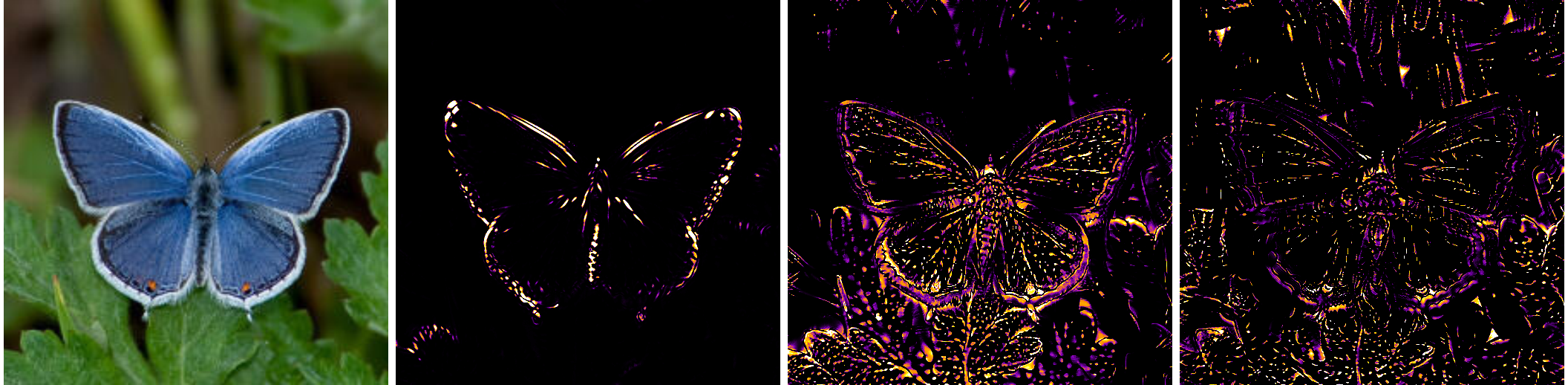}
\salrow{Violin}{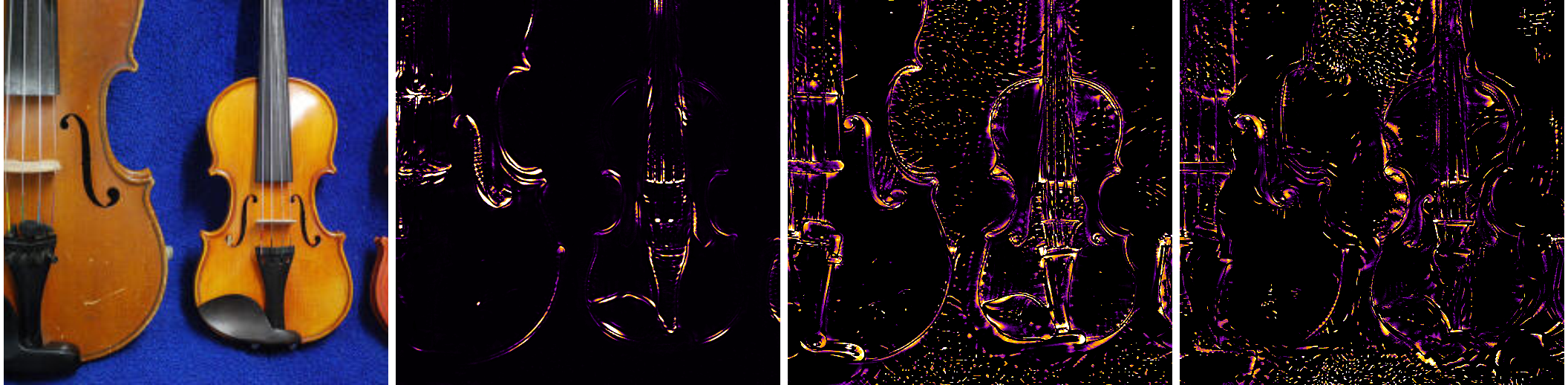}

\caption{\textbf{One optimisation pipeline, multiple explanatory targets.} Each row corresponds to a single ImageNet image, while each column preserves a different target output: a class probability, the CLS embedding, or the patch-token representation. The optimisation procedure, regularisation, and image composite remain unchanged. Different targets produce different saliency patterns, showing that distinct components of a visual representation rely on different subsets of image pixels.}
\label{fig:compare-g}
\vspace{-0.4cm}
\end{figure}

In this work, we adopt a sufficiency-oriented perspective inspired by perturbation-based explanations \cite{fong2017interpretable,fong2019understanding,carter2019sis}. Rather than measuring local sensitivity, we search for the smallest image region that preserves a selected model output. The key result is illustrated in~\cref{fig:teaser-pair}. Starting from the original image, we learn a continuous soft mask that determines which pixels remain unchanged and which can be replaced without significantly altering the target representation. The resulting masks are compact, visually interpretable, and remain effective across different image resolutions and backbone architectures. \Cref{fig:teaser-pair} demonstrates that the pipeline transfers cleanly to higher input resolution and to a self-supervised backbone with no code changes beyond the input normalisation and image size: we use DINOv2 ViT-S/14 at the model's native $518\!\times\!518$ resolution, with the
same loss, composite, augmentation and hyperparameters.

\our{} is fully post hoc and model agnostic. Given a frozen encoder and a differentiable target output, the method optimises a soft mask so that the output computed from the masked image remains close to the output of the original image. Importantly, the target is modular and can represent different explanatory objectives, including a class probability, a global CLS embedding, or a dense token representation. Consequently, the same optimisation procedure can be used to study different aspects of a model without modifying the underlying architecture. This flexibility is demonstrated in~\cref{fig:compare-g}, where exactly the same optimisation pipeline is applied to three different targets. By changing only the preserved output, \our{} produces class-conditioned explanations, global representation masks, and token-level preservation maps. The resulting masks focus on different image regions, indicating that distinct components of a visual representation rely on different subsets of pixels. This observation highlights the importance of preserving a well-defined target when constructing explanations.

The optimisation is enabled by three simple design choices. First, unconstrained logits are transformed into a sparse soft mask through a normalise-and-clip parameterisation with a learnable budget. Second, removed image content is replaced using a three-way composite generated entirely from the query image, combining original content, a self-augmented distractor, and heavily blurred context. Third, augmentations are applied to the composite image while the reference representation remains fixed, encouraging robust and stable explanations. Unlike many perturbation-based methods, the framework requires no auxiliary distractor dataset, no architecture-specific attribution mechanism, and no differentiable sorting or top-k approximation.

Experiments on ViT-S/16 and ConvNeXt-Tiny demonstrate that the same optimisation framework generates compact and stable explanations across architectures while achieving competitive insertion and deletion performance. Interestingly, different architectures often identify different sufficient image regions despite achieving comparable preservation fidelity, suggesting that visual sufficiency is inherently architecture dependent. Our \textbf{contributions} are as follows:
\begin{itemize}[nosep]
  \item We introduce a sufficiency-based formulation of saliency that directly optimises preservation fidelity rather than local sensitivity.
  \item We propose a simple soft-mask optimisation framework that requires neither auxiliary distractor datasets nor specialised differentiable subset-selection techniques.
  \item We demonstrate that a single optimisation pipeline can generate multiple forms of explanation across different architectures by changing only the preserved target representation.
\end{itemize}

\section{Related work}
\label{sec:related}

Research on visual explanations for deep neural networks spans several related directions, including gradient-based attribution, class-discriminative localisation, perturbation-based explanation, and sufficiency-oriented reasoning. While these approaches differ in their objectives and assumptions, they all aim to identify image regions that contribute to a model's prediction or representation. Our work is most closely related to perturbation-based and sufficiency-driven methods, but differs in its use of a simple soft-mask optimisation framework that can preserve arbitrary differentiable targets without requiring auxiliary datasets, architecture-specific attribution rules, or differentiable subset-selection operators.

\smallskip
\noindent\textbf{Gradient-based attribution.}\, Gradient-based attribution methods explain model behaviour through local input sensitivity. Early work introduced saliency maps computed directly from image gradients \cite{simonyan2014deep}. Subsequent approaches addressed gradient saturation and noise through alternative attribution mechanisms, including DeepLIFT \cite{shrikumar2017learning}, Integrated Gradients \cite{sundararajan2017axiomatic}, SmoothGrad \cite{smilkov2017smoothgrad}, and Layer-wise Relevance Propagation \cite{bach2015pixel}. Although these methods differ in formulation, they all estimate how small input perturbations affect the output. As a result, they primarily capture sensitivity rather than identifying information sufficient to preserve a prediction or representation. Sanity checks based on model and data randomization have further highlighted the importance of evaluating whether saliency maps depend on the underlying model \cite{adebayo2018sanity}.

\smallskip
\noindent\textbf{Class-discriminative localisation.}\, Class Activation Mapping methods localise evidence responsible for a particular prediction. Representative examples include CAM \cite{zhou2016learning}, Grad-CAM~\cite{selvaraju2017grad}, Grad-CAM++ \cite{chattopadhyay2018grad}, Score-CAM \cite{wang2020score}, XGrad-CAM~\cite{fu2020axiom},  Layer-CAM~\cite{jiang2021layercam}, and sanity-preserving variants such as CAMERAS~\cite{jalwana2021cameras}. These techniques often produce intuitive object-level heatmaps. However, they are typically designed for class-specific outputs and depend on particular internal representations. In contrast, our formulation operates on an arbitrary differentiable target, including class probabilities, embeddings, and dense token representations. Recent work such as FAME \cite{zhang2026fame} further explores feature-activation-based visual explanations.

\smallskip
\noindent\textbf{Perturbation-based explanations.}\, Perturbation methods define importance through the effect of removing or revealing information. Meaningful Perturbations \cite{fong2017interpretable} introduced continuous mask optimisation, while Extremal Perturbations \cite{fong2019understanding} formulated explanation as selecting fixed-area regions that maximise evidence. RISE \cite{petsiuk2018rise} established insertion and deletion metrics through random masking, and Dabkowski and Gal \cite{dabkowski2017real} proposed amortised saliency generation. Later work improved optimisation procedures through integrated-gradient guidance including I-GOS \cite{qi2021visualizing} and iGOS++ \cite{khorram2021igos}. Vision DiffMask \cite{nalmpantis2023vision} extended preservation-based masking to Vision Transformers using differentiable patch selection. Our method belongs to this family but differs in several respects. The preserved target is fully modular, contextual fill is generated directly from the query image, and sparsity is obtained through a simple normalise-and-clip parameterisation rather than differentiable sorting or top-(k) approximations.

\smallskip
\noindent\textbf{Sufficiency and visual localisation.}\, The notion of explanation through sufficiency was formalised by Sufficient Input Subsets (SIS) \cite{carter2019sis}, which searches for minimal feature subsets that preserve a prediction. Our work follows the same conceptual direction but operates on continuous pixel-level masks and arbitrary differentiable outputs rather than discrete feature subsets and thresholded decisions. Related ideas also appear in Salient Object Detection (SOD), where the goal is to identify visually important regions \cite{wang2023salient,zhu2024visionmamba}. More recently, foundation models have become important sources of localisation signals. Segment Anything (SAM) \cite{kirillov2023segment} and its adaptations, including AlignSAM \cite{huang2024alignsam}, demonstrated strong generalisation capabilities for object localisation. Diffusion-based approaches such as DiffSeg \cite{tian2023diffseg} further showed that attention maps learned by generative models contain rich segmentation information. Unlike these methods, our objective is not object segmentation or saliency prediction. Instead, we directly optimise a compact set of pixels that is sufficient to preserve a chosen model output, providing a general framework for representation-oriented visual explanation.


\section{Method}
\label{sec:method}

\begin{figure*}[t]
\centering
\includegraphics[width=\linewidth]{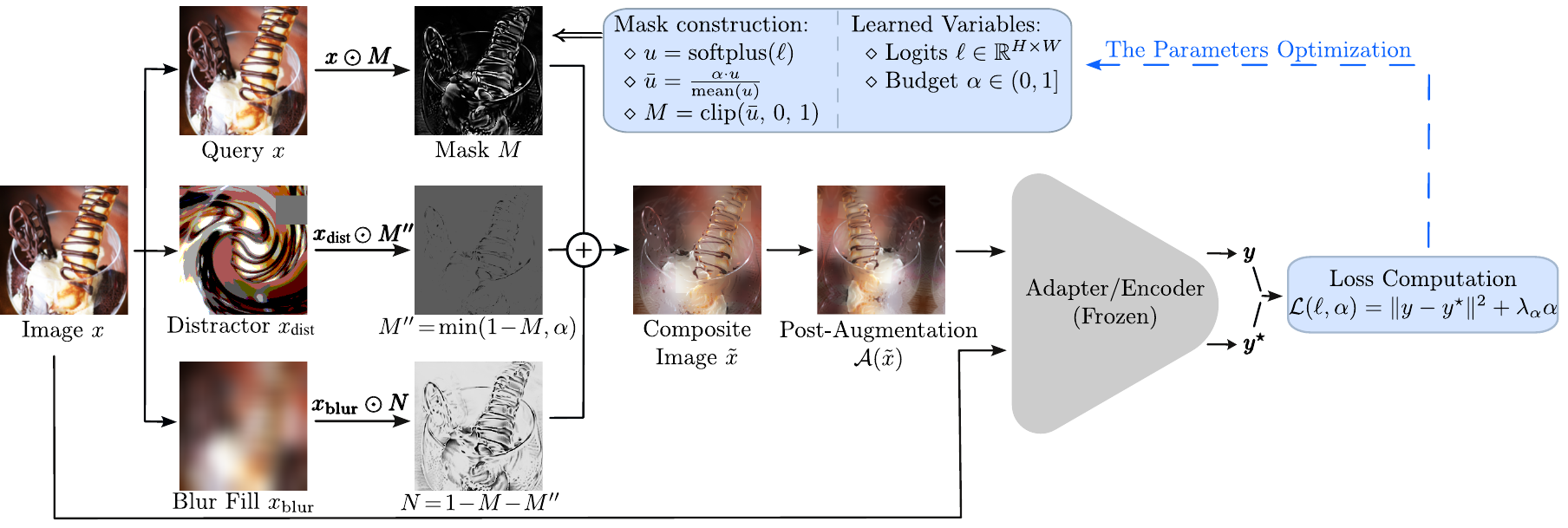}
\caption{\textbf{Overview of the \our{} framework.}
The target representation $y^{\star}$ is first computed from the original image and then kept fixed. A learnable soft mask determines which pixels are preserved, replaced by a self-generated distractor, or substituted with blurred context. The resulting composite image is augmented and evaluated by the frozen model. Optimisation updates only the mask logits and the sparsity budget, encouraging a compact image region that remains sufficient to preserve the selected target representation.}
\label{fig:pipeline}
\vspace{-0.5cm}
\end{figure*}

Our objective is to identify a compact subset of image pixels that is sufficient, within the optimization composite, to preserve a selected model output. Given an input image $\img$ and a frozen vision model, we optimise a continuous soft mask that determines which image regions must remain visible in order to maintain a target representation. Pixels selected by the mask are copied from the original image, whereas the remaining content is replaced by a combination of self-generated distractor content and blurred context. The optimisation seeks a trade-off between preservation fidelity and mask compactness. The overall framework is illustrated in \cref{fig:pipeline}.

\smallskip
\noindent\textbf{Preservation target.}\, Let
\(
g:\mathbb{R}^{C\times H\times W}\rightarrow\mathbb{R}^{d}
\)
denote a frozen vision encoder. Rather than explaining a particular internal component of the model, we define a generic differentiable target
\(
g(\img),
\)
which specifies the representation to be preserved. The formulation is independent of the choice of $g$. In this work we consider three targets: class probabilities, CLS embeddings, and dense token representations. The corresponding examples are shown in \cref{fig:compare-g}. For a given image, the reference target
\(
\bm{y}^{\star}=g(\img)
\)
is computed once and remains fixed throughout optimisation. Having fixed the target representation, we seek a compact subset of image pixels that preserves $\bm{y}^{\star}$. Rather than selecting a discrete set of pixels, which would lead to a combinatorial optimisation problem, we optimise a continuous soft mask defined over the image domain. The mask determines which image regions remain visible and which can be replaced without substantially changing the chosen model output. Preservation is evaluated on a composite image constructed from the original query and two automatically generated replacement sources. The mask, together with a learnable sparsity budget, is optimised directly through the preservation objective shown in \cref{fig:pipeline}.

\smallskip
\noindent\textbf{Mask parametrization.}\, The explanation is represented by a soft mask
\(
M\in[0,1]^{H\times W},
\)
defined at the full image resolution. Values close to one indicate pixels that should be preserved, whereas values close to zero indicate pixels that can be replaced. Instead of optimising the mask directly, we optimise unconstrained logits
\(
\ell\in\mathbb{R}^{H\times W},
\)
which are transformed into a valid mask through a sequence of differentiable operations:
\begin{align}
u &= \operatorname{softplus}(\ell), \nonumber\\
\bar u &= \alpha\frac{u}{\operatorname{mean}(u)}, \nonumber\\
M &= \operatorname{clip}(\bar u,0,1), \nonumber
\end{align}
where $\alpha\in(0,1]$ is a learnable sparsity budget. The softplus transformation ensures non-negative activations, while the normalisation step enforces a prescribed average mask mass before clipping. Intuitively, $\alpha$ controls how much image content can remain visible. Increasing the activation assigned to one region automatically reduces the relative allocation available to other regions through the shared normalisation term. The final clipping operation constrains the mask to the interval $[0,1]$ while preserving differentiability almost everywhere. This parameterisation avoids explicit subset-selection operators and does not require differentiable sorting, top-$k$ relaxations, Gumbel sampling, or temperature annealing. Nevertheless, as discussed later in this section, the resulting masks typically become nearly binary during optimisation.

\smallskip
\noindent\textbf{Composite image.}\, To evaluate whether a candidate mask is sufficient, we construct a composite image in which unselected regions are replaced while selected regions remain unchanged. The composite is formed from three image sources: the original query image $\img$, a synthetic distractor
$\img_{\mathrm{dist}}=\mathrm{heavyAug}(\img),
$
and a heavily blurred version
$\img_{\mathrm{blur}}=\mathrm{blur}_{\sigma}(\img).
$
The distractor is generated by applying strong geometric and photometric transformations to the query image, whereas the blurred image preserves only coarse spatial structure. Given the soft mask $M$, the composite image is defined as
\begin{equation}
\label{eq:composite}
\begin{cases}
\tilde{\img}
=
M\odot\img
+
M''\odot\img_{\mathrm{dist}}
+
N\odot\img_{\mathrm{blur}},\\
M+M''+N=1,
\end{cases}
\end{equation}
where
$
M''=\min(1-M,\alpha),
\quad
N=1-M-M''.
$
The coefficients $M$, $M''$, and $N$ therefore form a pixel-wise partition of unity. Pixels assigned to $M$ are copied directly from the original image and constitute the candidate explanation. Pixels assigned to $M''$ are replaced by distractor content that intentionally alters the visual representation, while the remaining pixels are replaced by blurred context that preserves coarse image layout but removes high-frequency details.

This three-way construction serves two complementary purposes. The distractor discourages the optimisation from relying on weak texture statistics or colour correlations that remain visible under simple blurring. At the same time, the blurred component preserves global scene structure and reduces optimisation artefacts that often occur when large image regions are completely removed. Since both replacement images are generated directly from the query image, the method does not require auxiliary distractor datasets or reference image collections. For pixels with $M\approx0$, the replacement approaches the mixture
\[
\alpha\,\img_{\mathrm{dist}}
+
(1-\alpha)\,\img_{\mathrm{blur}},
\]
whereas intermediate values of $M$ produce smooth transitions between preserved and replaced content. As a result, the optimisation remains fully differentiable and can be solved efficiently using standard gradient-based methods.

\smallskip
\noindent\textbf{Optimisation objective.}\, The optimisation objective is independent of the preserved target. Given the reference representation $\bm{y}^{\star}=g(\img)$ and the composite image $\tilde{\img}$ from~\cref{eq:composite}, we minimise a preservation loss defined as
\begin{equation}\label{eq:loss}
  \mathcal{L}(\bm{\ell},\alpha)
    = \underbrace{
        \mathbb{E}_{\augm}\Bigl[
          \bigl\lVert
            g\!\bigl(\augm(\tilde{\img})\bigr) - \bm{y}^{\star}
          \bigr\rVert^{2}
        \Bigr]
      }_{\text{preservation MSE on }g}
      + \;\lambda_{\alpha}\,\alpha.
\end{equation}
The first term measures preservation fidelity under random augmentations, while the second term encourages sparse explanations by penalising large mask budgets.

\smallskip
\noindent\textbf{Augmentation strategy.}\, Two augmentation pipelines are used during optimisation. The first, $\mathrm{heavyAug}(\cdot)$, generates the distractor image appearing in~\cref{eq:composite}. Its purpose is to produce image content whose representation differs substantially from that of the original image while preserving basic image statistics. The pipeline combines strong geometric and photometric transformations, including affine warps, flips, swirl distortion, colour perturbations, posterisation or solarisation, and random cutout operations.

The second augmentation pipeline, denoted by $\augm(\cdot)$, is applied to the composite image inside the preservation loss. Unlike the distractor generator, it uses moderate perturbations intended to improve robustness. The augmentation consists of random cropping, scaling, translation, rotation, horizontal flipping, blur, and colour jitter. Importantly, these perturbations are applied only to the composite image, whereas the reference target $\bm{y}^{\star}$ remains fixed. Consequently, the mask must preserve the selected representation not only for the original image but also under realistic evaluation-time perturbations. Implementation details of both augmentation pipelines are provided in \cref{app:aug}.

\smallskip
\noindent\textbf{Implicit sparsification.}\, Although the optimisation operates entirely on continuous variables, the resulting masks are typically close to binary. This behaviour emerges naturally from the normalise-and-clip parameterisation. The normalisation step enforces a fixed average mask mass, creating competition between pixels for the available budget. Increasing the activation of one region necessarily reduces the relative weight assigned to others. After clipping, pixels that exceed the available budget saturate at $M=1$, while less informative regions are driven towards $M=0$.

As a result, the optimisation tends to concentrate the available budget on a small subset of highly informative pixels, producing sparse masks without explicit binarisation mechanisms such as top-$k$ operators, Gumbel relaxations, straight-through estimators, or temperature annealing. In practice, the final masks contain only a narrow transition region between preserved and removed pixels, making them easy to interpret while remaining fully differentiable during optimisation.


\section{Experiments and Results}
\label{sec:experiments}

We evaluate \our{} along four dimensions: (1) flexibility across preservation targets, (2) faithfulness of the resulting masks, (3) cross-architecture generalisation, and (4) applicability beyond ImageNet on a private Retinopathy of Prematurity (ROP) dataset.
\begin{figure*}[ht!]

    \centering
    \includegraphics[width=\textwidth]{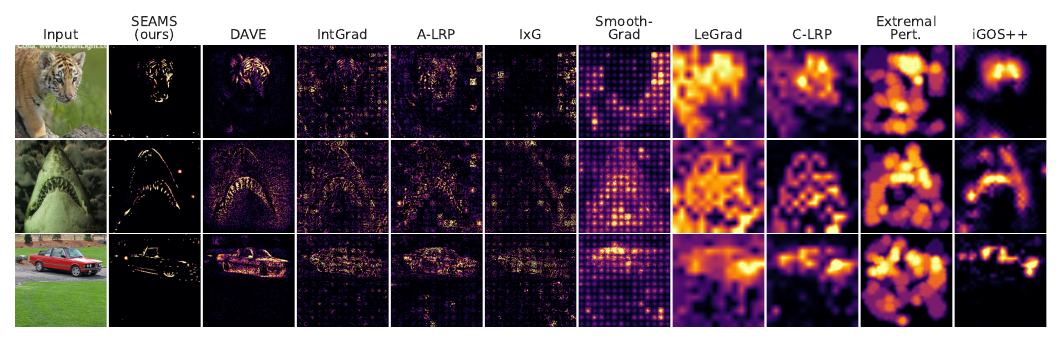}
    \caption{\textbf{Qualitative comparison of attribution maps.} For selected ImageNet-1K validation examples (rows), we display the input image (left) alongside attribution maps generated by \our{} (ours), DAVE, IntGrad, A-LRP, IxG, SmoothGrad,  LeGrad, C-LRP, Extremal Perturbations and IGOS++ (columns). Both DAVE and \our{} deliver sharper, more object-aligned, and spatially coherent explanations with significantly fewer patch-grid artifacts than prior techniques. Notably, compared to DAVE, \our{} produces sparser, binary activations that pinpoint only the most critical regions.}
    \label{fig:posthoc}
    \vspace{-0.3cm}
\end{figure*}

\smallskip
\noindent\textbf{Experimental Setup.}\, We evaluate \our{} on the ImageNet-1k validation set. To assess architectural generalization, we evaluate ConvNeXt-Tiny, ConvNeXt-Base, DINOv2 ViT-S/14, ViT-B/14, ViT-B/16, DeiT-B/16, and DeiT-III-B/16. All backbones use publicly available pretrained weights released by their respective authors. Input images are resized to $224 \times 224$ using bicubic interpolation (resize to 248 pixels followed by center cropping). Optimization is performed for $T=2000$ iterations with $D=4$ self-augmented distractors per iteration. Unless stated otherwise, all experiments use fixed hyperparameters $\lambda_{\alpha}=2.0$ and $\alpha_0=0.25$. To evaluate the applicability of the \our{} beyond natural images, we additionally conduct experiments on a private Retinopathy of Prematurity (ROP) dataset using a ConvNeXt-Small backbone. The optimization protocol and hyperparameters are identical to those used for ImageNet-1k. Details of the ROP dataset are provided in \cref{app:rop}.

\smallskip
\noindent\textbf{Qualitative comparisons.}\, \Cref{fig:posthoc} compares saliency maps produced by \our{} with several representative attribution methods, including C-LRP, LeGrad, SmoothGrad, AttnLRP, Integrated Gradients, DAVE, iGOS++, and Extremal Perturbations on ImageNet-1k validation examples. Visual differences reflect the conceptual distinction between sensitivity and sufficiency. Gradient-based methods identify pixels whose infinitesimal perturbation would affect the output, whereas \our{} directly optimises for the smallest subset of pixels that preserves it. Consequently, the resulting masks should not be interpreted as importance maps in the classical attribution sense, but rather as compact sufficient explanations of the model prediction. Additional qualitative examples are provided in \cref{app:qual}.

\smallskip
\noindent\textbf{Multi-target setting.} The optimisation objective of \our{} remains unchanged, while only the preserved output $g$ is modified. \Cref{fig:compare-g} illustrates this property on identical images with three targets: the global CLS embedding, predicted class probability, and full patch-token representation. These targets yield complementary explanations: CLS preservation produces compact object-level masks, class-probability optimisation highlights discriminative evidence, while token preservation distributes importance across the object to maintain local feature consistency. Results highlight the modularity of \our{}. Further results on cross-architecture saliency and the effect of backbone size across preservation targets are provided in \cref{app:b-models}.

We evaluate the resulting masks using insertion and deletion AUCs following~\cite{petsiuk2018rise}. As shown in~\cref{tab:ins-del}, \our{} consistently achieves a large insertion--deletion gap across targets. Class-probability optimisation performs best, reaching $0.822$ insertion and $0.257$ deletion AUC on ViT-S/16, while representation-based targets also retain substantial gaps. Similar results on ConvNeXt-T demonstrate that the formulation extends beyond transformer encoders. This experiment is an ablation of the optimisation target rather than a baseline comparison. Standard attribution baselines are not directly applicable to intermediate representations or uniformly across the two architectures.

\begin{table}[t]
\centering
\small
\begin{tabular}{@{}l@{\,}l@{\;\;}c@{\;\;}c@{\;\;}c@{}}
  \toprule
  Target & Backbone & ins-AUC $\uparrow$ & del-AUC $\downarrow$ & gap $\uparrow$ \\
  \midrule
  \multirow{2}{*}{$g\!=\!p_c$}
    & ViT-S/16     & $\mathbf{0.822}\!\pm\! 0.147$ & $0.257\!\pm\! 0.169$ & $\mathbf{0.565}$ \\
    & ConvNeXt-T   & $0.775\!\pm\! 0.093$ & $0.258\!\pm\! 0.136$ & $0.516$ \\
  \multirow{2}{*}{$g\!=\!\repr_{\text{CLS}}$}
    & ViT-S/16     & $0.814\!\pm\! 0.159$ & $0.267\!\pm\! 0.133$ & $0.547$ \\
    & ConvNeXt-T   & $0.778\!\pm\! 0.122$ & $0.270\!\pm\! 0.127$ & $0.508$ \\
  \multirow{2}{*}{$g\!=\!\repr_{\text{tok}}$}
    & ViT-S/16     & $0.789\!\pm\! 0.183$ & $0.284\!\pm\! 0.146$ & $0.505$ \\
    & ConvNeXt-T   & $0.772\!\pm\! 0.112$ & $0.285\!\pm\! 0.133$ & $0.487$ \\
  \bottomrule
\end{tabular}
\caption{\textbf{Insertion/deletion AUCs} averaged over randomly selected ImageNet validation set images. ins: AUC as top-ranked pixels are progressively revealed from a grey baseline (higher~$\uparrow$).  del: AUC as top-ranked pixels are progressively occluded (lower~$\downarrow$).  gap: ins$-$del (higher~$\uparrow$). Both backbones achieve a large gap on all three targets.}
\label{tab:ins-del}
\vspace{-0.6cm}
\end{table}

\begin{figure}[t]

  \centering
  \includegraphics[width=\linewidth]{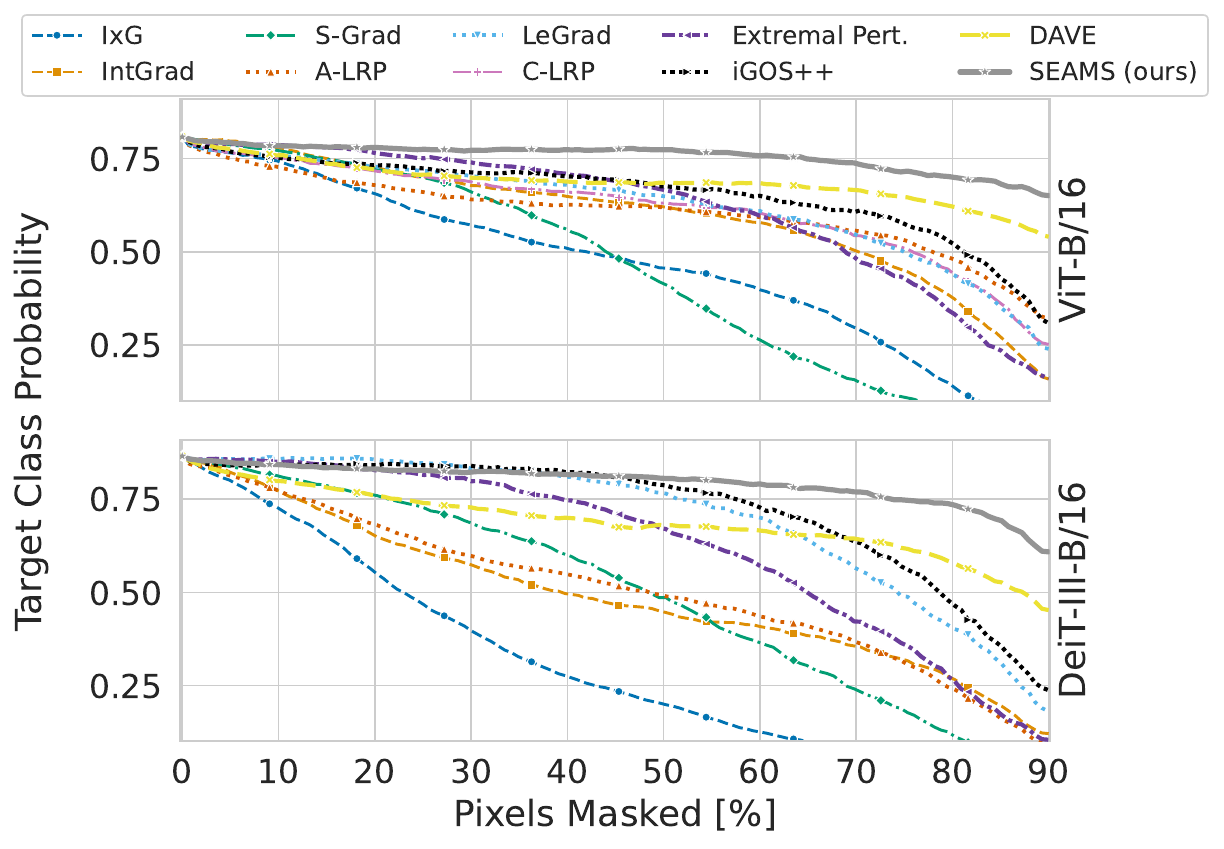}%
  \caption{{\textbf{Pixel-deletion faithfulness.}
Deletion curves for ViT-B/16 and DeiT-III-B/16. Higher curves indicate better faithfulness. \our{} (ours) consistently preserves the highest target-class probability across the deletion range on both architectures.}
  }
  \label{fig:pixdel-curves}
\end{figure}

\smallskip
\noindent\textbf{Faithfulness evaluation.} We evaluate whether \our{} identifies pixels that are informative for the model using standard pixel deletion~\cite{petsiuk2018rise} and ROAD~\cite{rong2022road}, which uses in-distribution imputation to reduce artifacts from constant-value masking. Following prior work~\citep{chefer2021transformer,wrobel2026dave}, for the standard pixel deletion protocol, we rank pixels by attribution score and progressively remove them from least to most relevant, replacing removed pixels with the channel-wise mean of the normalized input. We record the target-class probability after each step. Higher deletion curves indicate more faithful rankings. \Cref{fig:pixdel-curves} shows the curves and~\cref{tab:pixdel} reports the AUCs. \our{} achieves the highest score on both ViT-B/16 and DeiT-III-B/16, with AUCs of $68.1$ and $71.6$, respectively. The strongest competitors, DAVE ($61.9$) on ViT-B/16 and iGOS++ ($65.3$) on DeiT-III-B/16, fall short of \our{} by $6.2$ AUC points on both architectures. Paired statistical comparisons with Holm--Bonferroni correction are provided in \cref{app:stats}.

Constant-value masking can introduce distribution shifts~\cite{rong2022road}, we therefore additionally evaluate ROAD using noisy linear imputation. We consider both LeRF and MoRF removal orders and sufficiency at $10\%$ and $20\%$ pixel budgets. Full results are reported in \cref{app:faith}. \cref{tab:road-sufficiency-10} reports the more stringent $10\%$ sufficiency setting, using the same images and attribution maps as \cref{tab:pixdel}. \our{} achieves the highest target-class probability on both architectures, reaching $74.2\%$ on ViT-B/16 and $76.3\%$ on DeiT-III-B/16. This exceeds the strongest baseline, DAVE, by $17.8$ and $17.4$ percentage points, respectively. All differences are statistically significant after Holm--Bonferroni correction. Thus, \our{} concentrates substantial predictive evidence in only $10\%$ of the image pixels.

\begin{table}[t]
  \centering
  \small
  \begin{tabular}{@{}lcc@{}}
    \toprule
    Method & ViT-B/16 $\uparrow$ & DeiT-III-B/16 $\uparrow$ \\
    \midrule
    IxG & $41.7\!\pm\! 21.3$\, & $28.3\!\pm\! 19.0$\, \\
    SmoothGrad & $40.2\!\pm\! 13.3$\, & $45.1\!\pm\! 17.3$\,\\
    IntGrad & $54.0\!\pm\! 22.1$\, & $44.7\!\pm\! 25.2$\, \\
    A-LRP & $54.8\!\pm\! 22.5$\, & $46.1\!\pm\! 24.4$\, \\
    C-LRP & $55.7\!\pm\! 15.2$\, & --- \\
    LeGrad & $56.4\!\pm\! 14.6$\, & $63.4\!\pm\! 16.0$\, \\
    DAVE & $61.9\!\pm\! 19.8$\, & $62.1\!\pm\! 23.9$\, \\
     Extremal Pert. &  $55.7\!\pm\! 15.1$\, &  $56.5\!\pm\! 17.6$\, \\
     iGOS++ &  $59.3\!\pm\! 14.2$\,
    \, &  $65.3\!\pm\! 15.3$\, \\
    \midrule
    \rowcolor[HTML]{D5E4FF}
    \textbf{\our{} (ours)} & $\mathbf{68.1\!\pm\! 14.3}$\, & $\mathbf{71.6\!\pm\! 17.2}$\, \\
    \bottomrule
  \end{tabular}
  \caption{\textbf{Pixel-deletion faithfulness.}
Mean AUC~$\pm$~std integrated over $0\%$--$90\%$ pixel masking (higher is better). All methods differ significantly from \our{} after Holm--Bonferroni correction. Complete statistical results, including BCa 95\% confidence intervals, are provided in \cref{app:stats}. C-LRP is only supported for ViT-B/16.}
  \label{tab:pixdel}
  \vspace{-0.3cm}
\end{table}

\begin{table}[t]

  \centering
  \small
  \begin{tabular}{@{}lcc@{}}
    \toprule
    Method & ViT-B/16 $\uparrow$ & DeiT-III-B/16 $\uparrow$\\
    \midrule
    IxG & $27.5\!\pm\! 32.9$\,\, & $29.4\!\pm\! 36.1$\, \\
    SmoothGrad & $14.3\!\pm\! 26.0$\, & $13.5\!\pm\! 27.3$\, \\
    IntGrad & $51.0\!\pm\! 35.3$\, & $49.6\!\pm\! 39.4$\,\\
    A-LRP & $54.8\!\pm\! 34.2$\, & $50.6\!\pm\! 38.2$\, \\
    C-LRP & $30.8\!\pm\! 35.0$\, & --- \\
    LeGrad & $26.0\!\pm\! 34.0$\, & $27.3\!\pm\! 36.2$\, \\
    DAVE & $56.4\!\pm\! 34.3$\, & $58.9\!\pm\! 36.1$\, \\
    Extremal Pert. & $23.7\!\pm\! 32.4$\, & $26.2\!\pm\! 34.8$\, \\
    iGOS++ & $42.8\!\pm\! 36.1$\,& $45.1\!\pm\! 39.3$\,\\
    \midrule
    \rowcolor[HTML]{D5E4FF}
    \textbf{\our{} (ours)} & $\mathbf{74.2\!\pm\! 24.4}$\, & $\mathbf{76.3\!\pm\! 25.5}$\, \\
    \bottomrule
  \end{tabular}
  \caption{\textbf{Sufficiency under ROAD with top $\mathbf{10\%}$ kept.} Mean target-class probability (\%) when only the top $10\%$ of pixels are kept and everything else is imputed in-distribution. All methods differ significantly from \our{} after Holm–Bonferroni correction. Complete statistical results, including BCa 95\% confidence intervals, are provided in \cref{app:stats}.}
  \label{tab:road-sufficiency-10}
  \vspace{-0.4cm}
\end{table}

\begin{table}[t]
    \resizebox{\linewidth}{!}{%
    \begin{tabular}{@{}l@{\;}c@{\;\;}c@{\quad}c@{\quad}c@{}}
        \toprule
        Masks compared & Target &
        IoU@.5 & $\cos$ & $r$ \\
        \midrule

        \multirow{3}{*}{ ViT-S/16, 5 seeds}
        &  $g = p_c$
        &  $0.617\!\pm\! 0.090$
        &  $0.846\!\pm\! 0.065$
        &  $0.841\!\pm\! 0.064$ \\
        &  $g = f_{\mathrm{CLS}}$
        &  $0.619\!\pm\! 0.049$
        &  $0.890\!\pm\! 0.026$
        &  $0.874\!\pm\! 0.030$ \\
        &  $g = f_{\mathrm{tok}}$
        &  $0.555\!\pm\! 0.048$
        &  $0.863\!\pm\! 0.027$
        &  $0.847\!\pm\! 0.030$ \\

        \midrule

        \multirow{3}{*}{ ConvNeXt-Tiny, 5 seeds}
        &  $g = p_c$
        &  $0.601\!\pm\! 0.118$
        &  $0.846\!\pm\! 0.085$
        &  $0.836\!\pm\! 0.100$ \\
        &   $g = f_{\mathrm{CLS}}$
        &  $0.593\!\pm\! 0.051$
        &  $0.885\!\pm\! 0.029$
        &  $0.874\!\pm\! 0.032$ \\
        &  $g = f_{\mathrm{tok}}$
        &  $0.523\!\pm\! 0.054$
        &  $0.852\!\pm\! 0.040$
        &  $0.828\!\pm\! 0.044$ \\

        \midrule

        \multirow{3}{*}{ViT-S/16 vs ConvNeXt-Tiny}
        & $g = p_c$
        & $0.127\!\pm\! 0.057$
        & $0.282\!\pm\! 0.098$
        & $0.266\!\pm\! 0.102$ \\
        & $g = f_{\mathrm{CLS}}$
        & $0.128\!\pm\! 0.025$
        & $0.406\!\pm\! 0.043$
        & $0.333\!\pm\! 0.052$ \\
        & $g = f_{\mathrm{tok}}$
        & $0.116\!\pm\! 0.024$
        & $0.379\!\pm\! 0.045$
        & $0.295\!\pm\! 0.044$ \\

        \bottomrule
    \end{tabular}%
    }
    \centering
    \caption{\textbf{Mask agreement within and across architectures.} We measure agreement between masks using IoU@.5, cosine similarity ($\cos$), and Pearson correlation ($r$). For each architecture, values are averaged over pairs of masks obtained from 5 optimisation seeds. The final block reports agreement between ViT-S/16 and ConvNeXt-Tiny. Higher values indicate greater mask agreement.}
    \label{tab:mask-comparison}
\end{table}

\begin{figure}[h!]
\centering

\newcommand{\myfig}[1]{\includegraphics[width=\linewidth]{images/vit_convnext/#1}}

\newcommand{\salrow}[2]{%
    \par\noindent
    \begin{minipage}[c]{5px}
        \rotatebox{90}{\makebox[5px][c]{\tiny #1}}
    \end{minipage}%
    \begin{minipage}[c]{\dimexpr\columnwidth-5px\relax}
        \myfig{#2}
    \end{minipage}%
}

\newcommand{\salcol}[2]{%
    \begin{minipage}[c]{#1\columnwidth}
        \centering\tiny #2
    \end{minipage}%
}

\salcol{0.4}{ViT-S/16}
\salcol{0.13}{\phantom{Input}}
\salcol{0.39}{ConvNeXt-Tiny}
\par\noindent
\salcol{0.15}{\phantom{iii}Patch-Token}
\salcol{0.13}{CLS Embedding}
\salcol{0.13}{Class Probability}
\salcol{0.13}{Input}
\salcol{0.13}{Class Probability}
\salcol{0.13}{CLS Embedding}
\salcol{0.13}{Features}

\salrow{Ant}{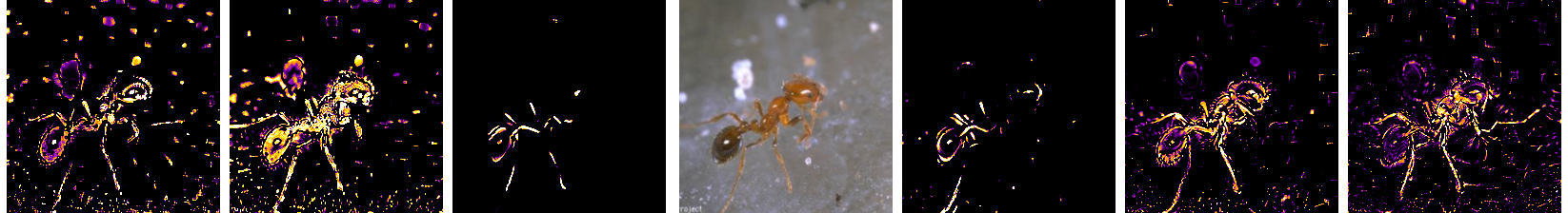}
\salrow{Sandbar}{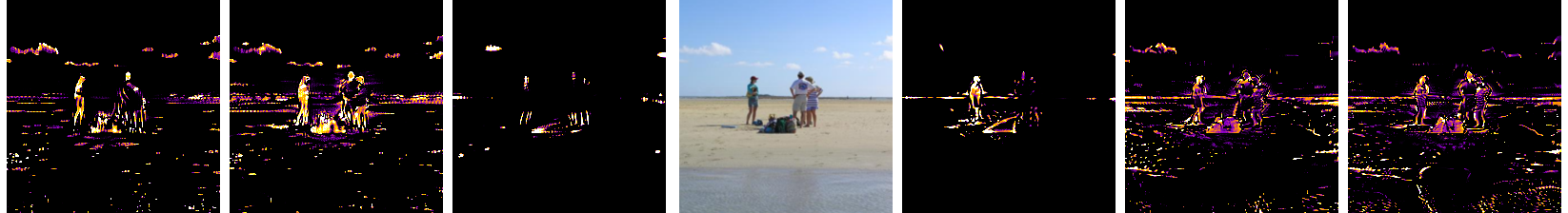}
\salrow{Hare}{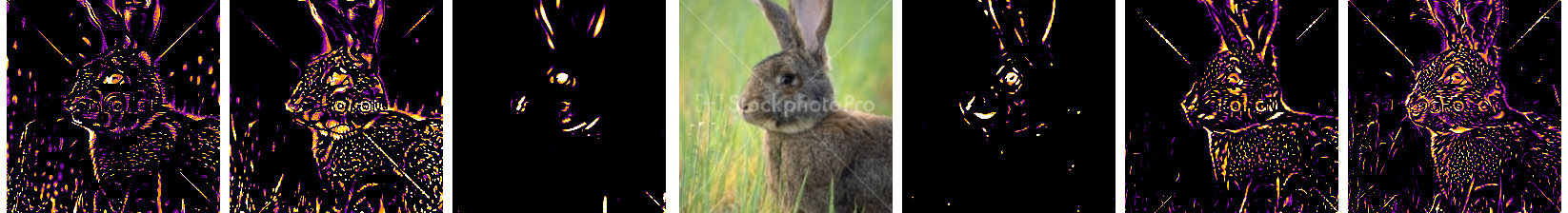}
\salrow{Garbage Truck}{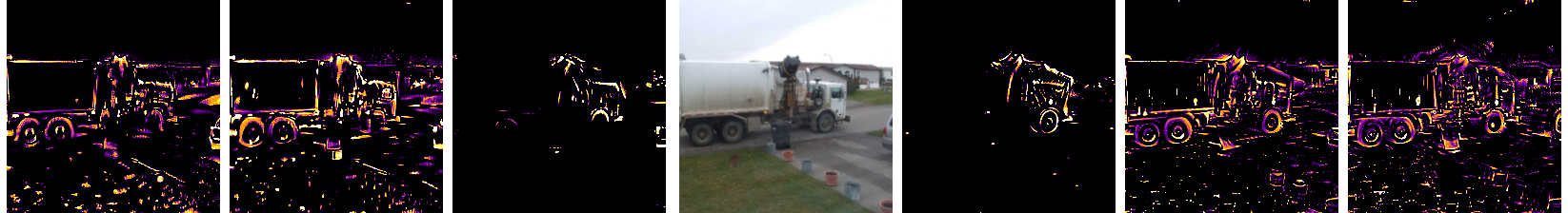}

\caption{\textbf{Cross-architecture saliency.}
Four ImageNet validation images with masks from ViT-S/16 (columns~2--4) and ConvNeXt-Tiny (columns~5--7) under $g\!=\!p_c$, $g\!=\!\repr_{\mathrm{CLS}}$, and $g\!=\!\repr_{\mathrm{tok}}$. Column~1 shows the originals. All
hyperparameters are shared across architectures except the encoder and input normalization. Quantitative results are in
\cref{tab:mask-comparison,tab:ins-del}.}
\label{fig:cross-arch}
\vspace{-0.3cm}
\end{figure}

\noindent\textbf{Cross-architecture generalisation.} Our optimisation is architecture-agnostic and requires only a differentiable forward pass through the preserved representation. To evaluate this claim, we compare the full optimisation between ViT-S/16 and ConvNeXt-Tiny without modifying any optimisation hyperparameters. \Cref{fig:cross-arch} shows that both backbones produce sparse, semantically meaningful explanations despite their substantially different representations. Quantitatively, cross-architecture agreement is low across all targets, with IoU@.5 below $0.15$ and Pearson correlation between $0.27$ and $0.33$, see \cref{tab:mask-comparison}. Importantly, this disagreement is not explained by optimisation variability. Repeating the optimisation with different seeds within the same architecture yields substantially higher agreement, with $r=0.83$--$0.87$. Thus, the masks are considerably more stable within an architecture than across architectures, indicating that the observed cross-architecture differences reflect the evidence selected by the backbones rather than optimisation noise.

Despite these differences in selected pixels, the resulting explanations achieve comparable insertion/deletion performance across architectures. This suggests that the two backbones can rely on substantially different visual evidence while supporting similarly faithful explanations. This form of architectural diversity is naturally exposed by sufficiency-based evaluation.

\begin{figure}[t]
    \centering
    \includegraphics[width=\linewidth,
    trim=0 0 8cm 0, clip]{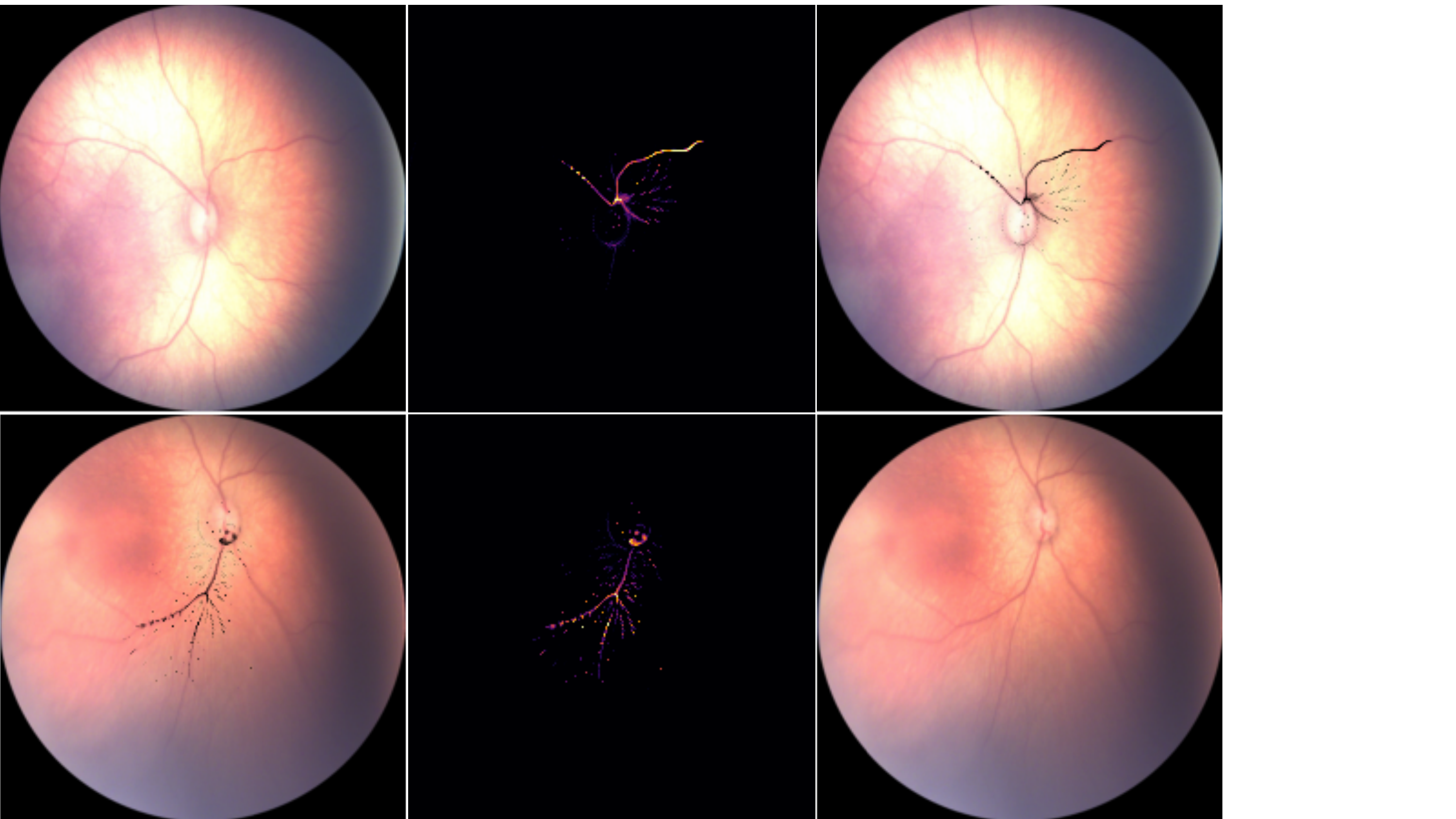}
    \caption{\textbf{Generalisation beyond ImageNet.}
Qualitative results on a private retinopathy of prematurity (ROP)
dataset. Columns left to right: original fundus image, soft mask,
and binarised mask overlaid on the original. Each row shows a
different patient. Without any modification to the optimisation
procedure or hyperparameters, \our{} produces sparse and clinically meaningful explanations, demonstrating transferability to a real-world medical imaging domain. Additional examples are provided in \cref{app:rop}.}
    \label{fig:medical}
    \vspace{-0.3cm}
\end{figure}

\smallskip
\noindent\textbf{Transfer to medical imaging.}\, Finally, we evaluate \our{} on a private clinical dataset for retinopathy of prematurity (ROP). Unlike ImageNet, this dataset consists of retinal fundus photographs
acquired in a real clinical environment. The optimisation pipeline and all hyperparameters remain unchanged. \Cref{fig:medical} demonstrates that \our{} successfully identifies clinically relevant retinal structures while maintaining sparse and spatially coherent explanations. Despite the substantial domain shift from natural images to medical imaging, the optimisation remains stable and produces semantically meaningful masks without any task-specific modifications. \our{} highlights key retinal image features associated with the development of severe stages of ROP. \our{} masks superimposed on fundus photographs emphasize vascular trajectory, tortuosity, and caliber. Additional mask components, visible as plume-like structures, are localized at the vascular–avascular junction (the demarcation line separating vascularized from avascular retina) or at the mesenchymal ridge, a pathological elevation that develops in the region of the demarcation line. These results indicate that \our{} transfers beyond standard computer vision benchmarks and can serve as a generic post-hoc explanation tool for specialised medical applications. Quantitative analysis on the ROP dataset is provided in \cref{app:rop}.

\section{Conclusions}

We introduced a sufficiency-based formulation of sharp saliency that identifies the minimal image evidence required to preserve a chosen model output, within the optimization composite.  By changing only the target function $g$, the same optimisation framework produces object-level, class-specific and
patch-token explanations without architecture-specific attribution rules or external distractor datasets. Experiments demonstrate compact and semantically meaningful masks together with competitive insertion/deletion performance, while revealing that different architectures rely on distinct visual evidence for the same prediction.

\smallskip
\noindent\textbf{Limitations and Future Work.}\, \our{} requires iterative optimization and is therefore considerably more computationally expensive than gradient-based attribution techniques. The resulting masks also depend on the chosen image composition strategy and should be interpreted as regions sufficient for preserving a model output together with the optimization composite. While \our{} is model-agnostic in principle, the current evaluation focuses exclusively on image encoders. Extending preservation optimization to video, multimodal, or generative architectures remains an interesting direction for future work.

\section*{Acknowledgments}
The work of Jacek Tabor was supported by the Foundation for Polish Science (FNP) grant 'Centre for Credible AI' No. FENG.02.01-IP.05-0058/24. The work of Łukasz Struski was supported by the National Science Centre, Poland, grants no. 2023/49/B/ST6/01137. Some experiments were performed on servers purchased with funds from the flagship project entitled ``Artificial Intelligence Computing Center Core Facility'' from the DigiWorld Priority Research Area within the Excellence Initiative -- Research University program at Jagiellonian University in Kraków. We also gratefully acknowledge Polish high-performance computing infrastructure PLGrid (HPC Center: ACK Cyfronet AGH) for providing computer facilities and support within computational grant no. PLG/2026/019974.

{
    \small
    \bibliographystyle{ieeenat_fullname}
    \bibliography{main}
}

\clearpage
\appendix

\section{Optimisation details}
\label{sec:optimisation}

For a single query image~$\img$, the learnable parameters are the logit tensor $\bm{\ell}\in\R^{H\times W}$ (initialised to zero) and the budget scalar $\alpha_{\text{raw}}$ (initialised so that $\alpha=0.25$).  We use Adam with learning rates
$\eta_{\bm{\ell}}=0.05$ for the mask logits and $\eta_\alpha=0.01$ for the budget, running for $T=2000$ steps, unless stated otherwise.  At each step we (i) recompute
$\alpha$ and $M,M'',N$ via normalise-and-clip; (ii) sample $D=4$ fresh synthetic distractors $\img_{dist}=\mathrm{heavyAug}(\img)$ and a fresh blur $\sigma_d$; (iii) form $D$ composites and average the preservation MSE; (iv) add the sparsity regulariser
$\lambda_\alpha\,\alpha$; and (v) back-propagate.  The one regulariser weight is $\lambda_\alpha=2.0$, kept fixed for every encoder and every target~$g$. \Cref{alg:main} states the procedure as pseudocode.

\vspace{0.6cm}
\subsection{Cost of the optimisation}

In batch mode ($N$ images optimised simultaneously) one batched forward pass replaces $N$ sequential ones. The wall-clock cost is ${\sim}40$\,s per image on a single RTX~5060 GPU, for the $T{=}2000$ configuration used throughout this appendix. The faithfulness benchmark of the main text uses the cheaper $T{=}500$, $n_{\mathrm{copies}}{=}3$ setting, measured at $49.4$\,s per image on an RTX~3080.

The budget of the benchmark was chosen to match the optimisation-based baselines, not to be minimal. Extremal Perturbations uses the same 6000 calls ($40.4$\,s) and iGOS++ $0.92\times$ that ($21.7$\,s). Reducing the optimisation to 50 steps costs $4.9$\,s per image and still beats every baseline on all ROAD faithfulness protocols, reaching $79.6$ on ROAD sufficiency at $10\%$, see \cref{tab:cost-quality}.

Therefore the cost is a tunable trade-off rather than a fixed price. Sufficiency in fact falls as the budget grows $79.6$ at $T=50$ against $73.1$ at $T=500$. A longer optimisation makes the mask sparser, and a sparser mask ranks fewer pixels, which is what a fixed-budget retention protocol reads.

\begin{table}[h!]
  \centering
  \small
  \setlength{\tabcolsep}{4pt}
  \resizebox{\columnwidth}{!}{%
  \begin{tabular}{@{}lrrccc@{}}
    \toprule
    Budget & calls & s/img & DAUC $\uparrow$ & ROAD @$10\%$ $\uparrow$ & ROAD @$20\%$ $\uparrow$ \\
    \midrule
    $T=50$ & 600 & $4.9$ & $64.9$ & $79.6$ & $83.4$ \\
    $T=100$ & 1\,200 & $9.9$ & $66.5$ & $79.7$ & $83.9$ \\
    $T=200$ & 2\,400 & $20.0$ & $68.5$ & $79.6$ & $82.7$ \\
    $T=500$ \emph{(paper)} & 6\,000 & $49.4$ & $68.3$ & $73.1$ & $80.1$ \\
    \midrule
    DAVE & --- & $1.1$ & $61.6$ & $60.9$ & $73.3$ \\
    iGOS++ & --- & $21.6$ & $59.3$ & $40.1$ & $68.6$ \\
    IntGrad & --- & $1.2$ & $55.3$ & $56.0$ & $76.4$ \\
    \bottomrule
  \end{tabular}}
  \caption{\textbf{Cost--quality trade-off of the optimization.} \our{} optimization with reduced numbers of steps on the first 100 images of the faithfulness evaluation subset. Each configuration is evaluated independently on a single GPU. ``Calls'' denotes the total number of forward--backward passes through the frozen backbone, computed as $T\times4$ distractor draws $\times3$ mask copies. The results quantify the trade-off between optimization cost and attribution faithfulness.}
  \label{tab:cost-quality}
  \vspace{-0.3cm}
\end{table}
\section{Larger backbones: ViT-B/16 and ConvNeXt-Base}
\label{app:b-models}

\begin{figure}[t]
\centering
\includegraphics[width=\columnwidth]{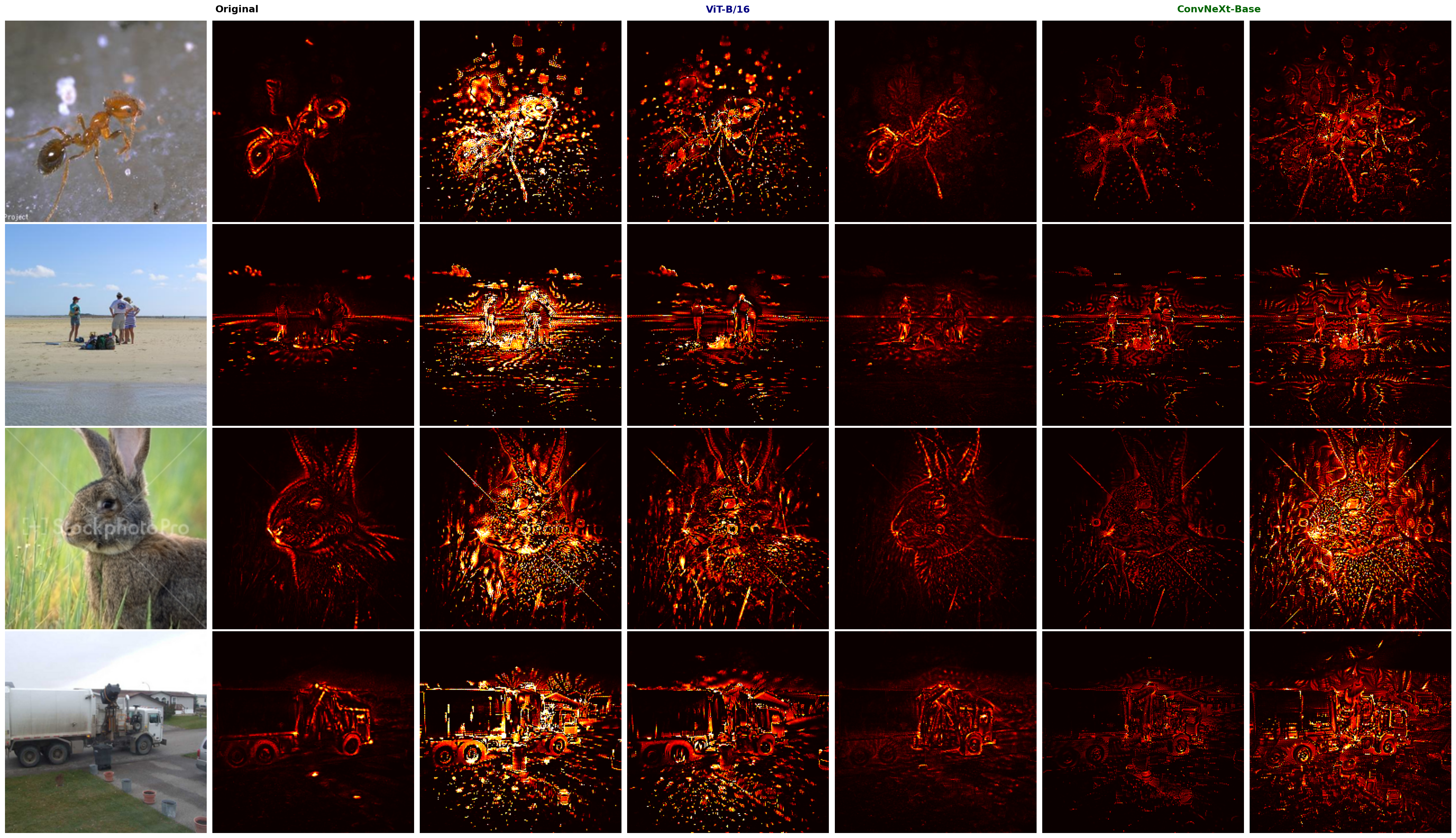}
\caption{\textbf{Cross-architecture saliency on the larger backbones.} Each row is one ImageNet val image; columns~1 the original, columns~2--4 ViT-B/16 masks (class-prob-mse, cls-embed, patch-mse), columns~5--7 ConvNeXt-Base masks (same three targets). Qualitative behaviour matches the smaller backbones. Quantitative agreement is reported in \cref{tab:s-vs-b}.}
\label{fig:cross-arch-b}
\end{figure}

The main text uses ViT-S/16 and ConvNeXt-Tiny, which both have ${\sim}28$M parameters, as the two backbones in the cross-architecture analysis.  We additionally repeated the same protocol on the next-size-up models, namely ViT-B/16 ($86.6$M, $768$-d (CLS representation) and ConvNeXt-Base ($88.6$M, $1024$ d pooled representation). For both architectures we use standard timm checkpoints.  All other hyperparameters were unchanged. \Cref{fig:cross-arch-b} shows qualitative results of this experiment and~\cref{tab:s-vs-b} summarises the per-target means.
\begin{algorithm}[th]
\caption{\our{} optimisation for one query image.}
\label{alg:main}
\begin{algorithmic}[1]
\REQUIRE Query image $\img$, frozen encoder $\repr$, output to preserve
         $g(\cdot)$, initial budget $\alpha_0$, steps $T$,
         sparsity weight $\lambda_\alpha$
\STATE $\bm{\ell}\gets\bm{0}\in\R^{H\times W}$;\;
       $\alpha_{\text{raw}}\gets\sigmoid^{-1}(\alpha_0)$
\STATE $\bm{y}^{\star}\gets g(\img)$ \hfill\COMMENT{fixed target output}
\FOR{$t=1,\dots,T$}
  \STATE $\alpha\gets\sigmoid(\alpha_{\text{raw}})$
  \STATE $\bm{u}\gets\operatorname{softplus}(\bm{\ell})$;\;
  $M\gets\operatorname{clip}(\alpha\cdot\bm{u}/\operatorname{mean}(\bm{u}),0,1)$
  \STATE $M''\gets\min(1-M,\alpha)$;\; $N\gets 1-M-M''$
  \STATE $\mathcal{L}\gets 0$
  \FOR{$d=1,\dots,D$}
    \STATE $\img_{dist}\gets\mathrm{heavyAug}(\img)$;\;
           sample $\sigma_d\sim\mathrm{U}[15,25]$;\;
           $\img_{\text{blur},d}\gets\mathrm{blur}_{\sigma_d}(\img)$
    \STATE $\tilde{\img}_d\gets
             M\odot\img+M''\odot\img_{dist}+N\odot\img_{\text{blur},d}$
    \STATE sample $\augm$;\;
           $\mathcal{L}\gets\mathcal{L}+\lVert g(\augm(\tilde{\img}_d))-\bm{y}^{\star}\rVert^2$
  \ENDFOR
  \STATE $\mathcal{L}\gets\mathcal{L}/D
           + \lambda_{\alpha}\,\alpha$
  \STATE $(\bm{\ell},\alpha_{\text{raw}})\gets
         \text{Adam-step}\bigl((\bm{\ell},\alpha_{\text{raw}}),
         \nabla\mathcal{L}\bigr)$
\ENDFOR
\RETURN $M$, $\alpha$
\end{algorithmic}
\end{algorithm}

It is worth noticing, that the masks remain visually meaningful on both larger backbones and the qualitative cross-architecture pattern (different pixels, comparable fidelity) is preserved.  Quantitatively the picture is a bit subtler. For the holistic CLS target, ViT-B uses a larger budget than ViT-S to
reach a similar cosine ($\bar{\alpha}\!=\!0.113$ vs.\ $0.086$,
$\overline{\cos}_{\text{blur}}\!=\!0.914$ vs.\ $0.918$), consistent with the harder constraint of preserving a $768$-d embedding instead of a $384$-d one. ConvNeXt-Base, in contrast, is somewhat more efficient than ConvNeXt-Tiny
($\bar{\alpha}\!=\!0.020$ vs.\ $0.028$ at the same $\overline{\cos}$). For the one-dimensional class-probability target, both larger backbones produce a slightly higher $p_{\text{masked}}$ on the blur composite than their smaller counterparts.  Differences are small ($\sim\!0.02$--$0.05$) and do not change the main story.

\section{Retinopathy of Prematurity}
\label{app:rop}
Retinopathy of prematurity (ROP) is an ocular disease affecting preterm infants. The
underlying pathology involves abnormal vascular proliferation and other pathological changes
occurring in the immature retina. These processes may ultimately lead to tractional retinal
detachment and blindness. ROP is currently recognized as the leading cause of vision loss in
newborns, with recent estimates indicating that at least 50,000 children worldwide are blind as
a consequence of the disease. Furthermore, ROP remains the primary cause of visual
impairment among children under five years of age in developed countries.

\begin{table}[t]
\centering
\begin{tabular}{@{}l@{\;\;}lcc@{}}
  \toprule
  Target & Backbone & $\bar{\alpha}$ &
  $\overline{\cos}_{\text{blur}}$ \,or\, $\bar{p}_{\text{masked,blur}}$ \\
  \midrule
  \multirow{2}{*}{$g\!=\!\repr_{\text{CLS}}$}
    & ViT-S/16     & 0.086 & 0.918 \\
    & ViT-B/16     & 0.113 & 0.914 \\
  \cmidrule(l{-2pt}r{15pt}){2-4}
    & ConvNeXt-T   & 0.028 & 0.907 \\
    & ConvNeXt-B   & 0.020 & 0.900 \\
  \midrule
  \multirow{2}{*}{$g\!=\!p_c$}
    & ViT-S/16     & 0.017 & 0.934 \\
    & ViT-B/16     & 0.015 & 0.960 \\
  \cmidrule(l{-2pt}r{15pt}){2-4}
    & ConvNeXt-T   & 0.012 & 0.846 \\
    & ConvNeXt-B   & 0.010 & 0.898 \\
  \midrule
  \multirow{2}{*}{$g\!=\!\repr_{\text{tok}}$}
    & ViT-S/16     & 0.063 & 0.689 \\
    & ViT-B/16     & 0.077 & 0.733 \\
  \cmidrule(l{-2pt}r{15pt}){2-4}
    & ConvNeXt-T   & 0.069 & 0.842 \\
    & ConvNeXt-B   & 0.085 & 0.823 \\
  \bottomrule
\end{tabular}
\caption{\textbf{Effect of backbone size across preservation targets.}
Means over the $10$ standard images for small (S) and base (B) backbones.
$\bar{\alpha}$ denotes the realised mask budget, while
$\overline{\cos}_{\mathrm{blur}}$ and $\bar{p}_{\mathrm{masked,blur}}$ denote,
respectively, the cosine similarity between the encoder representations of the blur-filled composite and the original image, and the corresponding target-class probability. The larger backbones achieve comparable preservation, with only modest changes in the realised mask budget.}
\label{tab:s-vs-b}
\vspace{-0.6cm}
\end{table}

\vspace{-0.3cm}
\paragraph{Private RetCam Neonatal Fundus Dataset.} The dataset consists of anonymized retinal fundus images collected retrospectively at the Department of Neonatal Pathology and Intensive Care, University Children's Hospital in Kraków, Poland. All images were acquired using a RetCam wide-field retinal imaging system during routine ophthalmological examinations performed between 2013 and 2023.
\begin{figure*}[th!]
    \centering
    \includegraphics[width=\textwidth,
    trim=0 15cm 6.5cm 0, clip]{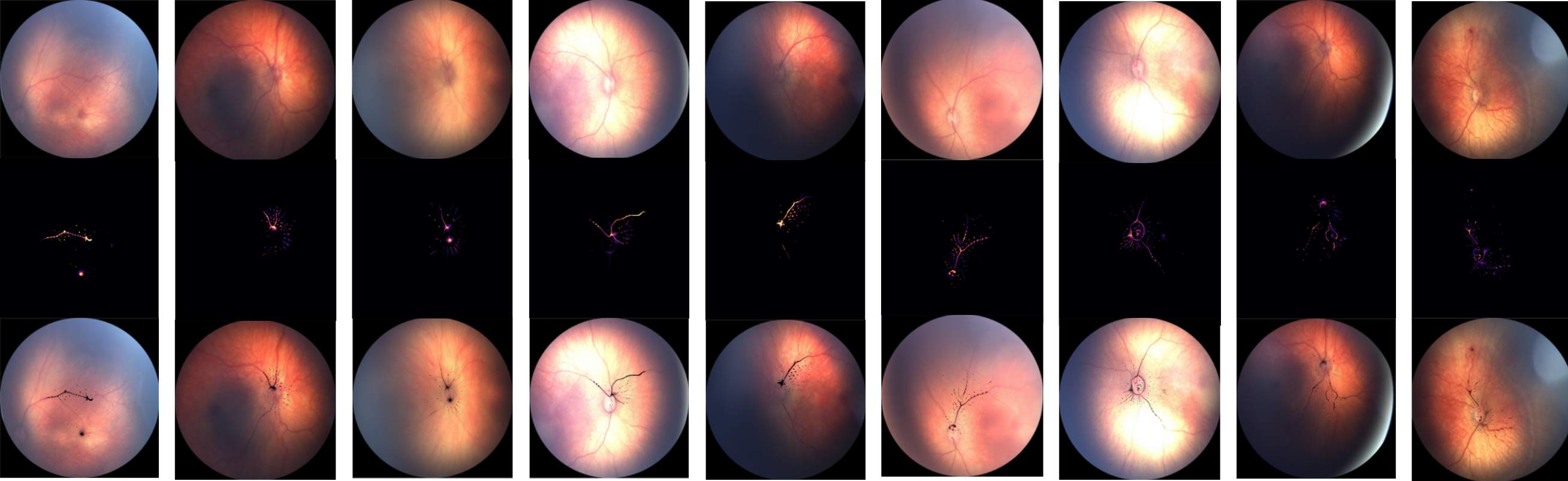}
    \caption{\textbf{Generalisation beyond ImageNet.}
Qualitative results on a private retinopathy of prematurity (ROP)
dataset. Without any changes to the optimisation procedure or
hyperparameters, \our{} produces sparse and clinically
meaningful explanations on retinal fundus images, demonstrating its
transferability to a real-world medical imaging domain.}
    \label{fig:medical-supp}
\end{figure*}
The study population includes neonates of both sexes who were hospitalized at the Department of Neonatal Pathology and Intensive Care and underwent retinal fundus imaging as part of their clinical care. Inclusion criteria comprised hospitalization in the department during the study period and availability of at least one RetCam examination. Patients with incomplete imaging records or insufficient image quality preventing clinical assessment were excluded from the dataset.

The dataset contains examinations from 243 unique patients and comprises 970 retinal imaging sessions, resulting in a total of 18690 retinal fundus photographs. All images were anonymized prior to inclusion in the dataset, and no personally identifiable patient information is available.

\vspace{0.2cm}
This dataset is not publicly available due to patient privacy regulations and institutional data protection policies. Access to the data is restricted and subject to approval by the data-owning institution and relevant ethics committees.

\vspace{-0.3cm}
\paragraph{Risk factors for ROP.}
The principal risk factors for ROP are prematurity-related immaturity, particularly gestational age and birth weight. Infants born before 28 weeks of gestation constitute a population at especially high risk. The global incidence of preterm birth continues to increase and varies considerably across regions, a pattern mirrored by the incidence of ROP. Over the past two decades, substantial advances in neonatal intensive care have significantly improved the
survival of preterm infants. At the same time, the number of children at risk of developing
severe forms of ROP has increased. Since the introduction of routine ophthalmic screening recommendations for preterm infants
in 2006, children at risk of irreversible vision loss have been eligible for timely treatment,
including laser photocoagulation of the avascular retina and intravitreal administration of anti-
VEGF (vascular endothelial growth factor) agents. These interventions have contributed to a
global reduction in blindness associated with prematurity. Current efforts by neonatologists
and ophthalmologists focus on optimizing screening protocols and treatment strategies to
effectively modify the course of this vision-threatening disease.

\vspace{-0.3cm}
\paragraph{ROP screening.} ROP screening is based on binocular indirect ophthalmoscopic examination performed by an ophthalmologist experienced in neonatal ocular diseases following pharmacological pupil dilation. The examination is technically demanding and requires the use of an eyelid speculum, a binocular indirect ophthalmoscope, a condensing lens, and a sterile instrument facilitating visualization of peripheral retinal abnormalities, such as a strabismus hook or scleral depressor. During the examination, the ophthalmologist evaluates multiple clinically relevant features, including vascular course and caliber, the extent and boundary of retinal vascularization, and the presence of other characteristic pathological findings. The retina is divided into anatomical zones, and observed abnormalities should be classified according to the current International Classification of Retinopathy of Prematurity (ICROP). Examinations are performed at predefined intervals and repeated as necessary, as assessment of disease dynamics is critical for appropriate management. In recent years, the RetCam imaging system (Clarity Medical Systems Inc.) has become a valuable adjunct to clinical examination. This wide-field digital retinal imaging device enables photographic documentation of fundus findings. It facilitates longitudinal comparison of retinal images during disease monitoring, supports treatment decision-making, and assists in evaluating therapeutic outcomes. Its utility is particularly evident in settings where access
to ophthalmologists experienced in neonatal retinal diseases is limited, enabling telemedicine-based consultation and remote assessment of retinal abnormalities.

Researchers have increasingly investigated RetCam-derived images to develop tools supporting the early detection of retinal changes associated with ROP, particularly those predictive of severe disease progression. Evidence suggests that artificial intelligence and neural network–based approaches can enhance ROP assessment but are not capable of replacing ophthalmologists in clinical practice. Clinical decision-making based solely on such models remains insufficient due to the complexity of the ROP phenotype and the inherent variability of disease progression. Nevertheless, given the substantial burden of vision-
threatening disability associated with ROP, research continues to focus on developing optimal decision-support solutions for specialists in neonatal ophthalmology, whose availability remains limited.
\vspace{-0.3cm}
\paragraph{Additional Examples.}

\our{} highlights key retinal image features associated with the development of severe stages of ROP. \our{} masks superimposed on fundus photographs emphasize vascular trajectory, tortuosity, and
caliber. Additional mask components, visible as plume-like structures, are localized at the vascular–avascular junction (the demarcation line separating vascularized from avascular retina) or at the mesenchymal ridge, a pathological elevation that develops in the region of the
demarcation line.
\vspace{-0.3cm}
\paragraph{Quantitative Results.}

For the medical dataset, we consider expert assessment particularly important. Standard XAI faithfulness metrics evaluate a specific computational property of an attribution map, but do not directly establish whether the highlighted regions correspond to clinically meaningful structures or lesions. In this setting, the central question is whether the model focuses on anatomically and clinically relevant evidence.

Importantly, we do not intend qualitative assessment to replace quantitative faithfulness evaluation. Rather, the two evaluations address complementary questions. Quantitative faithfulness measures whether the attribution identifies information that is important for the model's prediction, whereas expert inspection assesses whether this information is clinically meaningful.

To complement the qualitative evaluation on the private retinopathy of prematurity (ROP) dataset, we additionally evaluate attribution faithfulness using a pixel-deletion protocol on ConvNeXt-Small. As shown in \cref{tab:pixdel_convnext_small}, \our{} achieves the highest AUC ($65.89$), outperforming all gradient-based baselines. The improvement over SmoothGrad is not statistically significant, whereas the differences relative to the remaining baselines are significant under the paired Wilcoxon signed-rank test. These results provide quantitative evidence that the regions identified by \our{} remain more informative under progressive deletion.

\section{Augmentation Details}
\label{app:aug}
We employ two distinct augmentation pipelines, described in the main text. The first, \texttt{HeavyAug}, is used to generate synthetic distractors for the three-way image decomposition described in the main text. The second is a lighter differentiable augmentation applied after mask composition to improve robustness of the learned saliency masks.
\subsection{Synthetic Distractor Augmentation (HeavyAug)}
\label{app:heavyaug}

\begin{table}[h!]
\centering
\small
  \setlength{\tabcolsep}{4pt}
  \resizebox{\columnwidth}{!}{%
\begin{tabular}{lccc}
\toprule
Method & AUC $\uparrow$ & 95\% CI & $\Delta$ vs.\ SEAMS \\
\midrule
\textbf{SEAMS (ours)} & \textbf{65.89} & \textbf{[62.69, 69.08]} &  -- \\
SmoothGrad & 62.79 & [59.26, 66.40] &  -3.10 (ns) \\
attnlrp & 62.20 & [58.76, 65.70] & -3.70 (*) \\
ig & 61.44 & [58.14, 64.79] &  -4.45 (*) \\
IxG & 60.88 & [57.63, 64.16] & -5.01 (*) \\
vanilla grad & 58.58 & [55.53, 61.66] & -7.31 (*) \\
\bottomrule
\end{tabular}}
\caption{\textbf{Deletion faithfulness on ConvNeXt-Small.} Results on the private Retinopathy of Prematurity (ROP) dataset. Pixels are removed from least to most salient according to each attribution map; higher AUC indicates better
faithfulness. Brackets denote 95\% bootstrap confidence intervals from $10^{4}$ resamples of the per-image AUCs. $\Delta$ is the mean paired difference relative to \textbf{SEAMS}; significance is assessed with a paired Wilcoxon signed-rank test ($*$: $p<0.05$; ns: not significant).}
\label{tab:pixdel_convnext_small}
\vspace{-0.3cm}
\end{table}
The synthetic distractor image $I_{\mathrm{dist}}$ is generated using a deliberately aggressive augmentation pipeline designed to produce representations that are substantially different from those of the original image while remaining image-derived.
The following transformations are applied sequentially:
\begin{itemize}
\item Random affine transformation:
\begin{itemize}
\item rotation sampled from $[0^\circ,360^\circ]$,
\item scale sampled from $[0.4,1.6]$,
\item translation sampled from $[-0.3,0.3]$ of image size.
\end{itemize}
\item Horizontal flip ($p=0.5$).

\item Vertical flip ($p=0.5$).

\item Radial swirl transformation
\[
    \theta \mapsto \theta + s \exp(-r^2/\sigma^2),
\]
where the swirl strength is sampled as
$s \sim \mathcal{U}[-3,3]$.

\item Strong color perturbation:
\begin{itemize}
    \item brightness shift $\pm 0.4$,
    \item contrast factor sampled from $\mathcal{U}[0.5,1.5]$,
    \item saturation factor sampled from $\mathcal{U}[0.3,1.7]$,
    \item random RGB channel permutation with probability $0.5$.
\end{itemize}

\item Posterization (2--3 bits) or solarization.

\item Cutout augmentation using 1--3 randomly positioned rectangles, each covering up to approximately one-third of the image and filled with random gray values.
\end{itemize}

All transformations are implemented using GPU tensor operations. New augmentation parameters are sampled independently at every optimization step, ensuring that the model never observes the same synthetic distractor twice.

\subsection{Post-Mask Augmentation}
\label{app:postmaskaug}

After constructing the masked composite image, we apply a differentiable augmentation pipeline to improve robustness and discourage solutions that exploit low-frequency artifacts introduced by the blur channel.

\paragraph{Geometric transformations.}

A single affine transformation jointly models rotation, scaling, and translation. At each optimization step, a possibly rotated sub-window of the composite image is selected and resampled back to $224\times224$ resolution using bilinear interpolation with reflection padding.

The transformation parameters are sampled as follows:

\begin{itemize}
\item rotation angle in the range $\pm25^\circ$,
\item inverse-grid scale factor sampled from $[1.0,2.0]$,
\item translation up to $\pm0.15$ in normalized image coordinates.
\end{itemize}

Under this parameterization, the effective crop side length lies in the interval $[0.5,1.0]$ of the original image dimensions, corresponding to crop areas between approximately 25\% and 100\% of the image area.

\paragraph{Photometric transformations.}

The geometric transform is followed by a sequence of photometric augmentations:

\begin{itemize}
\item Gaussian blur with probability $0.5$, using
$\sigma \in [0.1,1.5]$ and a $7\times7$ depthwise convolution kernel;

\item brightness jitter of $\pm0.25$;

\item contrast jitter with multiplicative factor sampled from
$\mathcal{U}[0.75,1.25]$ and applied with probability $0.7$;

\item saturation jitter with multiplicative factor sampled from
$\mathcal{U}[0.6,1.4]$ and applied with probability $0.4$;

\item horizontal flip with probability $0.5$.

\end{itemize}

The reference target $y^\star = g(I)$ is computed once from the unaugmented image and does not undergo any of the above transformations. Consequently, the optimization is encouraged to identify image regions that preserve the selected model output under both geometric and photometric perturbations.

\section{Sufficiency vs.\ Sensitivity}
\label{sec:multi-instance}

\subsection{Multi-Instance Scenes}
Attribution methods based on local sensitivity, such as DAVE, assign importance to every image region whose perturbation would affect the model output. In scenes containing multiple instances of the same object class: a basket of strawberries, a row of monitors or a cluster of apples this results in attribution maps that highlight all visible instances simultaneously, since each one individually influences the prediction.

\our{} operates under a fundamentally different objective. Rather than identifying all influential regions, it searches for the smallest subset of pixels sufficient to preserve the model output. In
multi-instance scenes this distinction becomes particularly visible. Because a single instance of the target object already provides enough
evidence for the model to maintain its prediction, the sparsity constraint drives the optimised mask to concentrate on one or a small subset of the available instances, leaving the remaining occurrences uncovered.

\begin{figure*}[thbp]
\centering
\begin{minipage}[t]{0.49\linewidth}
\centering
\begin{minipage}[c]{0.33\linewidth}\centering\small Original\end{minipage}%
\begin{minipage}[c]{0.33\linewidth}\centering\small \our{}\end{minipage}%
\begin{minipage}[c]{0.33\linewidth}\centering\small DAVE\end{minipage}

\includegraphics[width=\linewidth]{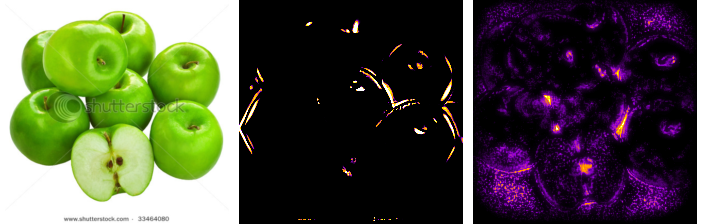}\\
\includegraphics[width=\linewidth]{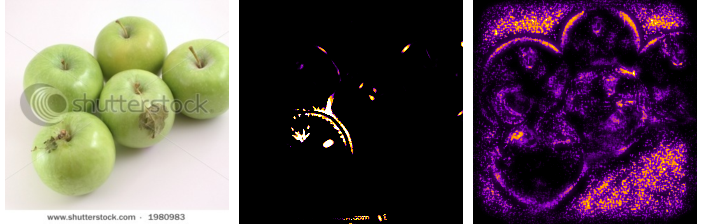}\\
\includegraphics[width=\linewidth]{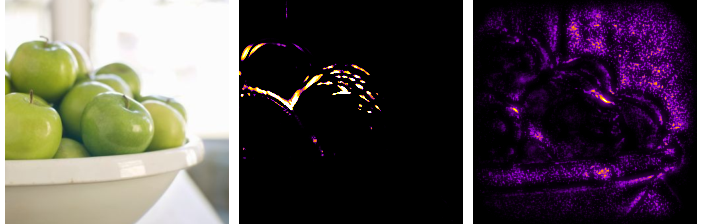}\\
\includegraphics[width=\linewidth]{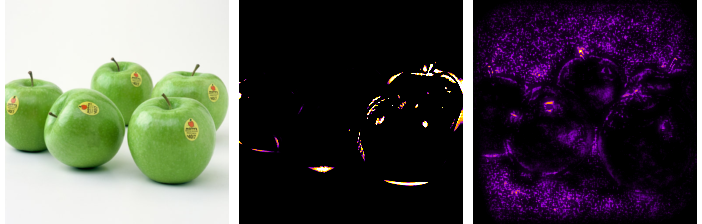}\\
\includegraphics[width=\linewidth]{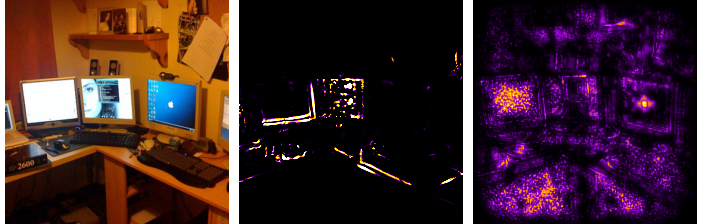}\\
\includegraphics[width=\linewidth]{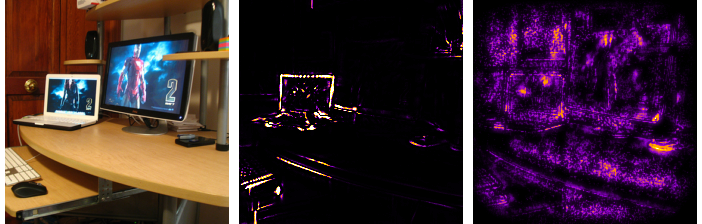}\\
\includegraphics[width=\linewidth]{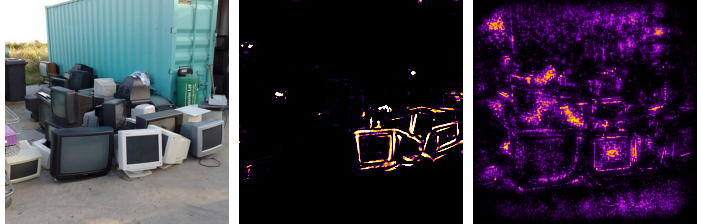}
\end{minipage}%
\hfill%
\begin{minipage}[t]{0.49\linewidth}
\centering
\begin{minipage}[c]{0.33\linewidth}\centering\small Original\end{minipage}%
\begin{minipage}[c]{0.33\linewidth}\centering\small \our{}\end{minipage}%
\begin{minipage}[c]{0.33\linewidth}\centering\small DAVE\end{minipage}

\includegraphics[width=\linewidth]{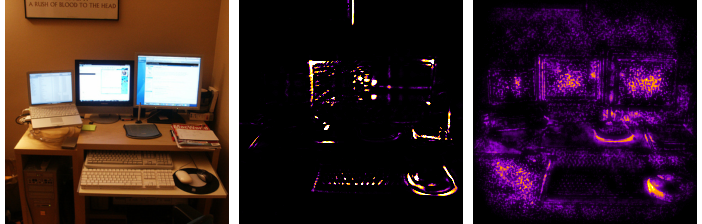}\\
\includegraphics[width=\linewidth]{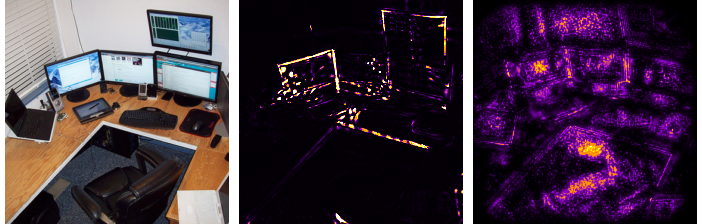}\\

\includegraphics[width=\linewidth]{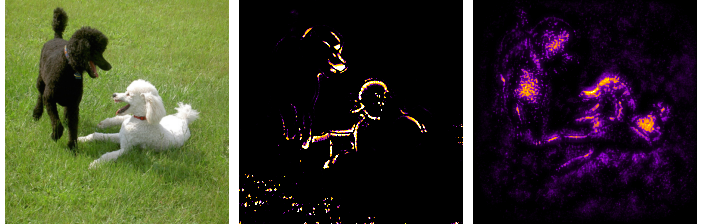}\\
\includegraphics[width=\linewidth]{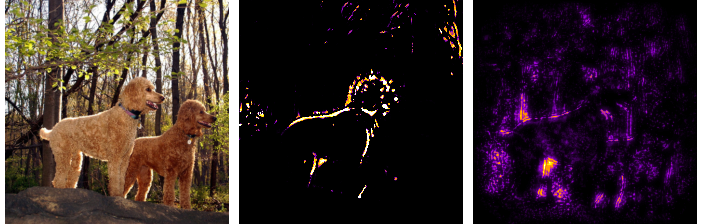}\\
\includegraphics[width=\linewidth]{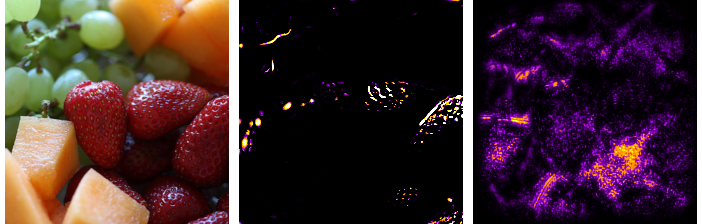}\\
\includegraphics[width=\linewidth]{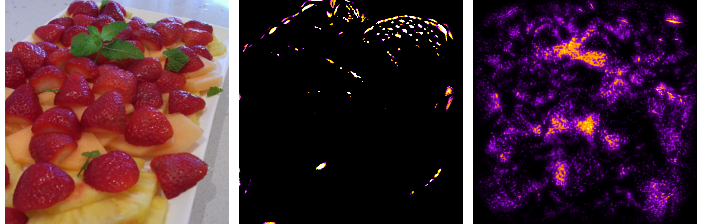}\\
\includegraphics[width=\linewidth]{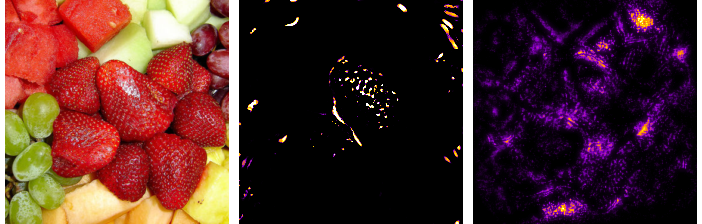}
\end{minipage}
\caption{\textbf{Multi-instance scenes: sufficiency vs.\ sensitivity.}
Each row shows an ImageNet validation image containing multiple
instances of the same object class (left), the \our{} soft mask
(centre), and the DAVE attribution map (right).
Sensitivity-based methods such as DAVE highlight all visible
instances, since each one independently influences the model output.
\our{} instead identifies the minimal sufficient evidence: because a
single instance already preserves the model prediction, the sparsity
constraint concentrates the mask on a small subset of the available
objects. This behaviour is a direct consequence of the sufficiency
objective rather than a failure of localisation.}
\label{fig:multi-instance}
\vspace{3cm}
\end{figure*}

This behaviour is not a limitation but a direct consequence of the sufficiency formulation: the method correctly identifies that only a fraction of the visible evidence is necessary, even when more
is available. \Cref{fig:multi-instance} illustrates this property across fourteen ImageNet validation examples spanning a variety of multi-instance scenes.

\subsection{Stability under Attribution-Guided Perturbations}

To quantitatively probe the distinction between sufficiency and sensitivity, we evaluate the stability of the selected evidence under attribution-guided adversarial perturbations. We perform a masked PGD attack on 100 correctly classified images from ImageNet, restricting
the perturbation to the top $5\%$ pixels selected by each attribution method. We report the smallest perturbation budget $\epsilon^*$ at which the attack succeeds and the corresponding attack success rate (ASR) at
$\epsilon=8/255$.

As shown in \cref{tab:adver_attack}, \our{} consistently requires at least as large a perturbation budget as the sensitivity-based baselines. On ViT-B/16, it requires $4/255$ and attains the lowest ASR (85.2\%). On DeiT-III-B/16, it requires $8/255$ and ties for the lowest ASR (56.5\%). These results indicate that the evidence selected by \our{} is less susceptible to attribution-guided perturbations than evidence selected by most sensitivity-based methods, providing quantitative support for the distinction between sufficient evidence and local sensitivity.

\begin{table}[h]
\centering
\small
\begin{tabular}{@{}lcccc@{}}
\toprule
& \multicolumn{2}{c}{ViT-B/16} & \multicolumn{2}{c}{DeiT-III-B/16}\\
Method & $\epsilon^*\downarrow$ & ASR$\uparrow$
& $\epsilon^*\downarrow$ & ASR$\uparrow$\\
\midrule
\our{} (ours) & 4/255 & 85.2 & 8/255 & 56.5\\
IxG & 2/255 & 97.5 & 4/255 & 61.2\\
SmoothGrad & 2/255 & 92.6 & 4/255 & 56.5\\
IntGrad & 2/255 & 98.8 & 4/255 & 71.8\\
DAVE & 4/255 & 92.6 & 8/255 & 62.4\\
LeGrad & 4/255 & 90.1 & 4/255 & 65.9\\
\bottomrule
\end{tabular}
\caption{\textbf{Stability of attribution-selected pixels under masked PGD.} The attack is restricted to the top $5\%$ pixels selected by each method over 100 correctly classified images from ImageNet. $\epsilon^*$ is the smallest perturbation budget at which the attack succeeds, while ASR is the attack success rate at $\epsilon=8/255$. Higher $\epsilon^*$ and lower ASR indicate greater stability of the selected evidence.}
\label{tab:adver_attack}
\vspace{-0.3cm}
\end{table}

\section{Sanity Check: Dependence on Learned Parameters}

The visual appearance of \our{} maps can resemble edge detectors. To evaluate whether the maps are determined primarily by low-level image structure rather than by the learned model parameters, we perform a parameter-randomization sanity check following the motivation of~\cite{adebayo2018sanity}. We compare masks produced by the original pretrained model with masks produced after fully randomizing all model parameters. For the randomized models, each parameter is independently reinitialized using an initialization distribution appropriate to its role (Kaiming, uniform, or normal initialization). We then rerun the complete \our{} mask-generation pipeline without changing any other component of the
procedure. The experiment is performed on 100 ImageNet validation images.

\Cref{tab:sanity_full_randomization} summarizes the agreement between the
masks produced by the pretrained and fully randomized models. The agreement
is low across all three measures: Pearson correlation is $0.278\pm0.059$,
cosine similarity is $0.357\pm0.066$, and IoU at a threshold of $0.5$ is
$0.042\pm0.038$.

These results show that the masks change substantially when the learned model
structure is destroyed. In particular, the low IoU indicates that the
pretrained and randomized models do not select the same image regions,
despite receiving identical inputs. This provides evidence against the
hypothesis that \our{} defaults to a fixed, input-driven edge detector
independent of the learned parameters.

\section{Evaluation composite: grey vs.\ blur fill}
\label{app:eval-composite}

The cosine and class-probability numbers reported throughout the main text and appendix use the saliency-canonical evaluation composite
$$\tilde{\img}_{\text{grey}}=M\!\cdot\!\img+(1{-}M)\!\cdot\!\text{grey}$$ in which non-salient pixels are replaced by neutral grey. This is the standard ``information removal'' baseline in saliency literature and gives the cleanest interpretation: what survives when only the mask pixels remain.

\begin{table}[t]
\centering
\small
\setlength{\tabcolsep}{6pt}

\begin{tabular}{lc}
\toprule
Metric & Original vs.\ 100\% random \\
\midrule
Pearson $\downarrow$ & 0.278 $\pm$ 0.059 \\
Cosine $\downarrow$ & 0.357 $\pm$ 0.066 \\
IoU@0.5 $\downarrow$ & 0.042 $\pm$ 0.038 \\
\bottomrule
\end{tabular}
\caption{\textbf{Positive-control sanity check} \our{} masks computed with the original pretrained model vs. with every model parameter freshly randomized over 100 ImageNet validation images. Low similarity confirms the pipeline can detect a maximal loss of learned structure.}
\label{tab:sanity_full_randomization}
\vspace{-0.3cm}
\end{table}
\begin{figure}[h!]
\centering
\includegraphics[width=\columnwidth]{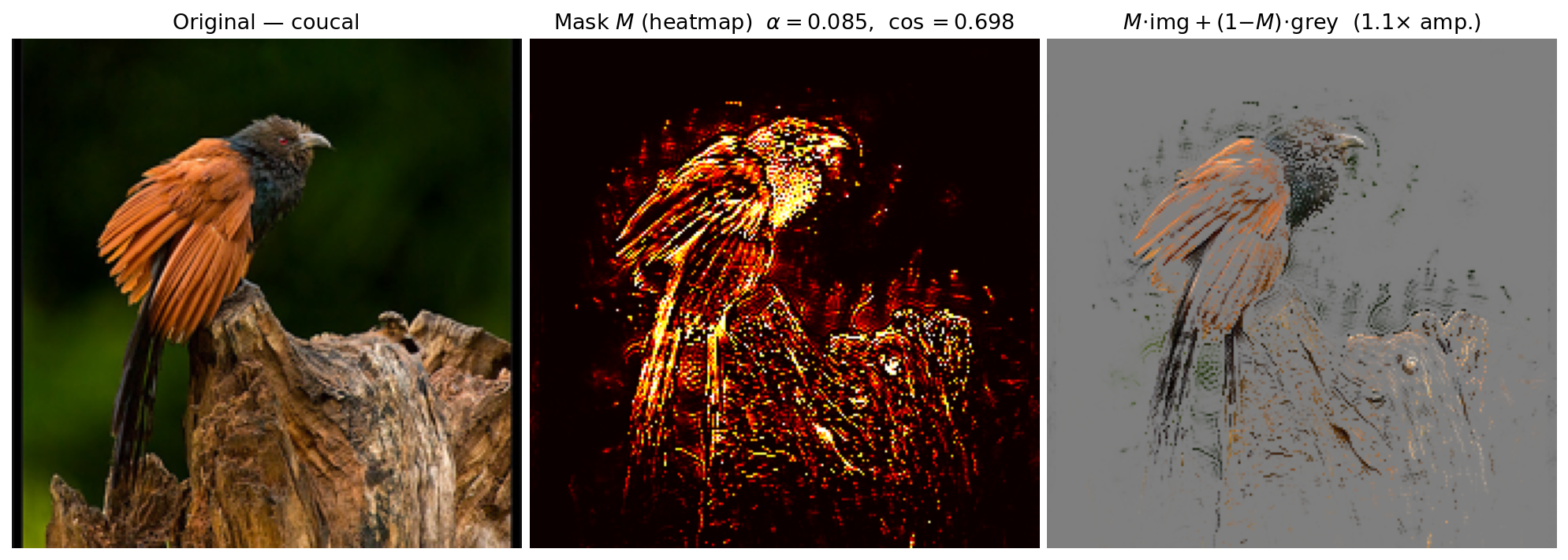}
\caption{Multi-class scene panel: ImageNet ice-cream image (idx\,$44710$), ViT-S/16, $T\!=\!2000$ steps, self-augmented
distractor + blur fill, no auxiliary dataset. From left to right: the original image; the optimised mask $M$ as a heatmap (colourmap on black, rescaled by per-image max and amplified by
$1.1\!\times$ for visibility) and the rescaled composite $M\!\cdot\!\img+(1{-}M)\!\cdot\! grey$. About ${\sim}10\%$
of pixels ($\alpha=0.105$, $\cos=0.770$ on the grey-fill composite) preserve much of the encoder's CLS representation, and the mask is obtained by per-image optimisation with no access to a distractor pool.}
\label{fig:teaser}
\end{figure}

\begin{figure}[!h]
\centering
\includegraphics[width=\columnwidth]{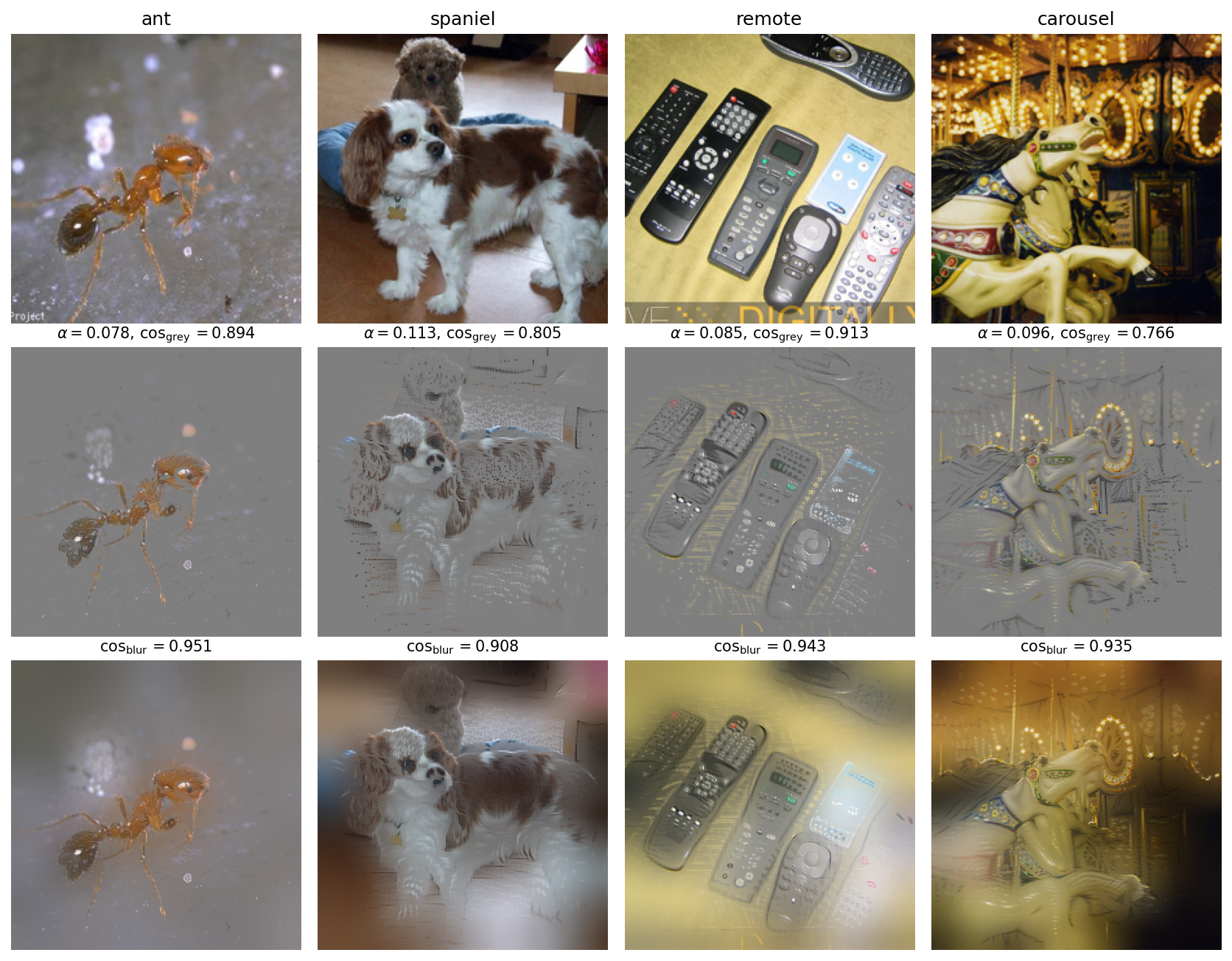}
\caption{Columns are the four standard images. Row~1: originals. Row~2: $\tilde{\img}_{\text{grey}}=M\!\cdot\!\img+(1{-}M)\!\cdot\!\text{grey}$. Row~3: $\tilde{\img}_{\text{blur}}=M\!\cdot\!\img+(1{-}M)\!\cdot\!\text{blur}_{\sigma=20}(\img)$ the training-matched composite. Cosine similarities to the original CLS representation rise substantially when the fill is changed from grey to blur (per-image numbers in titles); see \cref{tab:eval-composite}.}
\label{fig:grey-vs-blur}
\vspace{-0.3cm}
\end{figure}

However, it differs deliberately from the composite the loss, presented in the main text, optimises against, which always contains a heavy blur of the query in the non-salient region (and a capped self-distractor). The mask is therefore evaluated in a regime it did not train under, so
its reported $\cos$/$p_{\text{masked}}$ are conservative lower bounds.
\Cref{fig:grey-vs-blur} contrasts the two composites on the four
standard images, and \cref{tab:eval-composite} reports their means on the same images (and, where useful, on the full $10$-image set). Switching the non-salient fill from grey to
$\text{blur}_{\sigma=20}(\img)$, the training-matched composite,
raises the average CLS cosine from $0.845$ to $0.934$ on the four images, and from $0.740$ to $0.918$ across all~$10$
(\textit{carousel} alone jumps from $0.766$ to $0.935$). The
class-probability target gains even more: the apparent ``train/eval gap'' on \emph{spaniel}($p_{\text{grey}}\!=\!0.34$) and \emph{coucal} ($p_{\text{grey}}\!=\!0.26$) almost completely disappears once the non-salient context is restored ($p_{\text{blur}}\!=\!0.93$ and $1.00$ respectively). This confirms that the masks faithfully preserve $g$ under the composite they
were trained on. The apparent shortfall on the grey-fill metric is a measure of how much the mask relies on the blur fill's low-frequency context being present, not a defect of the optimisation.

\section{Class-conditioned saliency on a multi-class image}
\label{sec:class-prob}

The preservation-MSE objective, described in the main text, is the same regardless of $g$.  Setting $g=p_c$, the softmax probability of a specific ImageNet class~$c$, gives a class-conditioned saliency map: the smallest set of pixels sufficient for the model to recognise class~$c$. No other part of the pipeline changes. Argmax-class examples appear in the main text. Here we turn to the more interesting case where the same image contains two ImageNet classes recognised by the model with comparable confidence.

\begin{table}[!h]
\centering
\begin{tabular}{@{}l@{\;\;}ccc@{}}
  \toprule
  Target & metric & grey fill & blur fill \\
  \midrule
  $g\!=\!\repr_{\text{CLS}}$ ($10$-image mean)
                                          & $\overline{\cos}$       & 0.740 & \textbf{0.918} \\
  $g\!=\!p_c$  ($10$-image mean)
                                          & $\bar{p}_{\text{masked}}$ & 0.682 & \textbf{0.934} \\
  $g\!=\!p_c$ (\emph{spaniel}, $c\!=\!156$)   & $p_{\text{masked}}$     & 0.341 & \textbf{0.932} \\
  $g\!=\!p_c$ (\emph{coucal},  $c\!=\!91$)    & $p_{\text{masked}}$     & 0.261 & \textbf{1.000} \\
  $g\!=\!p_c$ (\emph{carousel}, $c\!=\!476$)  & $p_{\text{masked}}$     & 0.967 & \textbf{0.968} \\
  \bottomrule
\end{tabular}
\caption{Effect of the evaluation composite fill on reported fidelity. Means over the four standard images for the CLS-representation target; per-target-class probabilities for the class-prob target. The blur fill is the same one the loss is trained against. The grey fill removes all non-salient information. Reported $\alpha$ is unaffected, only the metric definition changes.}
\label{tab:eval-composite}
\vspace{-0.5cm}
\end{table}

\begin{figure*}[!h]
\centering
\includegraphics[width=\linewidth]{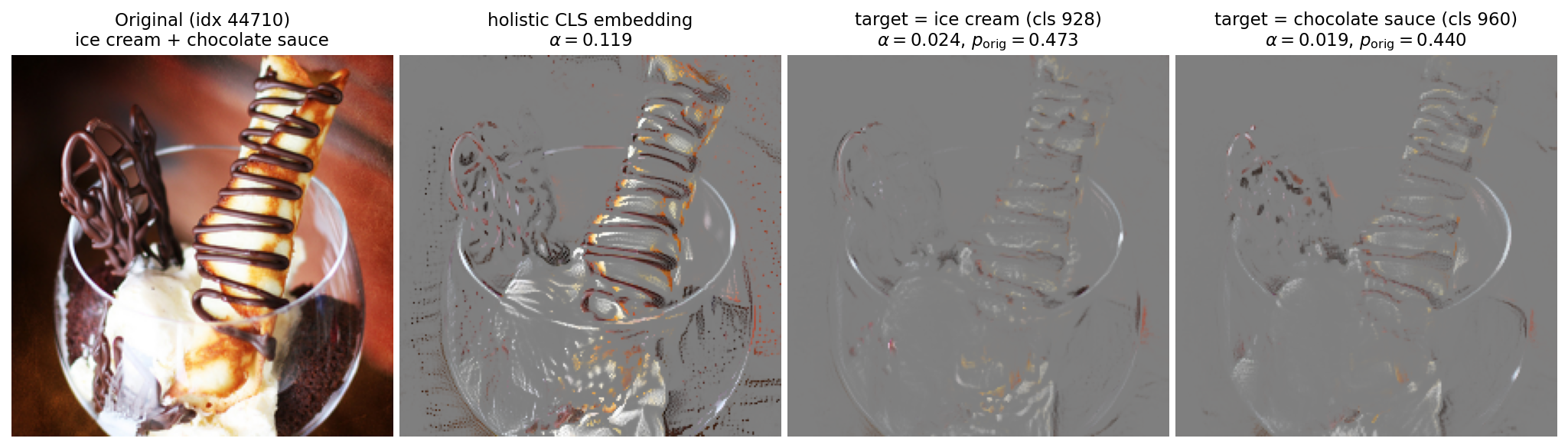}
\caption{Class-conditioned saliency on a multi-class scene (ImageNet validation image idx\,$44710$). The target output $g$ changes between columns. Column~1: original image.  Column~2: holistic mask ($g=\repr_{\text{CLS}}$), $\alpha=0.105$, covering both the ice-cream scoop and the chocolate drizzle.  Column~3: $g=p_{c=928}$ (\emph{ice cream}, $p_{\text{orig}}=0.47$), $\alpha=0.016$, concentrating on the white scoop. Column~4: $g=p_{c=960}$ (\emph{chocolate sauce}, $p_{\text{orig}}=0.44$), $\alpha=0.014$, concentrating on the brown drizzle. The two class-conditioned masks isolate visibly different regions of the same scene, each matching the part of the image that drives the model towards that class.}
\label{fig:multiclass}
\end{figure*}
\paragraph{Multi-class scene.}
We use ImageNet validation image idx\,$44710$, a photograph of an ice-cream sundae topped with chocolate drizzle.  On this image the frozen ViT-S/16 places substantial probability on two distinct ImageNet classes that naturally co-occur in the scene:
\emph{ice cream} (class~$928$, $p_{\text{orig}}=0.47$) and
\emph{chocolate sauce} (class~$960$, $p_{\text{orig}}=0.44$).  We run the optimisation twice with the same hyperparameters and the same image, changing only the target class.  \Cref{fig:multiclass} shows the result. The cls-embedding mask spreads over both the ice-cream scoop and the chocolate drizzle. The class-conditioned masks concentrate on visibly different regions: the ice-cream target ($c=928$) selects pixels on the white scoop at the bottom of the
glass, while the chocolate-sauce target ($c=960$) selects the brown drizzle running over the top of the sundae.

\paragraph{Three-panel view of the same image.}
For reference, \cref{fig:teaser} shows the same ice-cream sundae
under the standard three-panel schema used throughout the paper
(original\,/\,mask\,/\,mask on grey), with the holistic
CLS-embedding target only.

\section{Extended Faithfulness Evaluation}
\label{app:faith}

\begin{table*}[!th]
  \centering
  \small
  \setlength{\tabcolsep}{3pt}
  \begin{tabular}{@{}lcccc@{}}
    \toprule
     & \multicolumn{2}{c}{top $10\%$ kept $\uparrow$} & \multicolumn{2}{c}{top $20\%$ kept $\uparrow$} \\
    \cmidrule(lr){2-3} \cmidrule(lr){4-5}
    Method & ViT-B/16 & DeiT-III-B/16 & ViT-B/16 & DeiT-III-B/16 \\
    \midrule
    IxG & $27.5 \pm 32.9$\,\scriptsize[24.5, 30.7]\,\dag & $29.4 \pm 36.1$\,\scriptsize[26.1, 32.8]\,\dag & $64.8 \pm 30.8$\,\scriptsize[61.8, 67.6]\,\dag & $69.2 \pm 32.6$\,\scriptsize[66.0, 72.1]\,\dag \\
    SmoothGrad & $14.3 \pm 26.0$\,\scriptsize[11.9, 16.8]\,\dag & $13.5 \pm 27.3$\,\scriptsize[11.1, 16.3]\,\dag & $38.8 \pm 34.2$\,\scriptsize[35.6, 42.1]\,\dag & $41.9 \pm 38.9$\,\scriptsize[38.2, 45.6]\,\dag \\
    IntGrad & $51.0 \pm 35.3$\,\scriptsize[47.6, 54.3]\,\dag & $49.6 \pm 39.4$\,\scriptsize[45.8, 53.2]\,\dag & $75.3 \pm 22.7$\,\scriptsize[72.9, 77.3]\,\dag & $79.0 \pm 24.3$\,\scriptsize[76.5, 81.2] \\
    A-LRP & $54.8 \pm 34.2$\,\scriptsize[51.4, 57.9]\,\dag & $50.6 \pm 38.2$\,\scriptsize[47.0, 54.2]\,\dag & $71.8 \pm 24.2$\,\scriptsize[69.2, 73.9]\,\dag & $76.1 \pm 26.0$\,\scriptsize[73.4, 78.4]\,\dag \\
    C-LRP & $30.8 \pm 35.0$\,\scriptsize[27.4, 34.1]\,\dag & --- & $56.8 \pm 34.1$\,\scriptsize[53.5, 60.0]\,\dag & --- \\
    LeGrad & $26.0 \pm 34.0$\,\scriptsize[22.9, 29.4]\,\dag & $27.3 \pm 36.2$\,\scriptsize[23.9, 30.9]\,\dag & $53.4 \pm 35.2$\,\scriptsize[50.0, 56.8]\,\dag & $56.2 \pm 38.2$\,\scriptsize[52.5, 59.8]\,\dag \\
    DAVE & $56.4 \pm 34.3$\,\scriptsize[53.0, 59.5]\,\dag & $58.9 \pm 36.1$\,\scriptsize[55.4, 62.3]\,\dag & $70.7 \pm 25.9$\,\scriptsize[68.0, 73.0]\,\dag & $75.4 \pm 27.3$\,\scriptsize[72.6, 77.9]\,\dag \\
    Extremal Pert. & $23.7 \pm 32.4$\,\scriptsize[20.6, 26.8]\,\dag & $26.2 \pm 34.8$\,\scriptsize[22.9, 29.6]\,\dag & $47.7 \pm 36.3$\,\scriptsize[44.2, 51.1]\,\dag & $54.6 \pm 37.6$\,\scriptsize[50.9, 58.2]\,\dag \\
    iGOS++ & $42.8 \pm 36.1$\,\scriptsize[39.3, 46.2]\,\dag & $45.1 \pm 39.3$\,\scriptsize[41.3, 48.8]\,\dag & $67.5 \pm 29.6$\,\scriptsize[64.5, 70.3]\,\dag & $72.2 \pm 30.1$\,\scriptsize[69.1, 74.9]\,\dag \\
    \midrule
    \rowcolor[HTML]{D5E4FF}
    \textbf{\our{} (ours)} & $\mathbf{74.2 \pm 24.4}$\,\scriptsize[71.7, 76.4] & $\mathbf{76.3 \pm 25.5}$\,\scriptsize[73.7, 78.6] & $\mathbf{80.2 \pm 16.6}$\,\scriptsize[78.5, 81.6] & $\mathbf{83.2 \pm 18.7}$\,\scriptsize[81.2, 84.8] \\
    \bottomrule
  \end{tabular}
  \caption{\textbf{ROAD sufficiency at tight pixel budgets.}
Mean target-class probability (\%) when only the top $10\%$ or $20\%$ of pixels are retained and the remaining pixels are imputed using ROAD. Higher is better. Brackets denote BCa 95\% confidence intervals. $\dag$/$\ddag$ indicate significantly lower/higher performance than \our{} after Holm--Bonferroni correction.}
\label{tab:road-sufficiency-full}
\end{table*}

\subsection{Evaluation Protocol and Implementation Details}
We use the same evaluation subset and attribution maps as in the main text. For pixel deletion, pixels are ranked by attribution score and removed in ascending order, from least to most relevant. Removed pixels are replaced by the channel-wise mean of the normalized backbone input. We evaluate $M=128$ uniformly spaced masking levels up to $90\%$ removed pixels and record the target-class softmax probability at each level. Only correctly classified images are included.

The deletion score is computed using the trapezoidal rule,
\begin{equation}
\mathrm{AUC}=\int_0^{90}p(t)\,\mathrm{d}t,
\end{equation}
where $t$ is the percentage of removed pixels and $p(t)$ is the target-class probability. Higher AUC indicates a more faithful attribution ranking.

For \our{}, we use three independent mask optimizations per image and $T=500$ Adam iterations; the final attribution map is the element-wise median of the three optimized masks. DAVE uses its default Gaussian and bilateral post-processing. IxG, IntGrad, SmoothGrad, A-LRP, and LeGrad use their standard implementations. C-LRP is evaluated only on ViT-B/16 because the original implementation is incompatible with DeiT-III-B/16.

For ROAD, removed pixels are imputed using noisy linear imputation in image space before backbone normalization. We evaluate both LeRF and MoRF orders over the full $10$--$90\%$ range, as well as sufficiency at fixed $10\%$ and $20\%$ pixel budgets.

\subsubsection{Insertion Evaluation}

As a complementary evaluation to pixel deletion presented in the main text, we also measure insertion faithfulness. Starting from the fully masked input, pixels are progressively revealed in descending attribution order using the same channel-mean fill protocol. Higher AUC therefore indicates that the attribution ranking places predictive pixels earlier in the insertion sequence.

\Cref{tab:insertion} shows that \our{} obtains the highest insertion AUC on both architectures, with scores of $64.9$ and $66.3$ on ViT-B/16 and DeiT-III-B/16, respectively.

This provides a complementary view of the deletion results reported in the main text: the pixels ranked as most relevant by \our{} recover target-class evidence more rapidly when progressively revealed.

\begin{table}[th]
  \centering
  \small
  \setlength{\tabcolsep}{3pt}
  \resizebox{\columnwidth}{!}{%
  \begin{tabular}{@{}lcc@{}}
    \toprule
    Method & ViT-B/16 $\uparrow$ & DeiT-III-B/16 $\uparrow$ \\
    \midrule
    IxG & $34.6 \pm 20.6$\,\scriptsize[32.6, 36.6]\,\dag & $20.8 \pm 18.2$\,\scriptsize[19.1, 22.5]\,\dag \\
    SmoothGrad & $32.8 \pm 12.6$\,\scriptsize[31.7, 34.0]\,\dag & $37.2 \pm 16.7$\,\scriptsize[35.6, 38.8]\,\dag \\
    IntGrad & $47.0 \pm 21.8$\,\scriptsize[44.9, 49.1]\,\dag & $37.3 \pm 24.8$\,\scriptsize[34.9, 39.6]\,\dag \\
    A-LRP & $49.2 \pm 22.4$\,\scriptsize[46.9, 51.2]\,\dag & $38.7 \pm 23.7$\,\scriptsize[36.4, 40.9]\,\dag \\
    C-LRP & $49.4 \pm 14.9$\,\scriptsize[48.0, 50.8]\,\dag & --- \\
    LeGrad & $50.0 \pm 14.3$\,\scriptsize[48.6, 51.3]\,\dag & $55.9 \pm 16.1$\,\scriptsize[54.2, 57.3]\,\dag \\
    DAVE & $57.7 \pm 20.5$\,\scriptsize[55.6, 59.5]\,\dag & $56.5 \pm 24.3$\,\scriptsize[54.1, 58.7]\,\dag \\
    Extremal Pert. & $48.8 \pm 14.8$\,\scriptsize[47.3, 50.2]\,\dag & $48.8 \pm 17.4$\,\scriptsize[47.1, 50.4]\,\dag \\
    iGOS++ & $53.2 \pm 14.0$\,\scriptsize[51.8, 54.5]\,\dag & $58.2 \pm 15.5$\,\scriptsize[56.7, 59.7]\,\dag \\
    \midrule
    \rowcolor[HTML]{D5E4FF}
    \textbf{\our{} (ours)} & $\mathbf{64.9 \pm 15.0}$\,\scriptsize[63.4, 66.3] & $\mathbf{66.3 \pm 17.3}$\,\scriptsize[64.5, 67.8] \\
    \bottomrule
  \end{tabular}}
  \caption{\textbf{Insertion AUC under the channel-mean fill} pixels revealed in descending attribution order; higher is better, with BCa 95\,\% confidence intervals. Reported for completeness; the artefact-free counterpart is ROAD-LeRF in \cref{tab:road-lerf}. $\dag$/$\ddag$: significantly better/worse than \our{} after Holm-Bonferroni correction.}
  \label{tab:insertion}
  \vspace{-0.3cm}
\end{table}
\subsubsection{Per-Budget Deletion Results}

\begin{figure*}[h]
    \centering
    \includegraphics[width=\linewidth]{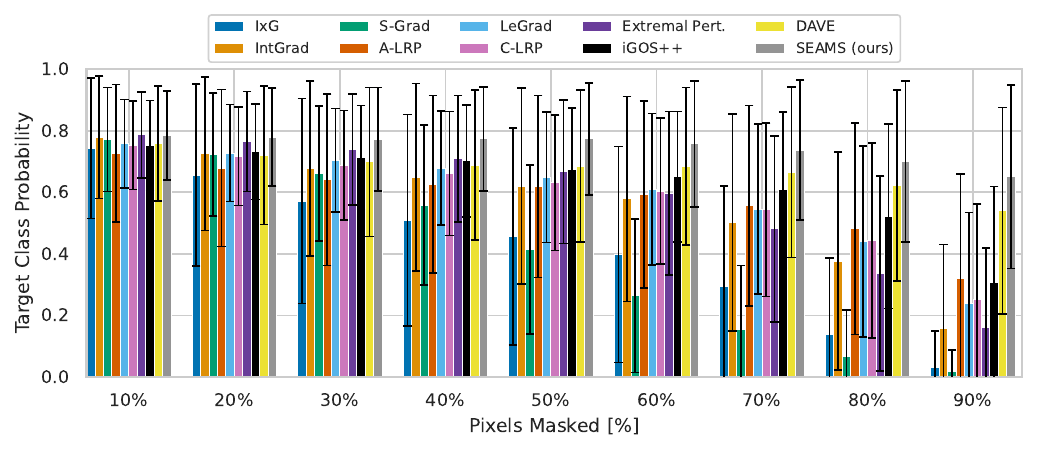}
    \caption{
    Mean target-class probability at each 10\% masking step for ViT-B/16. \our{} (grey) achieves the highest probability across all masking levels. C-LRP is evaluated only on ViT-B/16.
    }
    \label{fig:pixdel-bars-vit}
    \vspace{-6cm}
\end{figure*}

\begin{figure*}[h]
    \centering
    \includegraphics[width=\linewidth]{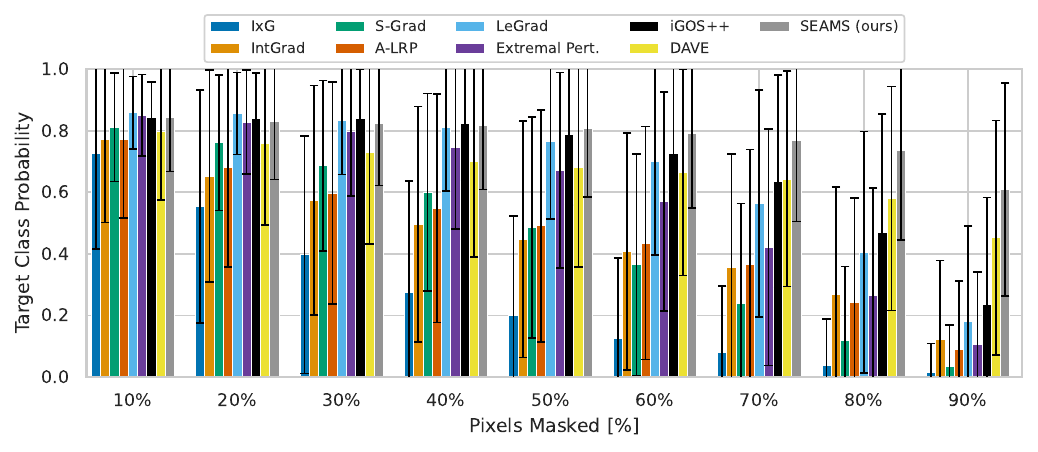}
    \caption{
    Mean target-class probability at each 10\% masking step for DeiT-III-B/16. \our{} (grey) consistently preserves the highest target-class probability. C-LRP is omitted due to architectural incompatibility.
    }
    \label{fig:pixdel-bars-deit}
\end{figure*}

\Cref{fig:pixdel-bars-vit,fig:pixdel-bars-deit} provide the mean target-class probability at each $10\%$ masking step for ViT-B/16 and DeiT-III-B/16, respectively. These plots complement the aggregate AUC results in the main text by exposing the behavior at individual masking budgets. On both architectures, \our{} remains above the competing methods throughout the evaluated masking range, with the separation becoming particularly pronounced beyond $40\%$ masking.

\subsection{Extended ROAD Results}

The main text reports the most stringent ROAD sufficiency setting, with only the top $10\%$ of pixels retained. Here we provide the complementary $20\%$ sufficiency results together with full-range LeRF and MoRF evaluations.

\begin{table}[!ht]
  \centering
  \small
  \setlength{\tabcolsep}{3pt}
  \resizebox{\columnwidth}{!}{%
  \begin{tabular}{@{}lcc@{}}
    \toprule
    Method & ViT-B/16 $\downarrow$ & DeiT-III-B/16 $\downarrow$ \\
    \midrule
    IxG & $55.1 \pm 26.9$\,\scriptsize[52.4, 57.6]\,\dag & $64.7 \pm 27.7$\,\scriptsize[61.9, 67.2]\,\dag \\
    SmoothGrad & $52.7 \pm 26.7$\,\scriptsize[50.1, 55.2]\,\dag & $51.6 \pm 27.8$\,\scriptsize[49.0, 54.2]\,\dag \\
    IntGrad & $51.2 \pm 29.0$\,\scriptsize[48.4, 53.9]\,\dag & $61.2 \pm 31.0$\,\scriptsize[58.2, 64.2]\,\dag \\
    A-LRP & $66.4 \pm 24.2$\,\scriptsize[64.0, 68.6]\,\dag & $72.3 \pm 23.0$\,\scriptsize[70.0, 74.5]\,\dag \\
    C-LRP & $23.2 \pm 19.0$\,\scriptsize[21.5, 25.1]\,\ddag & --- \\
    LeGrad & $23.2 \pm 19.1$\,\scriptsize[21.4, 25.1]\,\ddag & $23.7 \pm 20.3$\,\scriptsize[21.9, 25.7]\,\ddag \\
    DAVE & $64.7 \pm 24.1$\,\scriptsize[62.3, 66.9]\,\dag & $62.0 \pm 28.0$\,\scriptsize[59.3, 64.6]\,\dag \\
    Extremal Pert. & $44.4 \pm 25.6$\,\scriptsize[41.9, 46.8]\,\dag & $47.8 \pm 27.7$\,\scriptsize[45.1, 50.4]\,\dag \\
    iGOS++ & $27.4 \pm 22.8$\,\scriptsize[25.3, 29.6]\,\ddag & $27.9 \pm 24.8$\,\scriptsize[25.7, 30.4]\,\ddag \\
    \midrule
    \rowcolor[HTML]{D5E4FF}
    \our{} (ours) & $39.7 \pm 24.8$\,\scriptsize[37.3, 42.1] & $39.2 \pm 26.0$\,\scriptsize[36.7, 41.7] \\
    \bottomrule
  \end{tabular}}
  \caption{\textbf{ROAD MoRF AUC (lower is better)} with BCa 95\,\% confidence intervals. Removed pixels are replaced by ROAD's noisy linear imputation instead of a grey / channel-mean constant, which removes the out-of-distribution masking artefact and the class information leakage of the standard protocol. $\dag$/$\ddag$: significantly worse/better than \our{} after Holm-Bonferroni correction.}
  \label{tab:road-morf}
\end{table}

\Cref{tab:road-sufficiency-full} presents results for both $10\%$ and $20\%$ masking sufficiency. At $20\%$ kept, \our{} retains $80.2\%$ and $83.2\%$ of the target-class probability on ViT-B/16 and DeiT-III-B/16, respectively. The strongest baselines at this budget are IntGrad, with $75.3\%$ and $79.0\%$. Thus, the advantage remains positive at both budgets, although it becomes smaller as more pixels are retained.

\begin{figure*}[!ht]
  \centering
  \includegraphics[width=0.92\textwidth]{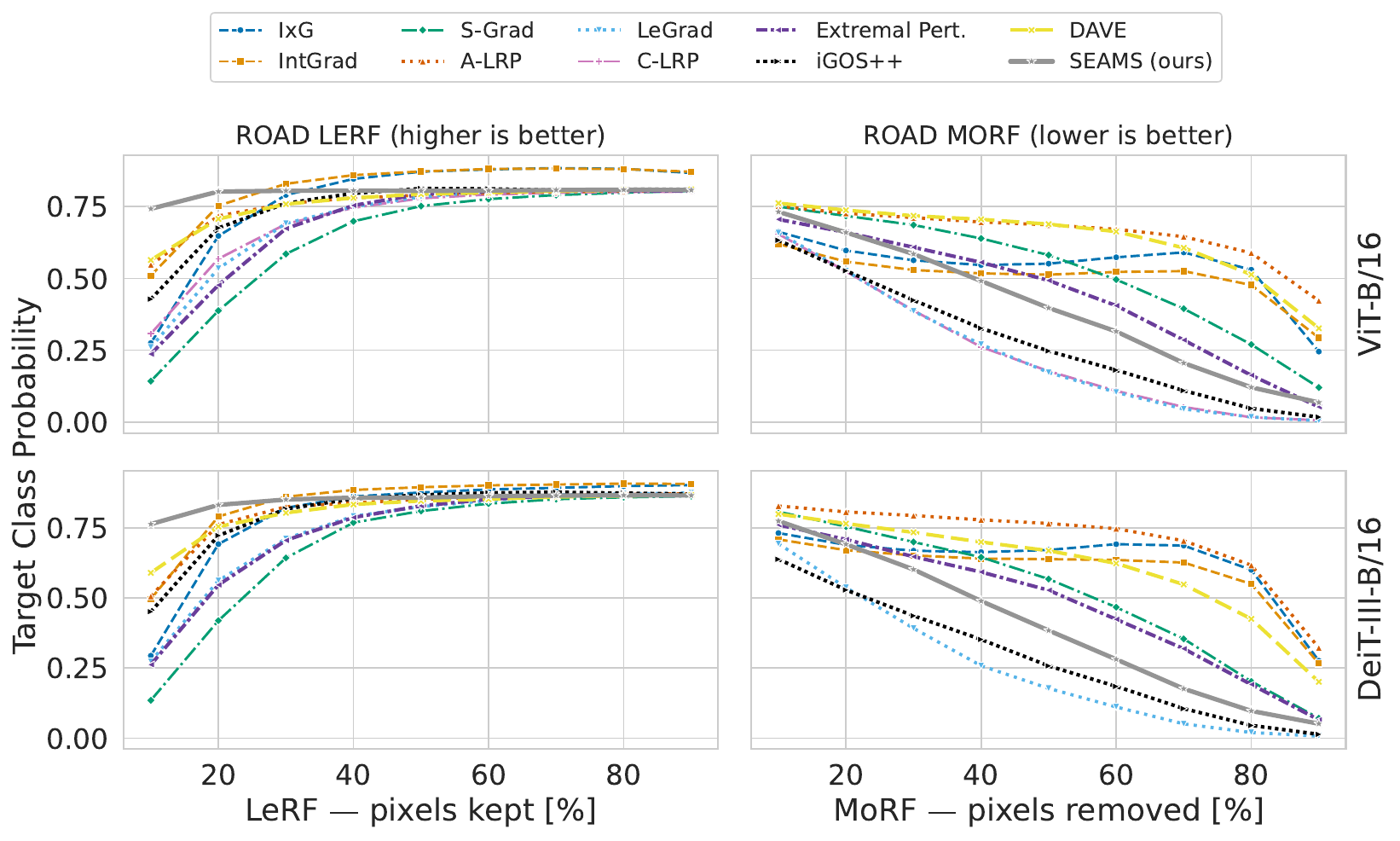}%
  \caption{\textbf{Faithfulness under ROAD.} Target-class probability when
  removed pixels are imputed in distribution instead of being replaced by a
  constant. \emph{Left}: LeRF, where only the top $p$ pixels are kept --- the
  sufficiency regime is the left edge. \emph{Right}: MoRF, where the top $p$
  pixels are removed. \our{} reaches the unmasked probability with about
  $20\%$ of the pixels and stays flat; the curves of signed attributions keep
  rising past that level because imputing their lowest-ranked, i.e.\ most
  negative, pixels removes counter-evidence.}
  \label{fig:road-curves}
  \vspace{-0.6cm}
\end{figure*}

The full-range ROAD results provide a complementary view of faithfulness. On LeRF, see \cref{tab:road-lerf}, \our{} obtains AUCs of $80.2$ and $85.1$, ranking second on both backbones behind IntGrad. On MoRF, \cref{tab:road-morf}, \our{} ranks fourth on ViT-B/16 and third on DeiT-III-B/16. These differences are expected because the three evaluations emphasize different properties: tight-budget LeRF measures sufficiency of a compact set of highly relevant pixels, whereas full-range LeRF is affected by saturation once most methods recover the original prediction, and MoRF emphasizes the concentration of attribution mass in a small number of pixels.

The corresponding curves are shown in \cref{fig:road-curves}. In the LeRF setting, most methods approach the unmasked probability once a sufficiently large fraction of pixels is retained, whereas the MoRF curves remain sensitive to the ranking of highly salient pixels. This saturation explains why full-range AUC and tight-budget sufficiency can induce different rankings.

\subsection{Statistical Analysis}
\label{app:stats}
All comparisons are paired at the image level. For each metric and backbone, we test the paired differences for normality using Shapiro--Wilk and for equal variances using Levene's test. We use a paired $t$-test when both assumptions are satisfied and the Wilcoxon signed-rank test otherwise, with $\alpha=0.05$. Confidence intervals are computed using BCa bootstrap with 10,000 resamples and a fixed seed. $p$-values are corrected using Holm--Bonferroni within each metric--backbone family.

The complete paired comparisons, including effect sizes, corrected $p$-values, and BCa confidence intervals, are reported in \cref{tab:stats-full}. Across the evaluated settings, the few cases in which \our{} does not significantly outperform the strongest baseline occur for full-range ROAD LeRF, ROAD MoRF, and $20\%$ sufficiency on DeiT-III-B/16.

\begin{table}[t]
  \centering
  \small
  \setlength{\tabcolsep}{3pt}
  \resizebox{\columnwidth}{!}{%
  \begin{tabular}{@{}lcc@{}}
    \toprule
    Method & ViT-B/16 $\uparrow$ & DeiT-III-B/16 $\uparrow$ \\
    \midrule
    IxG & $79.6 \pm 11.9$\,\scriptsize[78.4, 80.7]\,\dag & $81.6 \pm 15.3$\,\scriptsize[80.0, 82.9]\,\dag \\
    SmoothGrad & $65.8 \pm 16.7$\,\scriptsize[64.1, 67.3]\,\dag & $71.1 \pm 16.2$\,\scriptsize[69.5, 72.5]\,\dag \\
    IntGrad & $83.1 \pm 9.7$\,\scriptsize[82.1, 83.9]\,\ddag & $85.6 \pm 11.0$\,\scriptsize[84.4, 86.5] \\
    A-LRP & $76.6 \pm 16.0$\,\scriptsize[74.9, 78.0]\,\dag & $82.0 \pm 14.5$\,\scriptsize[80.4, 83.2]\,\dag \\
    C-LRP & $71.6 \pm 17.0$\,\scriptsize[69.8, 73.1]\,\dag & --- \\
    LeGrad & $71.3 \pm 16.2$\,\scriptsize[69.7, 72.8]\,\dag & $75.6 \pm 16.7$\,\scriptsize[73.9, 77.1]\,\dag \\
    DAVE & $76.7 \pm 15.4$\,\scriptsize[75.1, 78.1]\,\dag & $81.8 \pm 15.3$\,\scriptsize[80.2, 83.1]\,\dag \\
    Extremal Pert. & $70.4 \pm 16.0$\,\scriptsize[68.8, 71.8]\,\dag & $75.2 \pm 16.4$\,\scriptsize[73.6, 76.7]\,\dag \\
    iGOS++ & $76.2 \pm 14.5$\,\scriptsize[74.6, 77.4]\,\dag & $81.8 \pm 11.7$\,\scriptsize[80.6, 82.8]\,\dag \\
    \midrule
    \rowcolor[HTML]{D5E4FF}
    \our{} (ours) & $80.2 \pm 13.3$\,\scriptsize[78.8, 81.3] & $85.1 \pm 13.4$\,\scriptsize[83.7, 86.2] \\
    \bottomrule
  \end{tabular}}
  \caption{\textbf{ROAD LeRF AUC (higher is better)} with BCa 95\,\% confidence intervals. Removed pixels are replaced by ROAD's noisy linear imputation instead of a grey / channel-mean constant, which removes the out-of-distribution masking artefact and the class information leakage of the standard protocol. $\dag$/$\ddag$: significantly worse/better than \our{} after Holm-Bonferroni correction.}
  \label{tab:road-lerf}
  \vspace{-0.6cm}
\end{table}

\section{Qualitative Results}
\label{sec:qualitative}
\label{app:qual}

This section presents additional qualitative comparisons on ImageNet-1K validation images. We evaluate three Vision Transformer backbones: ViT-B/16-224, DeiT-B/16-224, and DeiT-III-B/16-224. For all models, no model parameters are modified after training.

\begin{figure*}[thbp]
\centering
\resizebox{\textwidth}{!}{%
\begin{minipage}[c]{.3cm}
    \centering
    \rotatebox{90}{%
      \makebox[87px][c]{DeiT-III-B/16-224}%
      \makebox[87px][c]{DeiT-B/16-224}%
      \makebox[87px][c]{ViT-B/16-224}%
      \makebox[97px][c]{Input\;}%
    }
\end{minipage}%
\begin{tabular}{@{}c@{}} {\small African Hunting Dog} \\ \includegraphics[width=0.18\textwidth]{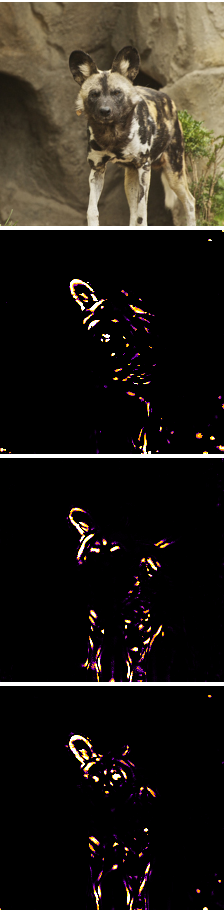} \end{tabular}
\begin{tabular}{@{}c@{}} Ambulance \\ \includegraphics[width=0.18\textwidth]{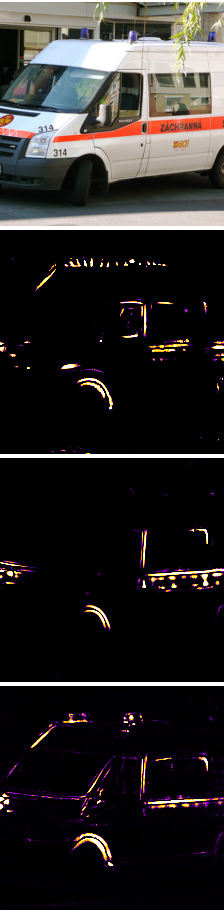} \end{tabular}
\begin{tabular}{@{}c@{}} Ballpoint \\ \includegraphics[width=0.18\textwidth]{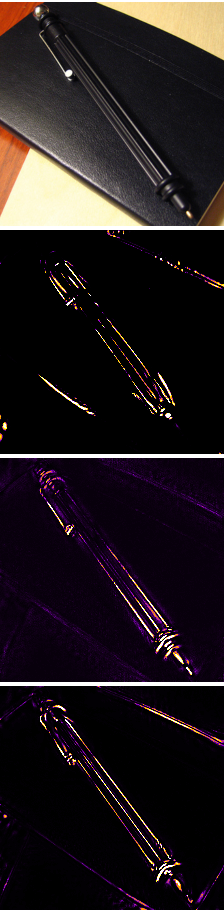} \end{tabular}
\begin{tabular}{@{}c@{}} Dome \\ \includegraphics[width=0.18\textwidth]{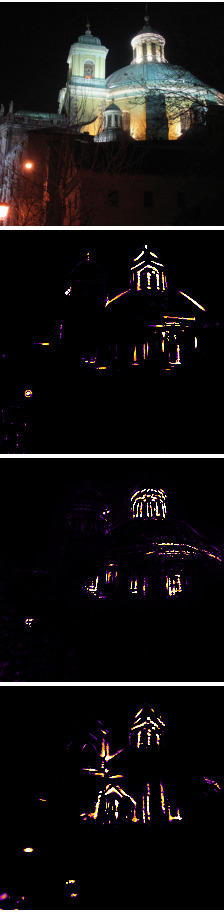} \end{tabular}
\begin{tabular}{@{}c@{}} Dowitcher \\ \includegraphics[width=0.18\textwidth]{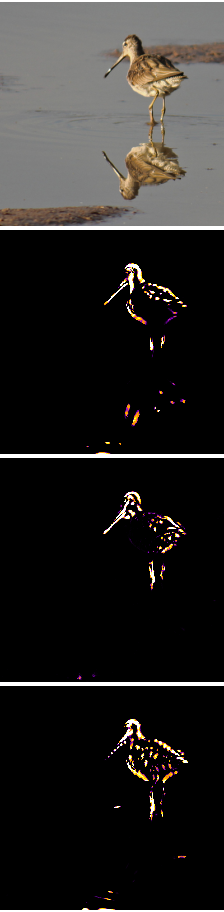} \end{tabular}
\begin{tabular}{@{}c@{}} Eft \\ \includegraphics[width=0.18\textwidth]{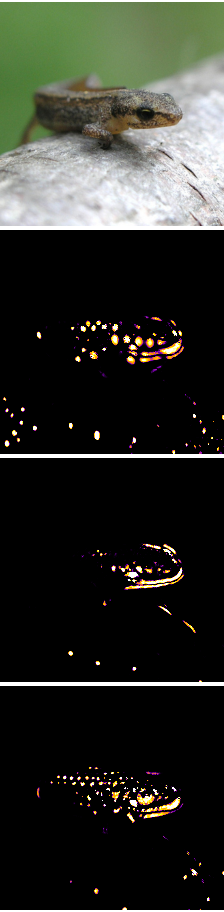} \end{tabular}
\begin{tabular}{@{}c@{}} Fountain \\ \includegraphics[width=0.18\textwidth]{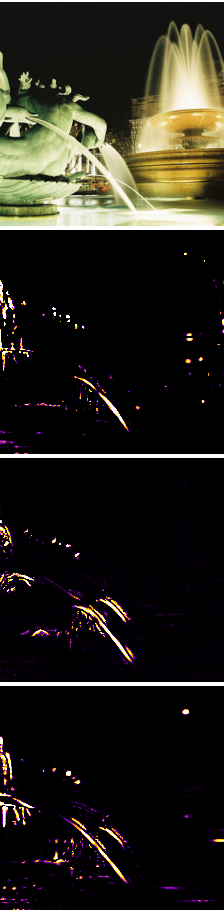} \end{tabular}
\begin{tabular}{@{}c@{}} Hair Spray \\ \includegraphics[width=0.18\textwidth]{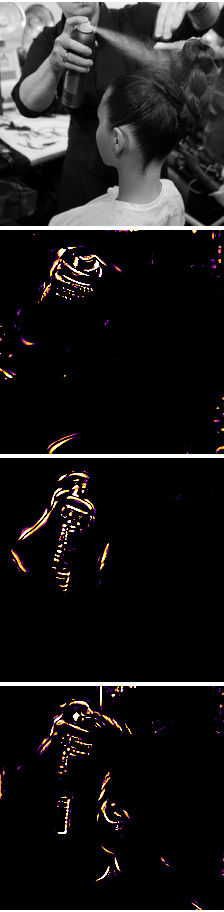} \end{tabular}
\begin{tabular}{@{}c@{}} Hamster \\ \includegraphics[width=0.18\textwidth]{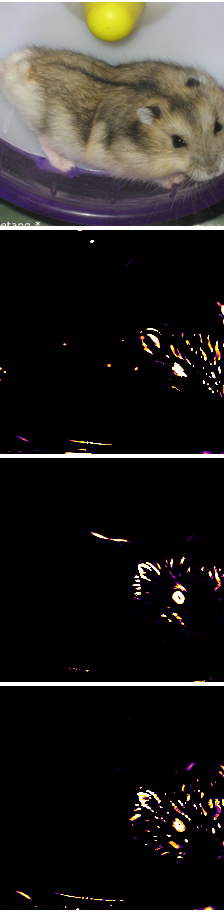} \end{tabular}
}%
\\
\resizebox{\textwidth}{!}{%
\begin{minipage}[c]{.3cm}
    \centering
    \rotatebox{90}{%
      \makebox[87px][c]{DeiT-III-B/16-224}%
      \makebox[87px][c]{DeiT-B/16-224}%
      \makebox[87px][c]{ViT-B/16-224}%
      \makebox[110px][c]{Input\quad}%
    }
\end{minipage}%
\begin{tabular}{@{}c@{}} \\ Hoopskirt \\ \includegraphics[width=0.18\textwidth]{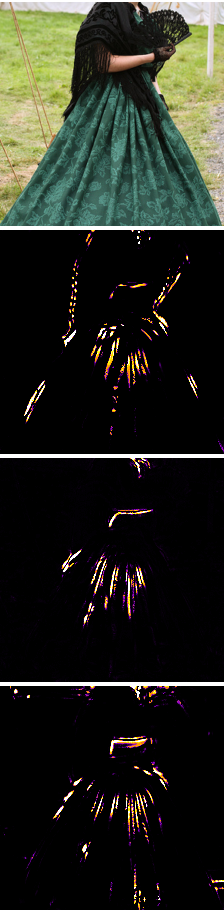} \end{tabular}
\begin{tabular}{@{}c@{}} \\ Hourglass \\ \includegraphics[width=0.18\textwidth]{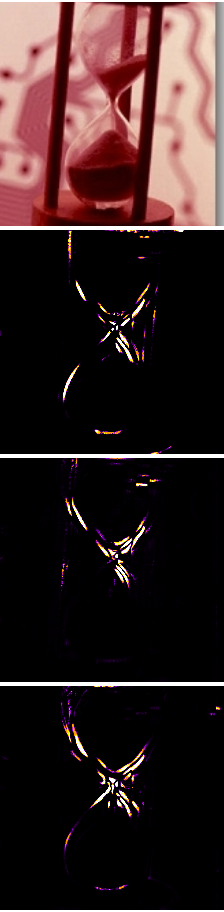} \end{tabular}
\begin{tabular}{@{}c@{}} \\ Hummingbird \\ \includegraphics[width=0.18\textwidth]{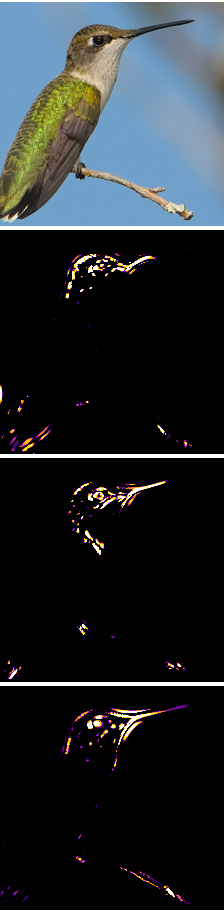} \end{tabular}
\begin{tabular}{@{}c@{}} \\ Iron \\ \includegraphics[width=0.18\textwidth]{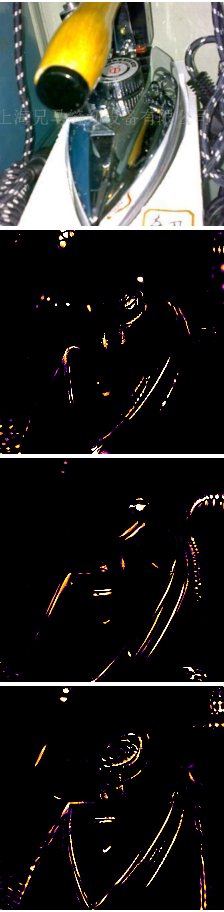} \end{tabular}
\begin{tabular}{@{}c@{}} \\ Macaw \\ \includegraphics[width=0.18\textwidth]{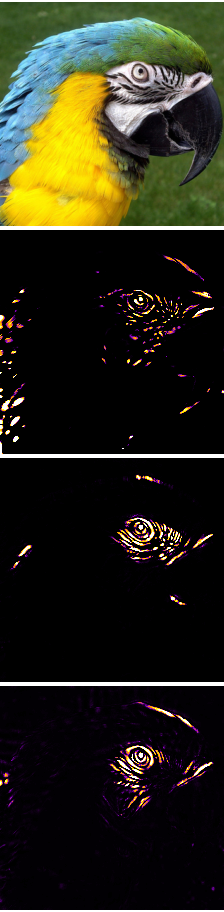} \end{tabular}
\begin{tabular}{@{}c@{}} \\ Mobile Home \\ \includegraphics[width=0.18\textwidth]{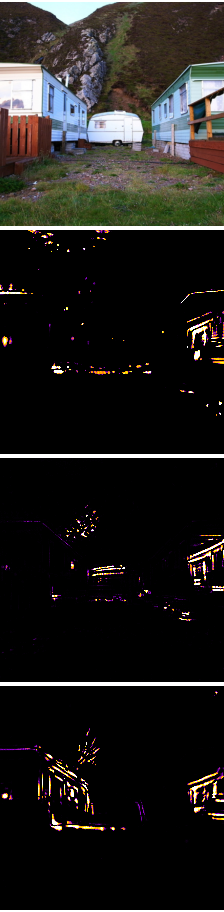} \end{tabular}
\begin{tabular}{@{}c@{}} \\ Revolver \\ \includegraphics[width=0.18\textwidth]{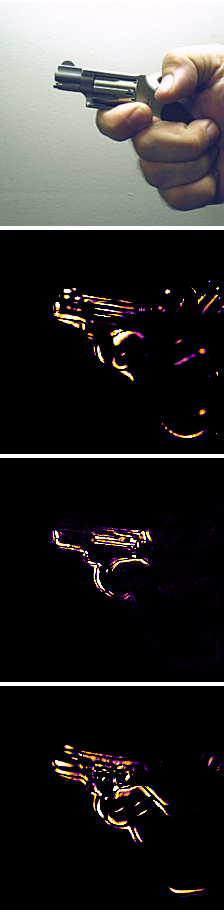} \end{tabular}
\begin{tabular}{@{}c@{}} \\ Streetcar \\ \includegraphics[width=0.18\textwidth]{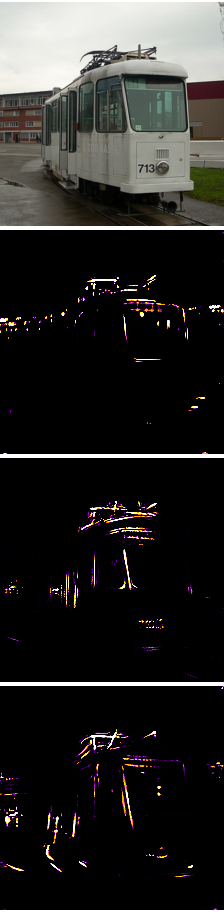} \end{tabular}
\begin{tabular}{@{}c@{}} West Highland \\ White Terrier \\ \includegraphics[width=0.18\textwidth]{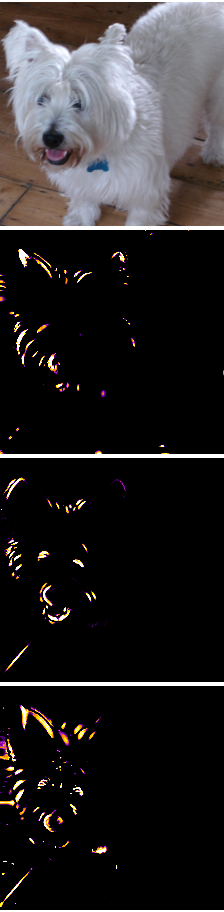} \end{tabular}
}%
\caption{Visual explanations across different ViT architectures. Using validation examples from ImageNet-1K (columns), we present the original input images (top row) alongside the corresponding attribution maps generated by ViT-B/16-224, DeiT-B/16-224, and DeiT-III-B/16-224 (subsequent rows). The attribution maps are computed by weighting token-level contributions with the full class probability vector produced by each model, allowing the explanations to reflect the distribution of confidence across all predicted classes rather than only the top prediction. While all three models generally focus on similar, class-relevant object regions, they occasionally highlight distinct yet semantically meaningful features, showcasing variations in their learned representations.}
\label{fig:vit_backbones}
\end{figure*}

\Cref{fig:vit_backbones} compares attribution maps obtained from class-probability outputs. The explanations show that all three backbones focus on object regions that are relevant for the predicted category. However, the spatial distribution of the highlighted regions differs between architectures, indicating differences in the learned visual representations.

\begin{figure*}[htbp]
\centering
\resizebox{\textwidth}{!}{%
\begin{minipage}[c]{.3cm}
    \centering
    \rotatebox{90}{%
      \makebox[87px][c]{DeiT-III-B/16-224}%
      \makebox[87px][c]{DeiT-B/16-224}%
      \makebox[87px][c]{ViT-B/16-224}%
      \makebox[97px][c]{Input\;}%
    }
\end{minipage}%
\begin{tabular}{@{}c@{}} {\small African Hunting Dog} \\ \includegraphics[width=0.18\textwidth]{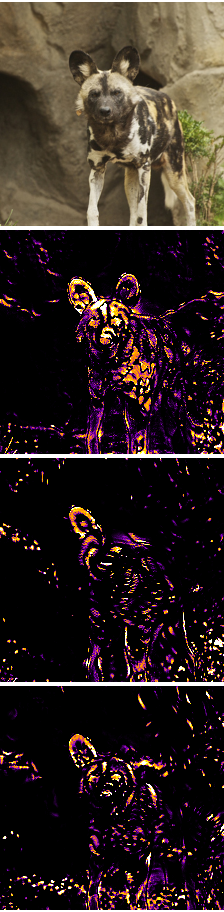} \end{tabular}
\begin{tabular}{@{}c@{}} Ambulance \\ \includegraphics[width=0.18\textwidth]{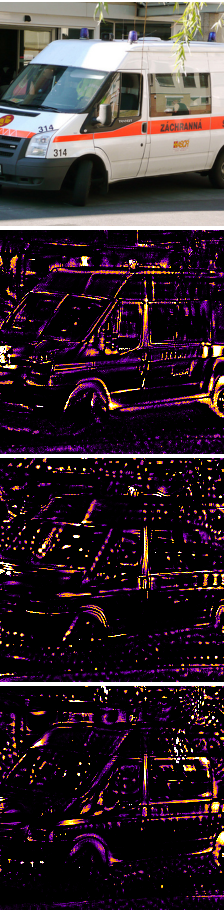} \end{tabular}
\begin{tabular}{@{}c@{}} Ballpoint \\ \includegraphics[width=0.18\textwidth]{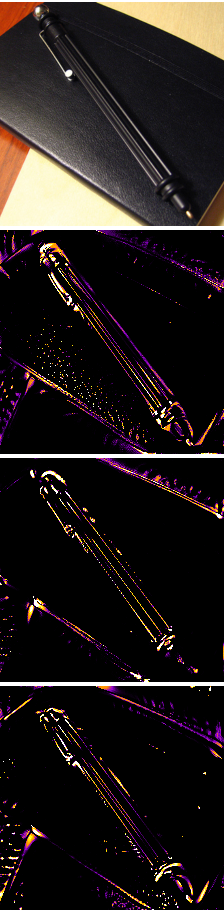} \end{tabular}
\begin{tabular}{@{}c@{}} Dome \\ \includegraphics[width=0.18\textwidth]{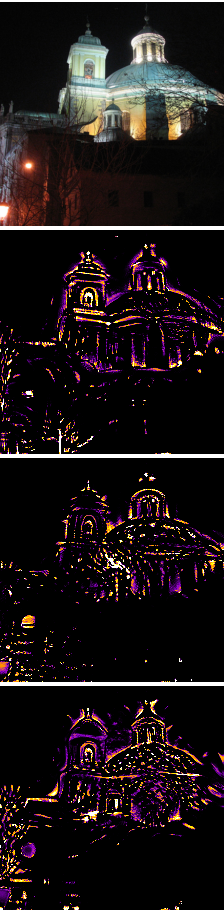} \end{tabular}
\begin{tabular}{@{}c@{}} Dowitcher \\ \includegraphics[width=0.18\textwidth]{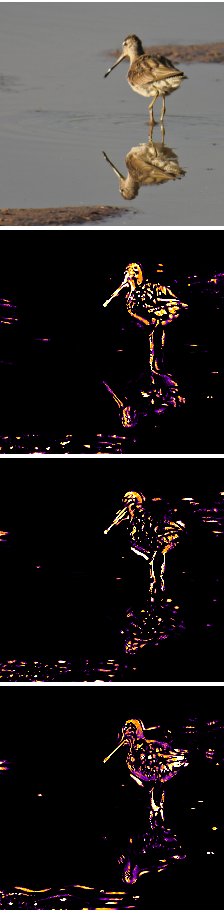} \end{tabular}
\begin{tabular}{@{}c@{}} Eft \\ \includegraphics[width=0.18\textwidth]{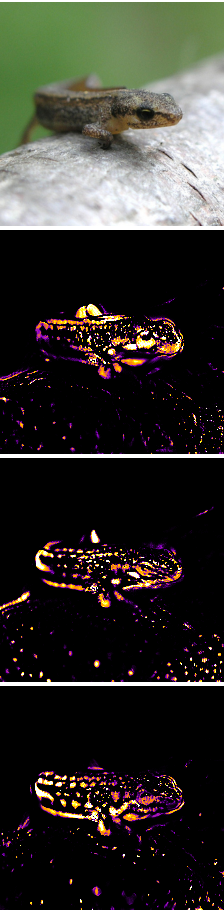} \end{tabular}
\begin{tabular}{@{}c@{}} Fountain \\ \includegraphics[width=0.18\textwidth]{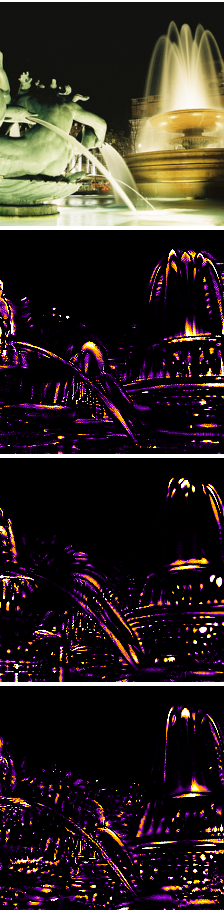} \end{tabular}
\begin{tabular}{@{}c@{}} Hair Spray \\ \includegraphics[width=0.18\textwidth]{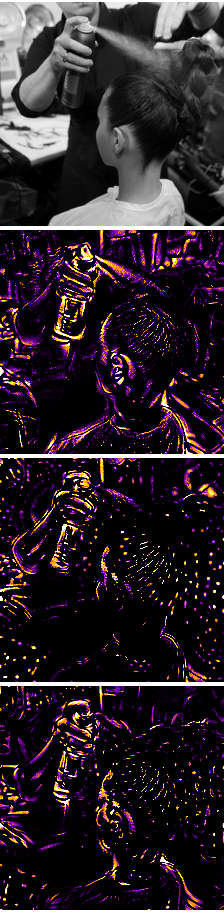} \end{tabular}
\begin{tabular}{@{}c@{}} Hamster \\ \includegraphics[width=0.18\textwidth]{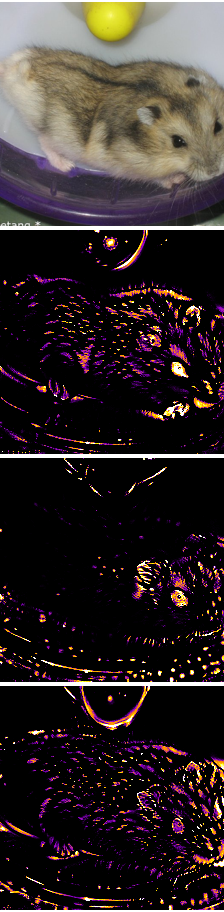} \end{tabular}
}%
\\
\resizebox{\textwidth}{!}{%
\begin{minipage}[c]{.3cm}
    \centering
    \rotatebox{90}{%
      \makebox[87px][c]{DeiT-III-B/16-224}%
      \makebox[87px][c]{DeiT-B/16-224}%
      \makebox[87px][c]{ViT-B/16-224}%
      \makebox[110px][c]{Input\quad}%
    }
\end{minipage}%
\begin{tabular}{@{}c@{}} \\ Hoopskirt \\ \includegraphics[width=0.18\textwidth]{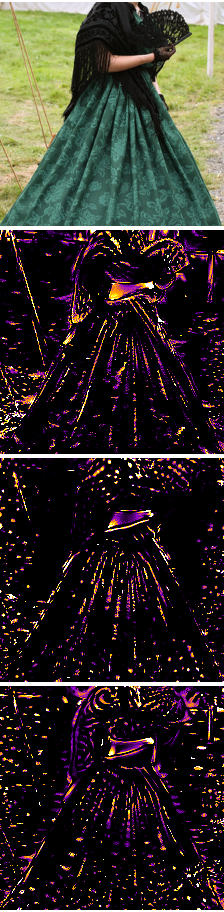} \end{tabular}
\begin{tabular}{@{}c@{}} \\ Hourglass \\ \includegraphics[width=0.18\textwidth]{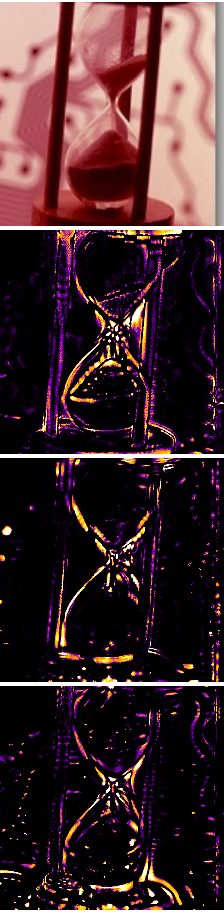} \end{tabular}
\begin{tabular}{@{}c@{}} \\ Hummingbird \\ \includegraphics[width=0.18\textwidth]{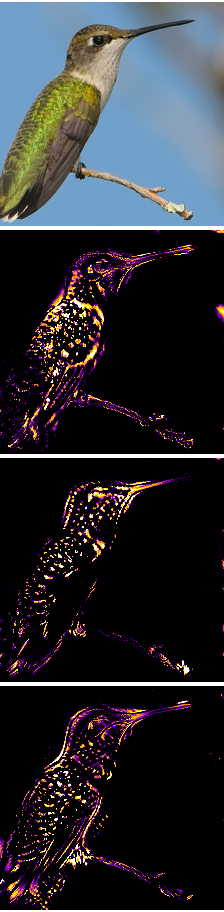} \end{tabular}
\begin{tabular}{@{}c@{}} \\ Iron \\ \includegraphics[width=0.18\textwidth]{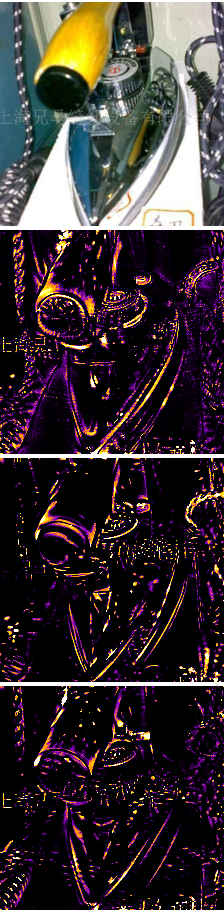} \end{tabular}
\begin{tabular}{@{}c@{}} \\ Macaw \\ \includegraphics[width=0.18\textwidth]{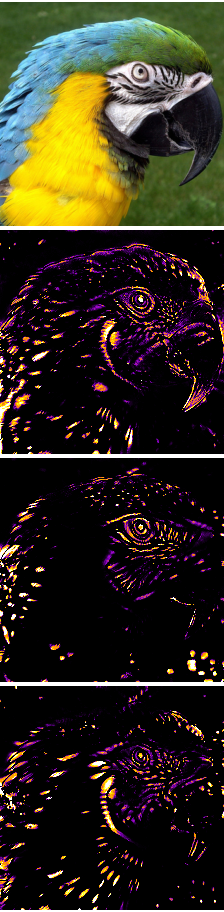} \end{tabular}
\begin{tabular}{@{}c@{}} \\ Mobile Home \\ \includegraphics[width=0.18\textwidth]{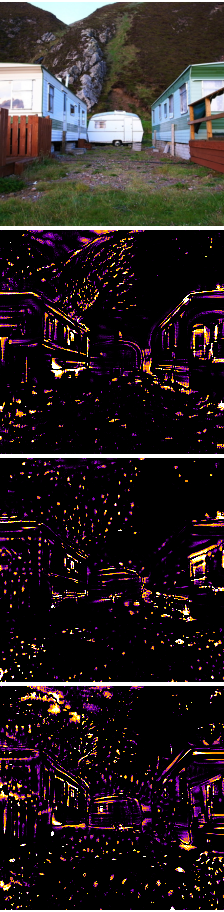} \end{tabular}
\begin{tabular}{@{}c@{}} \\ Revolver \\ \includegraphics[width=0.18\textwidth]{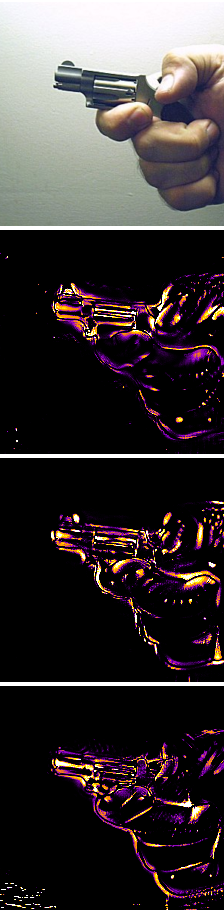} \end{tabular}
\begin{tabular}{@{}c@{}} \\ Streetcar \\ \includegraphics[width=0.18\textwidth]{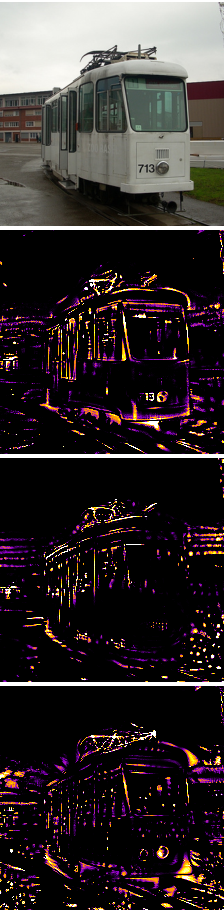} \end{tabular}
\begin{tabular}{@{}c@{}} West Highland \\ White Terrier \\ \includegraphics[width=0.18\textwidth]{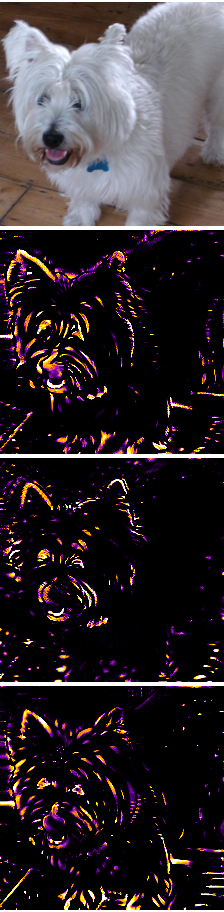} \end{tabular}
}
\caption{Comparison of class-token representations across different ViT architectures. For each ImageNet-1K validation sample, we visualize spatial maps derived from the class tokens extracted immediately before the final classification layer of ViT-B/16-224, DeiT-B/16-224, and DeiT-III-B/16-224. These maps reveal how information from image patches is aggregated into the global image representation used for recognition. Although the models often encode similar semantic cues, notable differences can be observed in the spatial distribution and concentration of the highlighted regions. Such variations reflect architecture-specific strategies for integrating local visual evidence into a compact class-level embedding.}
\label{fig:vit_token_class}
\end{figure*}

\Cref{fig:vit_token_class} presents explanations generated from class-token representations extracted immediately before the final classification layer. These maps visualize how information is aggregated into the global image representation used for recognition. While the models often attend to similar semantic concepts, they differ in the relative importance assigned to object parts and surrounding context. The final qualitative comparison includes attribution maps produced by \our{} and several established explanation approaches, including C-LRP, LeGrad, SmoothGrad, AttnLRP, Integrated Gradients, and DAVE. Compared with these baselines, our method produces compact and spatially coherent regions while preserving the semantic content required by the model. The examples also demonstrate that preservation-based explanations can highlight sufficient visual evidence without relying on architecture-specific attribution rules. These results support the quantitative findings presented in the main text and show that \our{} generalizes across different Vision Transformer architectures and explanation targets.

\section{ Sufficiency without the optimised model}

To test whether the sufficiency scores are specific to the model used to optimize the mask, we evaluate the same masks on a different backbone. At a 10\% pixel budget, a mask optimized on ViT-S/16 retains 53.6\% of the target-class probability on ConvNeXt-Tiny, compared with 58.8\% on the model used for optimization. This is higher than both IntGrad (28.1\%) and a random map (43.9\%). Conversely, a mask optimized on ConvNeXt-Tiny retains 60.3\% on ViT-S/16, compared with 67.7\% on its optimization model, again outperforming IntGrad (40.8\%) and a random map (51.6\%). Full results are reported in \cref{tab:cross-model-road}. Thus, the proposed masks retain most of their sufficiency when transferred to a backbone that did not participate in their optimization. In all cross-model comparisons, they outperform IntGrad, with the differences remaining statistically significant after Holm correction. These results suggest that the measured faithfulness is not solely an artifact of optimizing the masks against the evaluation model.

\begin{table}[h]
  \centering
  \small
  \setlength{\tabcolsep}{3pt}
  \resizebox{\columnwidth}{!}{%
  \begin{tabular}{@{}llcccc@{}}
    \toprule
     & & \multicolumn{2}{c}{scored on ViT-S/16} & \multicolumn{2}{c}{scored on ConvNeXt-Tiny} \\
    \cmidrule(lr){3-4} \cmidrule(lr){5-6}
    Mask from & Method & $10\%$ & $20\%$ & $10\%$ & $20\%$ \\
    \midrule
    ViT-S/16 & \our{} (ours) & ($\mathbf{58.8 \pm 39.3}$) & ($\mathbf{81.3 \pm 28.2}$) & $\mathbf{53.6 \pm 35.2}$ & $\mathbf{70.5 \pm 27.1}$ \\
     & IntGrad & ($28.3 \pm 36.5$) & ($74.9 \pm 31.1$) & $28.1 \pm 34.0$ & $61.2 \pm 32.3$ \\
     & random (scattered) & ($51.6 \pm 39.7$) & ($66.1 \pm 36.0$) & $43.9 \pm 35.6$ & $61.7 \pm 32.3$ \\
     & centre box & ($13.5 \pm 27.2$) & ($33.7 \pm 38.9$) & $17.4 \pm 29.5$ & $38.3 \pm 38.5$ \\
    \midrule
    ConvNeXt-Tiny & \our{} (ours) & $\mathbf{60.3 \pm 38.0}$ & $\mathbf{79.0 \pm 27.8}$ & ($\mathbf{67.7 \pm 29.3}$) & ($\mathbf{78.7 \pm 20.8}$) \\
     & IntGrad & $40.8 \pm 39.5$ & $68.4 \pm 37.4$ & ($47.1 \pm 37.7$) & ($72.9 \pm 27.4$) \\
     & random (scattered) & $51.6 \pm 39.7$ & $66.1 \pm 36.0$ & ($43.9 \pm 35.6$) & ($61.7 \pm 32.3$) \\
     & centre box & $13.5 \pm 27.2$ & $33.7 \pm 38.9$ & ($17.4 \pm 29.5$) & ($38.3 \pm 38.5$) \\
    \bottomrule
  \end{tabular}}
  \caption{\textbf{Sufficiency measured on a backbone that did not produce the mask.} Target-class probability (\%) after keeping the top $10\%$ / $20\%$ of pixels and imputing the rest with ROAD; mean $\pm$ std over 100 images classified correctly by both networks. Cells in parentheses are the circular measurement (mask and score come from the same network), while the other two columns of each row are the transfer, where the scoring network took no part in the optimisation. Two controls bracket the protocol: the centre box is a maximally concentrated keep-set and marks the floor, while the scattered random map is the best case for ROAD's linear imputer, which interpolates the image back from an even sample. It is therefore not a floor and \our{} does not beat it everywhere. Of the 24 paired comparisons 19 are significant after Holm-Bonferroni correction. }
  \label{tab:cross-model-road}
\end{table}

\begin{figure*}
    \centering
    \includegraphics[width=\textwidth]{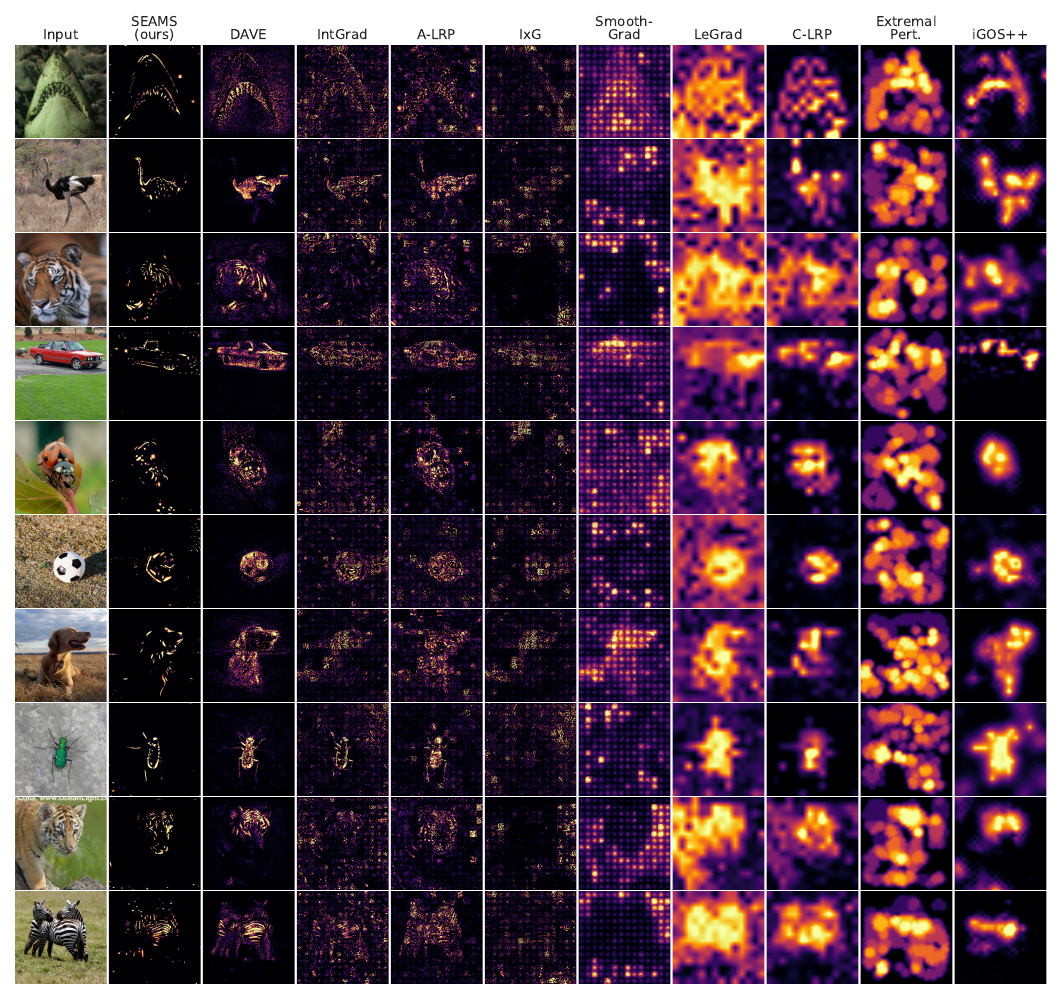}
    \caption{\textbf{Extended qualitative comparison.} Ten ImageNet-1K
    validation images (rows; indices $188$, $458$, $14617$, $40854$, $15069$,
    $40258$, $10395$, $15012$, $14602$, $17044$ of the sorted \texttt{val}
    split) with the input and the attribution map of every method in the
    benchmark (columns), including the two optimisation-based baselines added
    in this revision, Extremal Perturbations and iGOS++. The backbone is the
    frozen ViT-B/16 of \cref{tab:pixdel} and every method is given the
    ground-truth class; nine of the ten images are classified correctly, the
    exception being row~4 (\emph{sports car} predicted as \emph{convertible}).
    As in \cref{fig:posthoc}, \our{} is shown in the configuration of the
    paper ($T\!=\!2000$ steps, one optimisation copy) while each baseline keeps
    its benchmark configuration; the $T\!=\!500$, three-copy median masks the
    faithfulness benchmark actually scores have the same structure but are
    visibly noisier. Maps are min-max normalised after clipping the upper
    percentile. Both DAVE and \our{} give sharper, more object-aligned
    explanations with fewer patch-grid artifacts than the single-pass
    techniques; \our{} is the sparsest, concentrating on the few regions that
    are sufficient on their own.}
    \label{fig:posthoc-supp}
\end{figure*}

\begin{table*}[p]
  \centering
  \scriptsize
  \setlength{\tabcolsep}{4pt}
  \begin{tabular}{@{}lrrrlrrrl@{}}
    \toprule
    \multicolumn{9}{c}{\textbf{ViT-B/16}} \\
    \midrule
    Metric & \multicolumn{1}{c}{Baseline} & $\Delta$ mean & 95\,\% CI of $\Delta$ & Test & $p$ & $p_{\text{Holm}}$ & Effect & Sig. \\
    \midrule
    Deletion (grey) & IxG & +26.35 & [+24.49, +28.17] & W & $1.70\!\times\!10^{-67}$ & $1.36\!\times\!10^{-66}$ & +0.97 & yes \\
    Deletion (grey) & SmoothGrad & +27.87 & [+26.55, +29.10] & W & $7.72\!\times\!10^{-70}$ & $6.95\!\times\!10^{-69}$ & +0.99 & yes \\
    Deletion (grey) & IntGrad & +14.09 & [+12.43, +15.79] & W & $4.54\!\times\!10^{-46}$ & $1.36\!\times\!10^{-45}$ & +0.80 & yes \\
    Deletion (grey) & A-LRP & +13.30 & [+11.74, +15.07] & W & $3.49\!\times\!10^{-49}$ & $1.75\!\times\!10^{-48}$ & +0.82 & yes \\
    Deletion (grey) & C-LRP & +12.35 & [+11.15, +13.58] & W & $2.63\!\times\!10^{-52}$ & $1.58\!\times\!10^{-51}$ & +0.85 & yes \\
    Deletion (grey) & LeGrad & +11.74 & [+10.49, +13.03] & W & $6.89\!\times\!10^{-49}$ & $2.76\!\times\!10^{-48}$ & +0.82 & yes \\
    Deletion (grey) & DAVE & +6.18 & [+4.92, +7.71] & W & $1.12\!\times\!10^{-17}$ & $1.12\!\times\!10^{-17}$ & +0.48 & yes \\
    Deletion (grey) & iGOS++ & +8.82 & [+7.70, +9.95] & W & $2.46\!\times\!10^{-42}$ & $4.92\!\times\!10^{-42}$ & +0.76 & yes \\
    Deletion (grey) & Extremal Pert. & +12.38 & [+11.14, +13.61] & W & $5.29\!\times\!10^{-53}$ & $3.70\!\times\!10^{-52}$ & +0.86 & yes \\
    Insertion (grey) & IxG & +30.37 & [+28.53, +32.14] & W & $1.50\!\times\!10^{-69}$ & $1.20\!\times\!10^{-68}$ & +0.99 & yes \\
    Insertion (grey) & SmoothGrad & +32.11 & [+30.66, +33.42] & W & $2.36\!\times\!10^{-70}$ & $2.12\!\times\!10^{-69}$ & +0.99 & yes \\
    Insertion (grey) & IntGrad & +17.90 & [+16.25, +19.55] & W & $2.53\!\times\!10^{-60}$ & $1.77\!\times\!10^{-59}$ & +0.92 & yes \\
    Insertion (grey) & A-LRP & +15.77 & [+14.18, +17.49] & W & $2.77\!\times\!10^{-56}$ & $1.11\!\times\!10^{-55}$ & +0.88 & yes \\
    Insertion (grey) & C-LRP & +15.51 & [+14.21, +16.82] & W & $2.21\!\times\!10^{-57}$ & $1.11\!\times\!10^{-56}$ & +0.89 & yes \\
    Insertion (grey) & LeGrad & +14.97 & [+13.63, +16.31] & W & $8.03\!\times\!10^{-55}$ & $2.41\!\times\!10^{-54}$ & +0.87 & yes \\
    Insertion (grey) & DAVE & +7.24 & [+5.90, +8.84] & W & $1.29\!\times\!10^{-21}$ & $1.29\!\times\!10^{-21}$ & +0.53 & yes \\
    Insertion (grey) & iGOS++ & +11.76 & [+10.54, +12.96] & W & $2.48\!\times\!10^{-49}$ & $4.95\!\times\!10^{-49}$ & +0.83 & yes \\
    Insertion (grey) & Extremal Pert. & +16.14 & [+14.82, +17.45] & W & $3.76\!\times\!10^{-58}$ & $2.26\!\times\!10^{-57}$ & +0.90 & yes \\
    ROAD LeRF & IxG & +0.54 & [-0.57, +1.68] & W & $0.0447$ & $0.0447$ & +0.11 & yes \\
    ROAD LeRF & SmoothGrad & +14.42 & [+13.21, +15.69] & W & $8.58\!\times\!10^{-63}$ & $7.73\!\times\!10^{-62}$ & +0.94 & yes \\
    ROAD LeRF & IntGrad & -2.91 & [-3.87, -2.06] & W & $4.14\!\times\!10^{-12}$ & $8.28\!\times\!10^{-12}$ & -0.39 & yes \\
    ROAD LeRF & A-LRP & +3.58 & [+2.83, +4.51] & W & $6.19\!\times\!10^{-29}$ & $3.10\!\times\!10^{-28}$ & +0.62 & yes \\
    ROAD LeRF & C-LRP & +8.59 & [+7.40, +9.80] & W & $2.39\!\times\!10^{-41}$ & $1.80\!\times\!10^{-40}$ & +0.75 & yes \\
    ROAD LeRF & LeGrad & +8.83 & [+7.57, +10.15] & W & $1.08\!\times\!10^{-40}$ & $6.46\!\times\!10^{-40}$ & +0.75 & yes \\
    ROAD LeRF & DAVE & +3.49 & [+2.75, +4.32] & W & $9.05\!\times\!10^{-24}$ & $3.62\!\times\!10^{-23}$ & +0.56 & yes \\
    ROAD LeRF & iGOS++ & +4.02 & [+2.99, +5.05] & W & $6.32\!\times\!10^{-19}$ & $1.90\!\times\!10^{-18}$ & +0.50 & yes \\
    ROAD LeRF & Extremal Pert. & +9.75 & [+8.50, +11.04] & W & $2.25\!\times\!10^{-41}$ & $1.80\!\times\!10^{-40}$ & +0.75 & yes \\
    ROAD @10\% & IxG & +46.72 & [+43.34, +49.88] & W & $9.26\!\times\!10^{-64}$ & $6.48\!\times\!10^{-63}$ & +0.94 & yes \\
    ROAD @10\% & SmoothGrad & +59.97 & [+56.74, +63.03] & W & $1.31\!\times\!10^{-68}$ & $1.18\!\times\!10^{-67}$ & +0.98 & yes \\
    ROAD @10\% & IntGrad & +23.26 & [+20.12, +26.52] & W & $5.53\!\times\!10^{-39}$ & $1.11\!\times\!10^{-38}$ & +0.73 & yes \\
    ROAD @10\% & A-LRP & +19.46 & [+16.77, +22.26] & W & $6.73\!\times\!10^{-41}$ & $2.02\!\times\!10^{-40}$ & +0.75 & yes \\
    ROAD @10\% & C-LRP & +43.46 & [+39.97, +46.97] & W & $3.56\!\times\!10^{-59}$ & $1.78\!\times\!10^{-58}$ & +0.91 & yes \\
    ROAD @10\% & LeGrad & +48.19 & [+44.59, +51.79] & W & $3.82\!\times\!10^{-61}$ & $2.29\!\times\!10^{-60}$ & +0.92 & yes \\
    ROAD @10\% & DAVE & +17.83 & [+15.20, +20.74] & W & $1.31\!\times\!10^{-34}$ & $1.31\!\times\!10^{-34}$ & +0.69 & yes \\
    ROAD @10\% & iGOS++ & +31.44 & [+27.99, +34.89] & W & $2.34\!\times\!10^{-47}$ & $9.36\!\times\!10^{-47}$ & +0.81 & yes \\
    ROAD @10\% & Extremal Pert. & +50.57 & [+47.08, +53.95] & W & $5.73\!\times\!10^{-64}$ & $4.59\!\times\!10^{-63}$ & +0.94 & yes \\
    ROAD @20\% & IxG & +15.45 & [+12.78, +18.34] & W & $1.50\!\times\!10^{-25}$ & $5.99\!\times\!10^{-25}$ & +0.58 & yes \\
    ROAD @20\% & SmoothGrad & +41.46 & [+38.06, +44.81] & W & $7.67\!\times\!10^{-62}$ & $6.90\!\times\!10^{-61}$ & +0.93 & yes \\
    ROAD @20\% & IntGrad & +4.94 & [+3.03, +7.08] & W & $0.0003$ & $0.0003$ & +0.20 & yes \\
    ROAD @20\% & A-LRP & +8.48 & [+6.81, +10.43] & W & $7.10\!\times\!10^{-29}$ & $3.55\!\times\!10^{-28}$ & +0.62 & yes \\
    ROAD @20\% & C-LRP & +23.41 & [+20.38, +26.59] & W & $3.31\!\times\!10^{-35}$ & $1.99\!\times\!10^{-34}$ & +0.69 & yes \\
    ROAD @20\% & LeGrad & +26.78 & [+23.51, +30.25] & W & $3.73\!\times\!10^{-39}$ & $2.61\!\times\!10^{-38}$ & +0.73 & yes \\
    ROAD @20\% & DAVE & +9.52 & [+7.60, +11.78] & W & $1.28\!\times\!10^{-23}$ & $3.84\!\times\!10^{-23}$ & +0.56 & yes \\
    ROAD @20\% & iGOS++ & +12.69 & [+10.21, +15.39] & W & $3.18\!\times\!10^{-17}$ & $6.36\!\times\!10^{-17}$ & +0.47 & yes \\
    ROAD @20\% & Extremal Pert. & +32.55 & [+29.09, +36.06] & W & $2.41\!\times\!10^{-44}$ & $1.93\!\times\!10^{-43}$ & +0.78 & yes \\
    ROAD MoRF & IxG & -15.39 & [-17.37, -13.37] & W & $1.65\!\times\!10^{-36}$ & $8.27\!\times\!10^{-36}$ & -0.71 & yes \\
    ROAD MoRF & SmoothGrad & -13.05 & [-14.87, -11.29] & W & $2.59\!\times\!10^{-35}$ & $7.78\!\times\!10^{-35}$ & -0.69 & yes \\
    ROAD MoRF & IntGrad & -11.56 & [-13.52, -9.57] & W & $1.59\!\times\!10^{-23}$ & $3.17\!\times\!10^{-23}$ & -0.56 & yes \\
    ROAD MoRF & A-LRP & -26.72 & [-28.51, -24.88] & W & $1.15\!\times\!10^{-66}$ & $1.04\!\times\!10^{-65}$ & -0.96 & yes \\
    ROAD MoRF & C-LRP & +16.45 & [+14.63, +18.38] & W & $4.42\!\times\!10^{-46}$ & $2.65\!\times\!10^{-45}$ & +0.80 & yes \\
    ROAD MoRF & LeGrad & +16.52 & [+14.69, +18.45] & W & $1.25\!\times\!10^{-46}$ & $8.77\!\times\!10^{-46}$ & +0.80 & yes \\
    ROAD MoRF & DAVE & -25.02 & [-26.83, -23.11] & W & $3.00\!\times\!10^{-64}$ & $2.40\!\times\!10^{-63}$ & -0.95 & yes \\
    ROAD MoRF & iGOS++ & +12.33 & [+10.57, +14.13] & W & $4.99\!\times\!10^{-36}$ & $2.00\!\times\!10^{-35}$ & +0.70 & yes \\
    ROAD MoRF & Extremal Pert. & -4.68 & [-6.44, -2.89] & W & $5.74\!\times\!10^{-7}$ & $5.74\!\times\!10^{-7}$ & -0.28 & yes \\
    \bottomrule
  \end{tabular}
  \caption{\textbf{Paired comparisons of \our{} against every baseline.} $\Delta$ is the per-image mean difference \our{}~$-$~baseline with its BCa bootstrap 95\,\% confidence interval (10\,000 resamples, fixed seed). The test is a paired $t$-test (t) when Shapiro-Wilk on the paired differences and Levene's test both pass at $\alpha=0.05$, and the Wilcoxon signed-rank test (W) otherwise. $p_{\text{Holm}}$ is corrected within each metric-backbone family. Effect size is Cohen's $d_z$ for $t$ and the matched-pairs rank-biserial correlation for W.}
  \label{tab:stats-full}
\end{table*}

\begin{table*}[p]
  \centering
  \scriptsize
  \setlength{\tabcolsep}{4pt}
  \begin{tabular}{@{}lrrrlrrrl@{}}
    \toprule
    \multicolumn{9}{c}{\textbf{DeiT-III-B/16}} \\
    \midrule
    Metric & \multicolumn{1}{c}{Baseline} & $\Delta$ mean & 95\,\% CI of $\Delta$ & Test & $p$ & $p_{\text{Holm}}$ & Effect & Sig. \\
    \midrule
    Deletion (grey) & IxG & +43.30 & [+41.29, +45.25] & W & $2.41\!\times\!10^{-69}$ & $1.93\!\times\!10^{-68}$ & +0.99 & yes \\
    Deletion (grey) & SmoothGrad & +26.50 & [+24.73, +28.12] & W & $2.73\!\times\!10^{-64}$ & $1.91\!\times\!10^{-63}$ & +0.95 & yes \\
    Deletion (grey) & IntGrad & +26.91 & [+24.76, +29.09] & W & $4.61\!\times\!10^{-61}$ & $2.30\!\times\!10^{-60}$ & +0.92 & yes \\
    Deletion (grey) & A-LRP & +25.43 & [+23.46, +27.44] & W & $5.59\!\times\!10^{-64}$ & $3.35\!\times\!10^{-63}$ & +0.95 & yes \\
    Deletion (grey) & LeGrad & +8.22 & [+6.66, +9.79] & W & $3.70\!\times\!10^{-30}$ & $1.11\!\times\!10^{-29}$ & +0.64 & yes \\
    Deletion (grey) & DAVE & +9.48 & [+7.94, +11.27] & W & $3.17\!\times\!10^{-25}$ & $6.33\!\times\!10^{-25}$ & +0.58 & yes \\
    Deletion (grey) & iGOS++ & +6.25 & [+4.77, +7.65] & W & $9.19\!\times\!10^{-25}$ & $9.19\!\times\!10^{-25}$ & +0.58 & yes \\
    Deletion (grey) & Extremal Pert. & +15.04 & [+13.32, +16.70] & W & $4.29\!\times\!10^{-47}$ & $1.71\!\times\!10^{-46}$ & +0.81 & yes \\
    Insertion (grey) & IxG & +45.54 & [+43.59, +47.46] & W & $6.49\!\times\!10^{-70}$ & $5.19\!\times\!10^{-69}$ & +0.99 & yes \\
    Insertion (grey) & SmoothGrad & +29.12 & [+27.30, +30.77] & W & $3.87\!\times\!10^{-66}$ & $2.32\!\times\!10^{-65}$ & +0.96 & yes \\
    Insertion (grey) & IntGrad & +29.02 & [+26.90, +31.15] & W & $5.74\!\times\!10^{-64}$ & $2.87\!\times\!10^{-63}$ & +0.95 & yes \\
    Insertion (grey) & A-LRP & +27.66 & [+25.70, +29.61] & W & $1.87\!\times\!10^{-66}$ & $1.31\!\times\!10^{-65}$ & +0.96 & yes \\
    Insertion (grey) & LeGrad & +10.44 & [+8.84, +12.05] & W & $5.82\!\times\!10^{-36}$ & $1.75\!\times\!10^{-35}$ & +0.70 & yes \\
    Insertion (grey) & DAVE & +9.84 & [+8.27, +11.69] & W & $1.09\!\times\!10^{-24}$ & $1.09\!\times\!10^{-24}$ & +0.57 & yes \\
    Insertion (grey) & iGOS++ & +8.07 & [+6.53, +9.56] & W & $1.05\!\times\!10^{-28}$ & $2.10\!\times\!10^{-28}$ & +0.62 & yes \\
    Insertion (grey) & Extremal Pert. & +17.53 & [+15.74, +19.26] & W & $6.89\!\times\!10^{-51}$ & $2.76\!\times\!10^{-50}$ & +0.84 & yes \\
    ROAD LeRF & IxG & +3.49 & [+2.15, +4.94] & W & $1.17\!\times\!10^{-12}$ & $2.34\!\times\!10^{-12}$ & +0.40 & yes \\
    ROAD LeRF & SmoothGrad & +14.00 & [+12.61, +15.42] & W & $3.18\!\times\!10^{-56}$ & $2.54\!\times\!10^{-55}$ & +0.88 & yes \\
    ROAD LeRF & IntGrad & -0.50 & [-1.62, +0.60] & W & $0.7220$ & $0.7220$ & +0.02 & no \\
    ROAD LeRF & A-LRP & +3.08 & [+2.19, +3.97] & W & $3.53\!\times\!10^{-22}$ & $1.06\!\times\!10^{-21}$ & +0.54 & yes \\
    ROAD LeRF & LeGrad & +9.47 & [+7.95, +11.01] & W & $4.53\!\times\!10^{-44}$ & $3.17\!\times\!10^{-43}$ & +0.78 & yes \\
    ROAD LeRF & DAVE & +3.31 & [+2.52, +4.20] & W & $4.34\!\times\!10^{-24}$ & $2.17\!\times\!10^{-23}$ & +0.57 & yes \\
    ROAD LeRF & iGOS++ & +3.24 & [+2.03, +4.38] & W & $1.38\!\times\!10^{-23}$ & $5.53\!\times\!10^{-23}$ & +0.56 & yes \\
    ROAD LeRF & Extremal Pert. & +9.88 & [+8.33, +11.55] & W & $8.94\!\times\!10^{-42}$ & $5.36\!\times\!10^{-41}$ & +0.76 & yes \\
    ROAD @10\% & IxG & +46.93 & [+43.32, +50.54] & W & $3.84\!\times\!10^{-60}$ & $2.69\!\times\!10^{-59}$ & +0.92 & yes \\
    ROAD @10\% & SmoothGrad & +62.82 & [+59.31, +66.12] & W & $4.54\!\times\!10^{-67}$ & $3.63\!\times\!10^{-66}$ & +0.97 & yes \\
    ROAD @10\% & IntGrad & +26.69 & [+23.40, +30.23] & W & $7.80\!\times\!10^{-40}$ & $2.34\!\times\!10^{-39}$ & +0.74 & yes \\
    ROAD @10\% & A-LRP & +25.70 & [+22.64, +28.95] & W & $6.30\!\times\!10^{-42}$ & $2.52\!\times\!10^{-41}$ & +0.76 & yes \\
    ROAD @10\% & LeGrad & +48.98 & [+45.01, +52.90] & W & $2.16\!\times\!10^{-57}$ & $1.08\!\times\!10^{-56}$ & +0.89 & yes \\
    ROAD @10\% & DAVE & +17.37 & [+14.49, +20.43] & W & $2.88\!\times\!10^{-33}$ & $2.88\!\times\!10^{-33}$ & +0.67 & yes \\
    ROAD @10\% & iGOS++ & +31.23 & [+27.42, +35.20] & W & $2.49\!\times\!10^{-38}$ & $4.98\!\times\!10^{-38}$ & +0.72 & yes \\
    ROAD @10\% & Extremal Pert. & +50.13 & [+46.31, +53.92] & W & $1.27\!\times\!10^{-59}$ & $7.62\!\times\!10^{-59}$ & +0.91 & yes \\
    ROAD @20\% & IxG & +13.98 & [+11.20, +16.97] & W & $1.68\!\times\!10^{-14}$ & $5.04\!\times\!10^{-14}$ & +0.43 & yes \\
    ROAD @20\% & SmoothGrad & +41.27 & [+37.40, +45.10] & W & $4.45\!\times\!10^{-52}$ & $3.56\!\times\!10^{-51}$ & +0.85 & yes \\
    ROAD @20\% & IntGrad & +4.12 & [+2.11, +6.29] & W & $0.0542$ & $0.0542$ & +0.11 & no \\
    ROAD @20\% & A-LRP & +7.01 & [+5.14, +9.07] & W & $4.78\!\times\!10^{-14}$ & $9.56\!\times\!10^{-14}$ & +0.42 & yes \\
    ROAD @20\% & LeGrad & +27.00 & [+23.39, +30.74] & W & $2.59\!\times\!10^{-38}$ & $1.56\!\times\!10^{-37}$ & +0.72 & yes \\
    ROAD @20\% & DAVE & +7.75 & [+5.80, +9.97] & W & $1.90\!\times\!10^{-16}$ & $7.59\!\times\!10^{-16}$ & +0.46 & yes \\
    ROAD @20\% & iGOS++ & +10.94 & [+8.06, +14.07] & W & $6.63\!\times\!10^{-17}$ & $3.31\!\times\!10^{-16}$ & +0.47 & yes \\
    ROAD @20\% & Extremal Pert. & +28.53 & [+24.97, +32.28] & W & $1.99\!\times\!10^{-38}$ & $1.40\!\times\!10^{-37}$ & +0.73 & yes \\
    ROAD MoRF & IxG & -25.51 & [-27.99, -23.01] & W & $6.08\!\times\!10^{-50}$ & $3.65\!\times\!10^{-49}$ & -0.83 & yes \\
    ROAD MoRF & SmoothGrad & -12.45 & [-14.29, -10.59] & W & $6.57\!\times\!10^{-33}$ & $1.97\!\times\!10^{-32}$ & -0.67 & yes \\
    ROAD MoRF & IntGrad & -22.08 & [-24.52, -19.58] & W & $2.54\!\times\!10^{-42}$ & $1.27\!\times\!10^{-41}$ & -0.76 & yes \\
    ROAD MoRF & A-LRP & -33.18 & [-35.35, -30.97] & W & $1.66\!\times\!10^{-66}$ & $1.33\!\times\!10^{-65}$ & -0.96 & yes \\
    ROAD MoRF & LeGrad & +15.45 & [+13.38, +17.55] & W & $3.25\!\times\!10^{-35}$ & $1.30\!\times\!10^{-34}$ & +0.69 & yes \\
    ROAD MoRF & DAVE & -22.88 & [-25.01, -20.73] & W & $4.69\!\times\!10^{-54}$ & $3.28\!\times\!10^{-53}$ & -0.87 & yes \\
    ROAD MoRF & iGOS++ & +11.21 & [+9.35, +13.12] & W & $4.73\!\times\!10^{-26}$ & $9.47\!\times\!10^{-26}$ & +0.59 & yes \\
    ROAD MoRF & Extremal Pert. & -8.62 & [-10.83, -6.30] & t & $2.50\!\times\!10^{-13}$ & $2.50\!\times\!10^{-13}$ & -0.37 & yes \\
    \bottomrule
  \end{tabular}
  \caption{\textbf{Paired comparisons on DeiT-III-B/16} --- continued from \cref{tab:stats-full}, same conventions.}
  \label{tab:stats-full-deit3}
\end{table*}

\end{document}